%% file: paper_arxiv_preprint.tex
\documentclass{article}

\usepackage[final]{neurips_2019}

\PassOptionsToPackage{numbers, compress}{natbib}
\usepackage[utf8]{inputenc} 
\usepackage[T1]{fontenc}    
\usepackage[hidelinks]{hyperref}       
\usepackage{booktabs}       
\usepackage{amsfonts}       
\usepackage{nicefrac}       
\usepackage{microtype}      
\usepackage{bm}
\usepackage{amsmath}
\usepackage{amsthm}
\usepackage{amssymb}
\usepackage{graphicx,color}
\usepackage{wrapfig}
\usepackage{algorithm}
\usepackage[noend]{algpseudocode}
\usepackage{xspace}
\usepackage[table,xcdraw]{xcolor}
\usepackage{extarrows}
\usepackage{xfrac}
\usepackage{tcolorbox}
\usepackage{sidecap}
\usepackage{subfig}
\usepackage{multirow}
\usepackage{enumitem}
\sidecaptionvpos{figure}{t}

\input{macros}

\makeatletter
\newcommand{\printfnsymbol}[1]{%
  \textsuperscript{\@fnsymbol{#1}}%
}
\makeatother

\title{Reliable training and estimation of variance networks}

\author{%
  Nicki S. Detlefsen\thanks{Equal contribution} $\:$ \thanks{Section for Cognitive Systems, Technical University of Denmark} \\
  \texttt{nsde@dtu.dk} \\
  \And
  Martin Jørgensen\printfnsymbol{1} \printfnsymbol{2} \\
  \texttt{marjor@dtu.dk}
  \And
  Søren Hauberg \printfnsymbol{2}\\
  \texttt{sohau@dtu.dk}
}

\begin{document}

\maketitle

\begin{abstract}
\input{abstract}
\end{abstract}

\section{Introduction}\label{sec:intro}
\input{introduction}

\section{Related work}
\input{related_work}

\section{Methods}\label{sec:meth}
\input{methods}

\section{Experiments}\label{sec:results}
\input{experiments}

\section{Discussion \& Conclusion}
\input{discussion}

{\small
\bibliographystyle{abbrvnat}
\bibliography{references}
}
\newpage
\begin{appendix}
	\input{appendix}
\end{appendix}
\end{document}

%% file: macros.tex
\makeatletter
\DeclareRobustCommand\onedot{\futurelet\@let@token\@onedot}
\def\@onedot{\ifx\@let@token.\else.\null\fi\xspace}

\def\eg{\emph{e.g}\onedot} 

\def\ie{\emph{i.e}\onedot}

\makeatother

\renewcommand{\vec}[1]{\mathbf{#1}}

\newcommand{\x}[0]{\vec{x}}

\newcommand{\z}[0]{\vec{z}}

\newcommand{\Ncal}{\mathcal{N}}
\bmdefine\btheta{\theta}
\bmdefine\bgamma{\gamma}
\bmdefine\bmu{\mu}
\bmdefine\bsigma{\sigma}

\def\x{{\boldsymbol x}}
\def\z{{\boldsymbol z}}

\newcommand{\FIG}{Fig.~}
\newcommand{\FIGS}{Figs.~}
\newcommand{\EQN}{Eq.~}

\newcommand{\TAB}{Table~}
\newcommand{\SUPMAT}{supplementary material}



%% file: abstract.tex

We propose and investigate new complementary methodologies for estimating predictive variance networks in regression neural networks. We derive a locally aware mini-batching scheme that results in sparse robust gradients, and we show how to make unbiased weight updates to a variance network. Further, we formulate a heuristic for robustly fitting both the mean and variance networks post hoc. Finally, we take inspiration from posterior Gaussian processes and propose a network architecture with similar extrapolation properties to Gaussian processes. The proposed methodologies are complementary, and improve upon baseline methods individually. Experimentally, we investigate the impact of predictive uncertainty on multiple datasets and tasks ranging from regression, active learning and generative modeling. Experiments consistently show significant improvements in predictive uncertainty estimation over state-of-the-art methods across tasks and datasets.

%% file: introduction.tex
The quality of \emph{mean} predictions has dramatically increased in the last decade
with the rediscovery of neural networks \citep{lecun2015deeplearning}. 
The predictive \emph{variance}, however, has turned out to be a more elusive target,
with established solutions being subpar. The general finding is that neural networks
tend to make overconfident predictions \citep{Guo2017calibration} that can be harmful or offensive \citep{amodei2016safety}.
This may be explained by neural networks being general function estimators
that does not come with principled uncertainty estimates.
Another explanation is that \emph{variance} estimation is a fundamentally different
task than \emph{mean} estimation, and that the tools for mean estimation perhaps
do not generalize. We focus on the latter hypothesis within regression.

\begin{wrapfigure}[10]{R}{0.3\textwidth}
  \vspace{-8mm}
\includegraphics[width=0.3\textwidth]{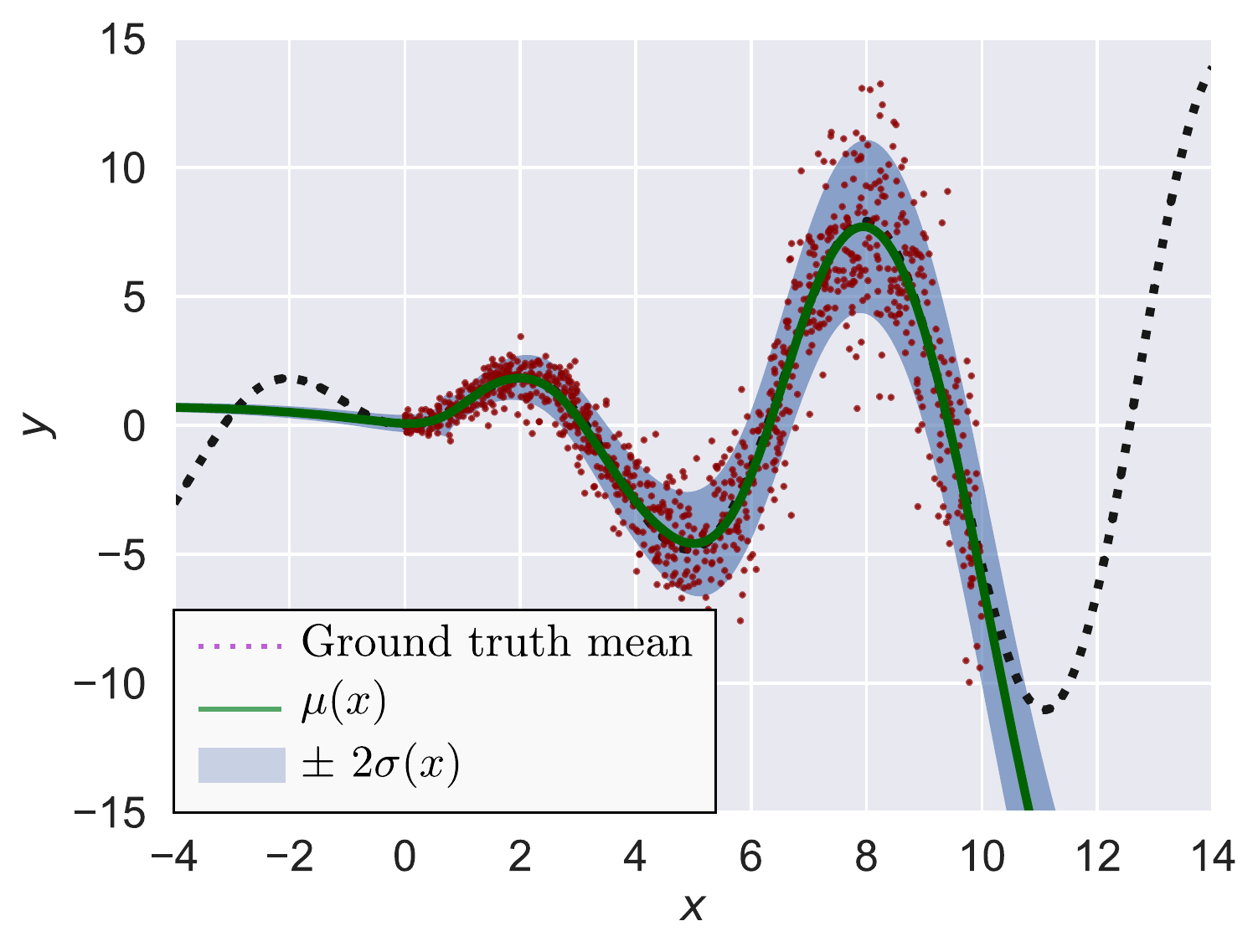}
  \caption{Max.\ likelihood fit of $\Ncal(\mu(x), \sigma^2(x))$ to data.}
  \label{fig:toy}
\end{wrapfigure}
To illustrate the main practical problems
in variance estimation, we consider a toy problem where data is generated as
$y = x\cdot \sin(x) + 0.3 \cdot \epsilon_{1} + 0.3 \cdot x \cdot \epsilon_{2}$,
with $\epsilon_{1}, \epsilon_{2} \sim \Ncal(0,1)$ and $x$ is uniform on $[0,10]$
(Fig.~\ref{fig:toy}). As is common, we do maximum likelihood estimation of
$\Ncal(\mu(x), \sigma^2(x))$, where $\mu$ and $\sigma^2$ are neural nets.
While $\mu$ provides an almost perfect fit to the ground truth, $\sigma^2$ shows two problems: $\sigma^2$ is significantly underestimated and $\sigma^2$ does not increase outside the data support to capture the poor mean predictions.

%
These findings are general (Sec.~\ref{sec:results}), and alleviating them is
the main purpose of the present paper. We find that this can be achieved by
a combination of methods that
{\emph{1)}} change the usual mini-batching to be location aware; 
{\emph{2)}} only optimize variance conditioned on the mean;
{\emph{3)}} for scarce data, we introduce a more robust likelihood function; and
{\emph{4)}} enforce well-behaved interpolation and extrapolation of variances. Points 1 and 2 are achieved through changes to the training algorithm, while 3 and 4 are changes to model specifications. We empirically demonstrate that these new tools significantly improve on
state-of-the-art across datasets in tasks ranging from regression to active learning, and generative modeling.

%% file: related_work.tex
\textbf{Gaussian processes (GPs)} are well-known function approximators with built-in uncertainty estimators \citep{rasmussen:book}. GPs are robust in settings with a low amount of data, and can model a rich class of functions with few hyperparameters. However, GPs are computationally intractable for large amounts of data and limited by the expressiveness of a chosen kernel. Advances like sparse and deep GPs \citep{snelson2006sparse, damianou} partially alleviate this, but neural nets still tend to have more accurate mean predictions.

\textbf{Uncertainty aware neural networks} model the predictive mean and variance
as two separate neural networks, often as multi-layer perceptrons. This originates with the work of \citet{Nix1994estimating}
and \citet{Bishop94mixturedensity}; today, the approach is commonly used for
making variational approximations \citep{kingma:iclr:2014,rezende2014stochastic},
and it is this general approach we investigate.

\textbf{Bayesian neural networks (BNN)} \citep{MacKay2008bnn} assume a prior distribution
over the network parameters, and approximate the posterior distribution. This
gives direct access to the approximate predictive uncertainty.
In practice, placing an informative prior over the parameters is non-trivial.
Even with advances in stochastic variational inference \citep{kingma:iclr:2014,rezende2014stochastic,Hoffman2012vi}
and expectation propagation \citep{Hernandez2015backprob}, it is still challenging
to perform inference in BNNs.

\textbf{Ensemble methods} represent the current state-of-the-art.
\emph{Monte Carlo (MC) Dropout} \citep{Gal2015dropout} measure the uncertainty induced by Dropout layers \citep{hinton2012improving}
arguing that this is a good proxy for predictive uncertainty.
\emph{Deep Ensembles} \citep{Lakshminarayanan2016ensembles} form an ensemble
from multiple neural networks trained with different initializations.
Both approaches obtain ensembles of \emph{correlated} networks, and the extent to which this biases the predictive uncertainty is unclear.
Alternatives include estimating \emph{confidence intervals} instead of variances
\citep{Pearce2018intervals}, and gradient-based Bayesian model
averaging \citep{Maddox2019baseline}.

\textbf{Applications of uncertainty} include \emph{reinforcement learning},
\emph{active learning}, and \emph{Bayesian optimization} \citep{szepesvari2010rl, huang2010al, Frazier2018bayesoptim}.
Here, uncertainty is the crucial element that allows for systematically making a trade-off
between \emph{exploration} and \emph{exploitation}.
It has also been shown that uncertainty is required to learn the topology
of data manifolds \citep{Hauberg2018onlybayes}.

\textbf{The main categories of uncertainty} are \emph{epistemic} and \emph{aleatoric} uncertainty \citep{kiureghian2009doesitmatter, kendall2017uncertainties}.
Aleatoric uncertainty is induced by unknown or unmeasured features, and, hence, does not vanish in the limit of infinite data.
Epistemic uncertainty is often referred to as \emph{model uncertainty}, as it is the uncertainty due to model limitations. It is this type of uncertainty that Bayesian and ensemble methods generally estimate.
We focus on the overall \emph{predictive uncertainty}, which reflects both epistemic and aleatoric uncertainty.


%% file: methods.tex
The opening remarks (Sec.~\ref{sec:intro}) highlighted two common problems
that appear when $\mu$ and $\sigma^2$ are neural networks. In this section
we analyze these problems and propose solutions.

\paragraph{Preliminaries.}
We assume that datasets
$\mathcal{D} = \{\x_i,y_i\}_{i=1}^N$ contain i.i.d.\ observations
$y_i \in \mathbb{R}, \x_i \in \mathbb{R}^D$. The targets $y_i$ are assumed
to be conditionally Gaussian,
$p_{\theta} (y | \x) = \mathcal{N}(y | \mu(\x),\sigma^2(\x))$, 
where $\mu$ and $\sigma^2$ are continuous functions parametrized by $\theta = \{ \theta_{\mu}, \theta_{\sigma^2} \}$.
The maximum likelihood estimate (MLE) of the variance of i.i.d.\ observations
$\{y_i\}_{i=1}^N$ is $\frac{1}{N-1}\sum_i (y_i - \hat{\mu})^2$, where $\hat{\mu}$ is the sample mean.
This MLE does not exist based on a single observation, unless the mean $\mu$ is known,
i.e.\ the mean is not a free parameter. When $y_i$ is Gaussian, the residuals $(y_i - \mu)^2$
are gamma distributed.

\subsection{A local likelihood model analysis}
By assuming that both $\mu$ and $\sigma^2$ are continuous functions, we are implicitly saying 
that $\sigma^2(\x)$ is correlated with $\sigma^2(\x + \delta)$
for sufficiently small $\delta$, and similar for $\mu$. Consider the local likelihood
estimation problem \citep{localregression, local-tibshirani} at a point $\x_i$,
\begin{equation}\label{loc-llh-unk}
\log \tilde{p}_\theta(y_i | \x_i) = \sum_{j=1}^N w_j(\x_i)\log p_\theta(y_j | \x_j),
\end{equation}
where $w_j$ is a function that declines as $\| \x_j - \x_i \|$ increases,
implying that the local likelihood at $\x_i$ is dependent on the points nearest
to $\x_i$. Notice $\tilde{p}_\theta(y_i | \x_i)=p_\theta(y_i | \x_i)$ if $w_j(\x_i)=\mathbf{1}_{i=j}$.
Consider, with this $w$, a uniformly drawn subsample (i.e. a standard mini-batch) of the data $\{ \x_k \}_{k=1}^M$ and its corresponding
stochastic gradient of Eq.~\ref{loc-llh-unk} with respect to $\theta_{\sigma^2}$. If for a point, $\x_i$, no points near it are in the subsample, then no other point will influence the gradient of $\sigma^2(\x_i)$, which will point in the direction of the MLE, that is highly uninformative as it does not exist unless $\mu(\x_i)$ is known.
Local data scarcity, thus, implies that while we have sufficient data for fitting
a \emph{mean}, locally we have insufficient data for fitting a \emph{variance}. Essentially, if a point is isolated
in a mini-batch, all information it carries goes to updating $\mu$ and none is present for $\sigma^2$.

If we do not use mini-batches, we encounter that gradients wrt.\ $\theta_\mu$ and $\theta_{\sigma^2}$
will both be scaled with $\frac{1}{2\sigma^2(\x)}$ meaning that points with small variances
effectively have higher learning rates \citep{Nix1994estimating}. This implies a bias towards low-noise regions of data.

%
%


\subsection{Horvitz-Thompson adjusted stochastic gradients}
We will now consider a solution to this problem within the 
local likelihood framework, which will give us a reliable, but biased, stochastic
gradient for the usual (nonlocal) log-likelihood. We will then show how this can be turned into
an unbiased estimator.

If we are to add some local information, giving more reliable gradients, we should choose a $w$ in Eq.\ref{loc-llh-unk} that reflects this.
Assume for simplicity that $w_j(\x_i)=\textbf{1}_{\| \x_i - \x_j\| < d}$ for some
$d > 0$.
The gradient of $\log\tilde{p}_{\theta}(y | \x_i)$ will then be informative,
as more than one observation will contribute to the local variance if $d$ is chosen appropriately. Accordingly, we suggest
a practical mini-batching algorithm that samples a random point $\x_j$ and we let
the mini-batch consist of the $k$ nearest neighbors of $\x_j$.\footnote{By convention,
  we say that the nearest neighbor of a point is the point itself.} In order to
allow for more variability in a mini-batch, we suggest sampling $m$ points
uniformly, and then sampling $n$ points among the $k$ nearest neighbors of each of the $m$ initially sampled points. Note that this 
is a more informative sample, as all observations in the sample are likely to influence the same subset 
of parameters in $\theta$, effectively increasing the degrees of freedom\footnote{Degrees of freedom here refers to the parameters in a Gamma distribution -- the distribution of variance estimators under Gaussian likelihood. Degrees of freedom in general is a quite elusive quantity in regression problems.}, hence the quality of variance estimation. In other words, if the variance network is sufficiently expressive, our Monte Carlo gradients under this sampling scheme are of smaller variation and more sparse.
In the supplementary material, we empirically show that this estimator yields significantly more sparse gradients, which results in improved convergence.
  Pseudo-code of this sampling-scheme, can be found in the supplementary material.

While such a mini-batch would give rise to an informative stochastic gradient,
it would not be an unbiased stochastic gradient of the (nonlocal) log-likelihood.
This can, however, be adjusted by using the \emph{Horvitz-Thompson (HT)} algorithm \citep{horwitz},
i.e.\ rescaling the log-likelihood contribution of each sample $\x_j$ by its inclusion
probability $\pi_j$. With this, an unbiased estimate of the log-likelihood (up to
an additive constant) becomes
\begin{align}
  \sum_{i=1}^{N} \left\{ -\frac{1}{2}\log(\sigma^2(\x_i)) - \frac{(y_i - \mu(\x_i))^2}{2\sigma^2(\x_i)} \right\}
  \approx \sum_{\x_j \in \mathcal{O}} \frac{1}{\pi_j} \left\{ -\frac{1}{2}\log(\sigma^2(\x_j)) - \frac{(y_j-\mu(\x_j))^2}{2\sigma^2(\x_j)} \right\}
\end{align}
where $\mathcal{O}$ denotes the mini-batch. With the nearest neighbor mini-batching,
the inclusion probabilities can be calculated as follows.
The probability that observation $j$ is in the sample is $\nicefrac{n}{k}$ if
it is among the $k$ nearest neighbors of one of the initial $m$ points, which are
chosen with probability $m/N$, i.e. 
\begin{align}
  \pi_j = \frac{m}{N}\sum_{i=1}^N \frac{n}{k}\ \mathbf{1}_{j \in \mathcal{O}_k(i)}, 
\end{align}
where $\mathcal{O}_k(i)$ denotes the $k$ nearest neighbors of $\x_i$.

\paragraph*{Computational costs} The proposed sampling scheme requires an upfront computational cost of $O(N^2 D)$ before any training can begin. We stress that this is pre-training computation and not updated during training. The cost is therefore relative small, compared to training a neural network for small to medium size datasets. Additionally, we note that the search algorithm does not have to be precise, and we could therefore take advantage of fast approximate nearest neighbor algorithms \citep{fu2016}.

\subsection{Mean-variance split training}
The most common training strategy is to first optimize $\theta_{\mu}$ assuming
a constant $\sigma^2$, and then proceed to optimize $\theta = \{\theta_{\mu}, \theta_{\sigma^2} \}$
jointly, i.e.\ a \emph{warm-up} of $\mu$.
As previously noted, the MLE of $\sigma^2$ does not exist when only a single
observation is available and $\mu$ is unknown. However, the MLE \emph{does}
exist when $\mu$ is known, in which case it is $\hat{\sigma}^2(\x_i) = (y_i - \mu(\x_i))^2$,
assuming that the continuity of $\sigma^2$ is not crucial.
This observation suggests that the usual training strategy is substandard as
$\sigma^2$ is never optimized assuming $\mu$ is known.
This is easily solved: we suggest to never updating $\mu$ and $\sigma^2$ simultaneously,
i.e.\ only optimize $\mu$ conditioned on $\sigma^2$, and vice versa. This reads as sequentially optimizing $p_\theta(y|\theta_\mu)$ and $p_\theta(y|\theta_{\sigma^2})$, as we under these conditional distributions we may think of $\mu$ and $\sigma^2$ as known, respectively.
We will refer to this as \emph{mean-variance split training (MV)}.

\subsection{Estimating distributions of variance}
When $\sigma^2(\x_i)$ is influenced by few observations, underestimation is still likely due to the left skewness of the gamma distribution of $\hat{\sigma}_i^2 = (y_i - \mu(\x_i))^2$. As always, when in a low data regime, it is sensible to be Bayesian about it; hence instead of point estimating $\hat{\sigma}_i^2$ we seek to find a distribution. Note that we are not imposing a prior, we are training the parameters of a Bayesian model. We choose the inverse-Gamma distribution, as this is the conjugate prior of $\sigma^2$ when data is Gaussian. This means $\theta_{\sigma^2}=\{\theta_\alpha,\theta_\beta\}$ where $\alpha,\beta>0$ are the shape and scale parameters of the inverse-Gamma respectively. So the log-likelihood is now calculated by integrating out $\sigma^2$
\begin{align}
\log p_\theta(y_i)=\log \int \mathcal{N}(y_i|\mu_i,\sigma_i^2) \mathrm{d} \sigma_i^2 = \log t_{\mu_i,\alpha_i,\beta_i}(y_i),
\end{align}
where $\sigma_i^2\sim \textsc{Inv-Gamma}(\alpha_i,\beta_i)$ and $\alpha_i = \alpha(\x_i), \beta_i = \beta(\x_i)$ are modeled as neural networks. Having an inverse-Gamma prior changes the predictive distribution to a located-scaled\footnote{This means $y\sim F$, where $F= \mu + \sigma t(\nu)$. The explicit density can be found in the \SUPMAT.} Student-\textit{t} distribution, parametrized with $\mu,\alpha$ and $\beta$. Further, the \textit{t}-distribution is often used as a replacement of the Gaussian when data is scarce and the true variance is unknown and yields a \emph{robust} regression \citep{bda3,robust-t}. We let $\alpha$ and $\beta$ be neural networks that implicitly determine the degrees of freedom and the scaling of the distribution. Recall the higher the degrees of freedom, the better the Gaussian approximation of the $t$-distribution.
\subsection{Extrapolation architecture}
If we evaluate the local log-likelihood (Eq.~\ref{loc-llh-unk}) at a point $\x_0$ far away from all
data points, then the weights $w_i(\x_0)$ will all be near (or exactly) zero.
Consequently, the local log-likelihood is approximately 0 regardless of the
observed value $y(\x_0)$, which should be interpreted as a large entropy of $y(\x_0)$.
Since we are working with Gaussian and \textit{t}-distributed variables, we can recreate this behavior by exploiting the fact
that entropy is only an increasing function of the variance.
We can re-enact this behavior by letting the variance tend towards an \emph{a priori}
determined value $\eta$  if $\x_0$ tends away from the training data.
Let $\{ \vec{c}_i \}_{i=1}^L$ be points in $\mathbb{R}^D$ that represent the
training data, akin to inducing points in sparse GPs \citep{snelson2006sparse}.
Then define $\delta(\x_0)=\min_{i}\|\vec{c}_i-\x_0\|$ and
\begin{align}
  \label{eq:arch}
  \hat{\sigma}^2(\x_0) = \big(1-\nu(\delta(\x_0))\big)\hat{\sigma}^2_\theta + \eta\nu(\delta(\x_0)),
\end{align}
where $\nu:[0,\infty) \mapsto [0,1]$ is a surjectively increasing function. Then the variance estimate will go to $\eta$ as $\delta\rightarrow\infty$
at a rate determined by $\nu$. In practice, we choose $\nu$ to be a scaled-and-translated sigmoid function: $\nu(x) = \text{sigmoid}((x+a)/\gamma)$, where $\gamma$ is a free parameter we optimize during training and $a\approx-6.9077\gamma$ to ensure that $\nu(0) \approx 0$. The inducing points $\vec{c}_i$ are initialized with $k$-means and optimized during training.
This choice of architecture is similar to that attained by posterior Gaussian processes
when the associated covariance function is stationary. It is indeed the behavior
of these established models that we aim to mimic with Eq.~\ref{eq:arch}.

%% file: experiments.tex
\subsection{Regression}
To test our methodologies we conduct multiple experiments in various settings. We compare our method to state-of-the-art methods for quantifying uncertainty: Bayesian neural network (BNN) \citep{Hernandez2015backprob}, Monte Carlo Dropout (MC-Dropout) \citep{Gal2015dropout} and Deep Ensembles (Ens-NN) \citep{Lakshminarayanan2016ensembles}. Additionally we compare to two baseline methods: standard mean-variance neural network (NN) \citep{Nix1994estimating} and GPs (sparse GPs (SGP) when standard GPs are not applicable) \citep{rasmussen:book}. We refer to our own method(s) as \emph{Combined}, since we apply all the methodologies described in Sec.~\ref{sec:meth}. Implementation details and code can be found in the \SUPMAT. 
Strict comparisons of the models should be carefully considered; having two seperate networks to model mean and variance seperately (as NN, Ens-NN and Combined) means that all the predictive uncertainty, \ie both aleatoric and episteminc, is modeled by the variance networks alone. BNN and MC-Dropout have a higher emphasis on modeling epistemic uncertainty, while GPs have the cleanest separation of noise and model uncertainty estimation.
Despite the methods quantifying different types of uncertainty, their results can still be ranked by test set log-likelihood, which is a proper scoring function.

\begin{figure}[h!]
\centering
\begin{minipage}{0.6\textwidth} \hspace{-0.5cm}
  \includegraphics[height=4cm]{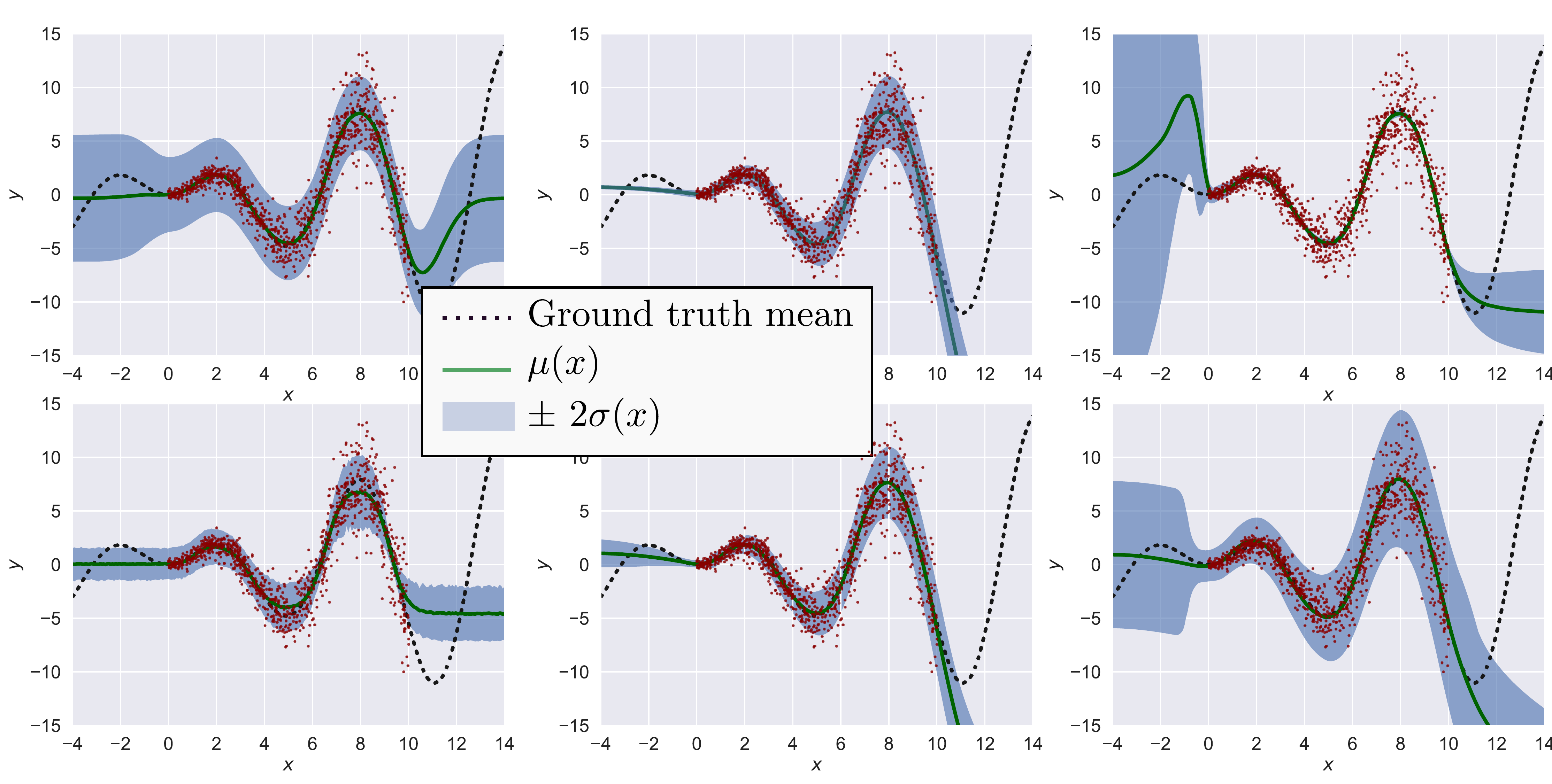}
\captionof{figure}{From top left to bottom right: GP, NN, \\BNN, MC-Dropout, Ens-NN, Combined.}
\label{fig:test1}
\end{minipage} \hspace{-0.5cm}
\begin{minipage}{0.40\textwidth} \vspace{0.4cm}
\vspace{-3mm}
\subfloat{\includegraphics[height=4.0cm]{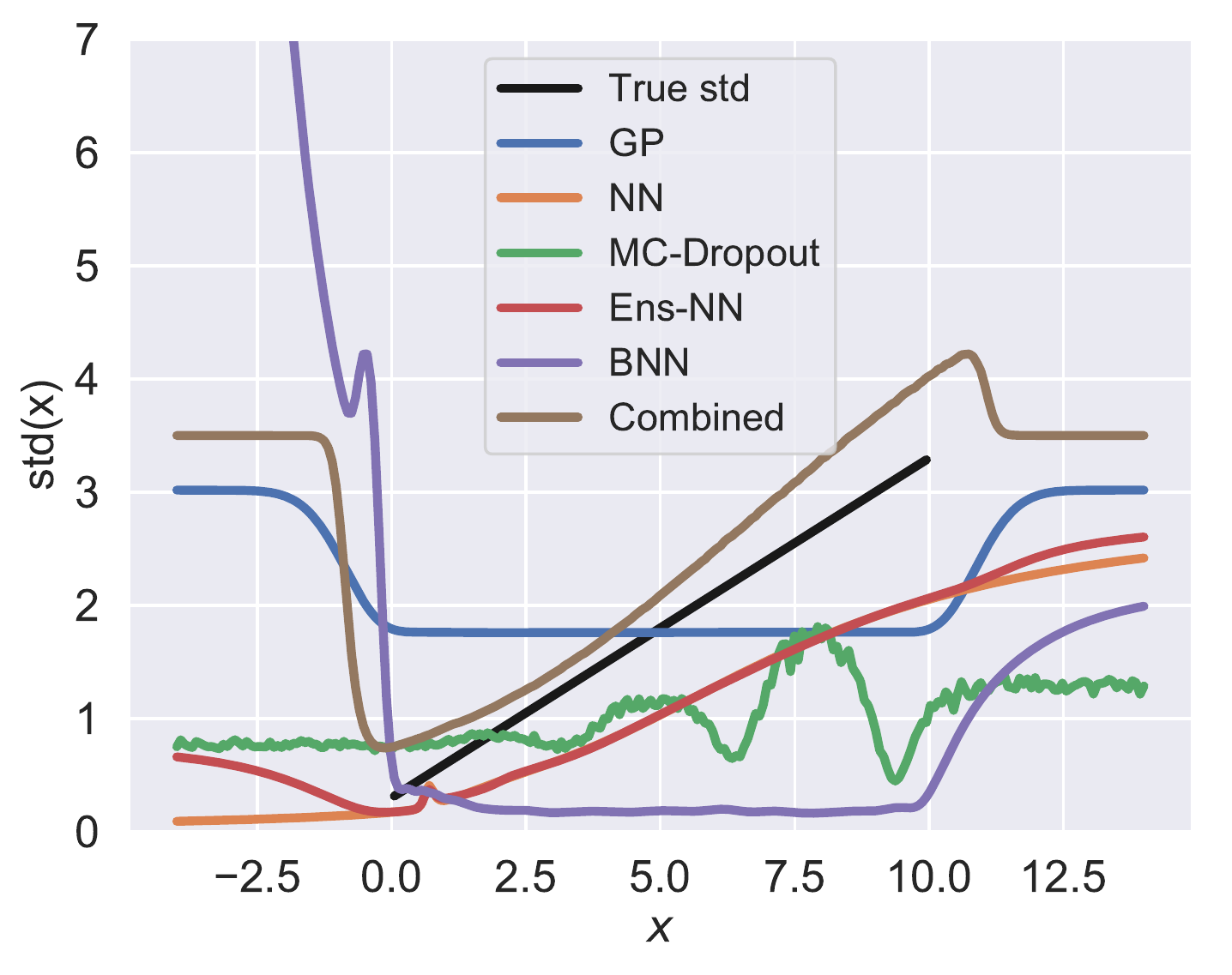}}
\vspace{-1mm}
\captionof{figure}{Standard deviation estimates as a function of $\x$.}
\label{fig:test2}
\end{minipage}
\end{figure}

\paragraph{Toy regression.}
We first return to the toy problem of Sec.~\ref{sec:intro}, where we consider 500 points from
$y=\x\cdot \sin(\x) + 0.3 \cdot \epsilon_{1} + 0.3 \cdot \x \cdot \epsilon_{2}$, with $\epsilon_{1}, \epsilon_{2} \sim \Ncal(0,1)$.
In this example, the variance is \emph{heteroscedastic}, and models should estimate larger variance for larger values of $\x$.
The results\footnote{The standard deviation plotted for \emph{Combined}, is the root mean of the inverse-Gamma.} can be seen in \FIGS~\ref{fig:test1} and \ref{fig:test2}.
Our approach is the only one to satisfy all of the following: capture the heteroscedasticity, extrapolate high variance outside data region and not underestimating within.


\begin{figure}[h!]
\subfloat[GP]{\includegraphics[width=0.17\textwidth, trim=2cm 2cm 0cm 0cm, clip]{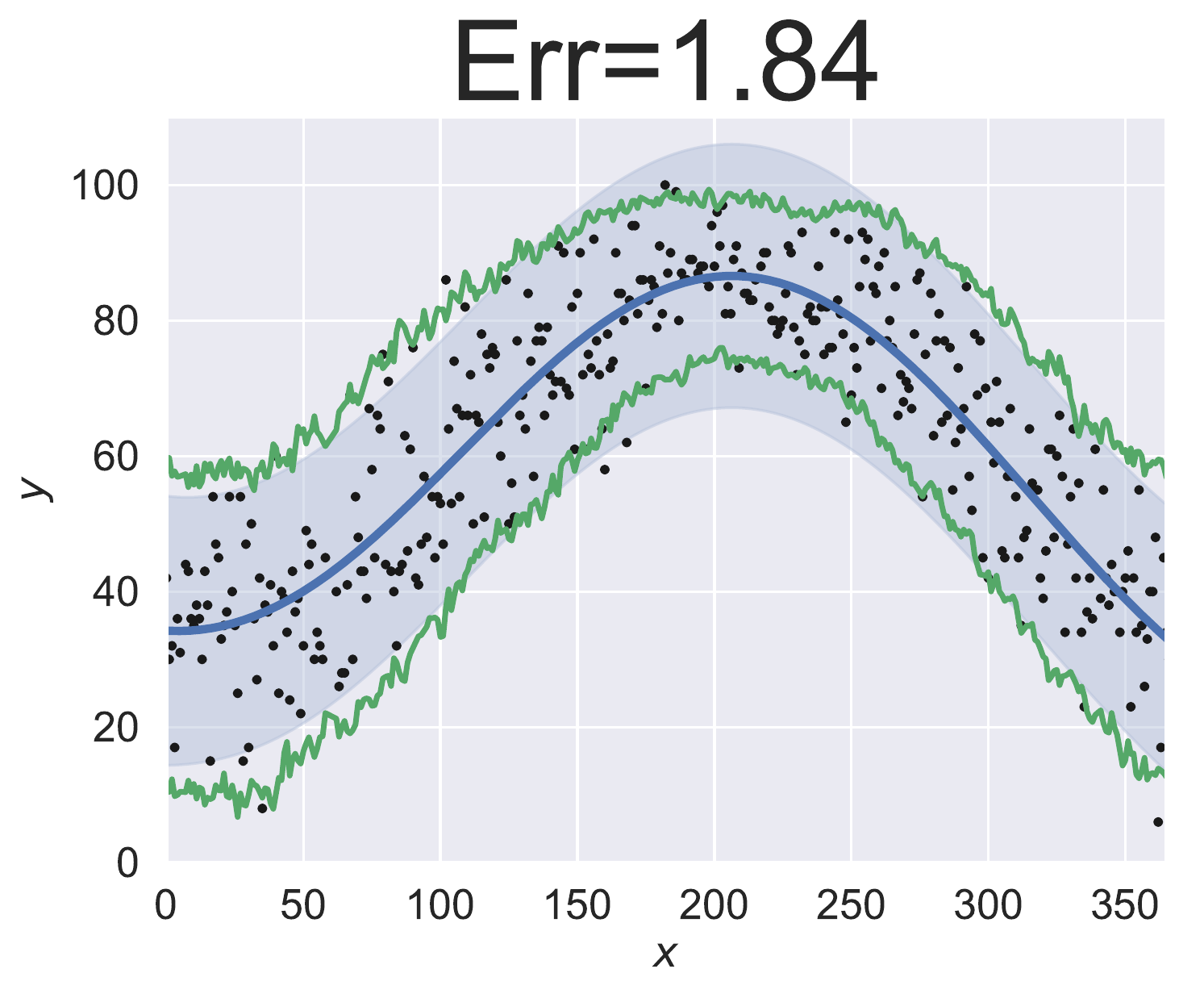}} 
\subfloat[NN]{\includegraphics[width=0.17\textwidth, trim=2cm 2cm 0cm 0cm, clip]{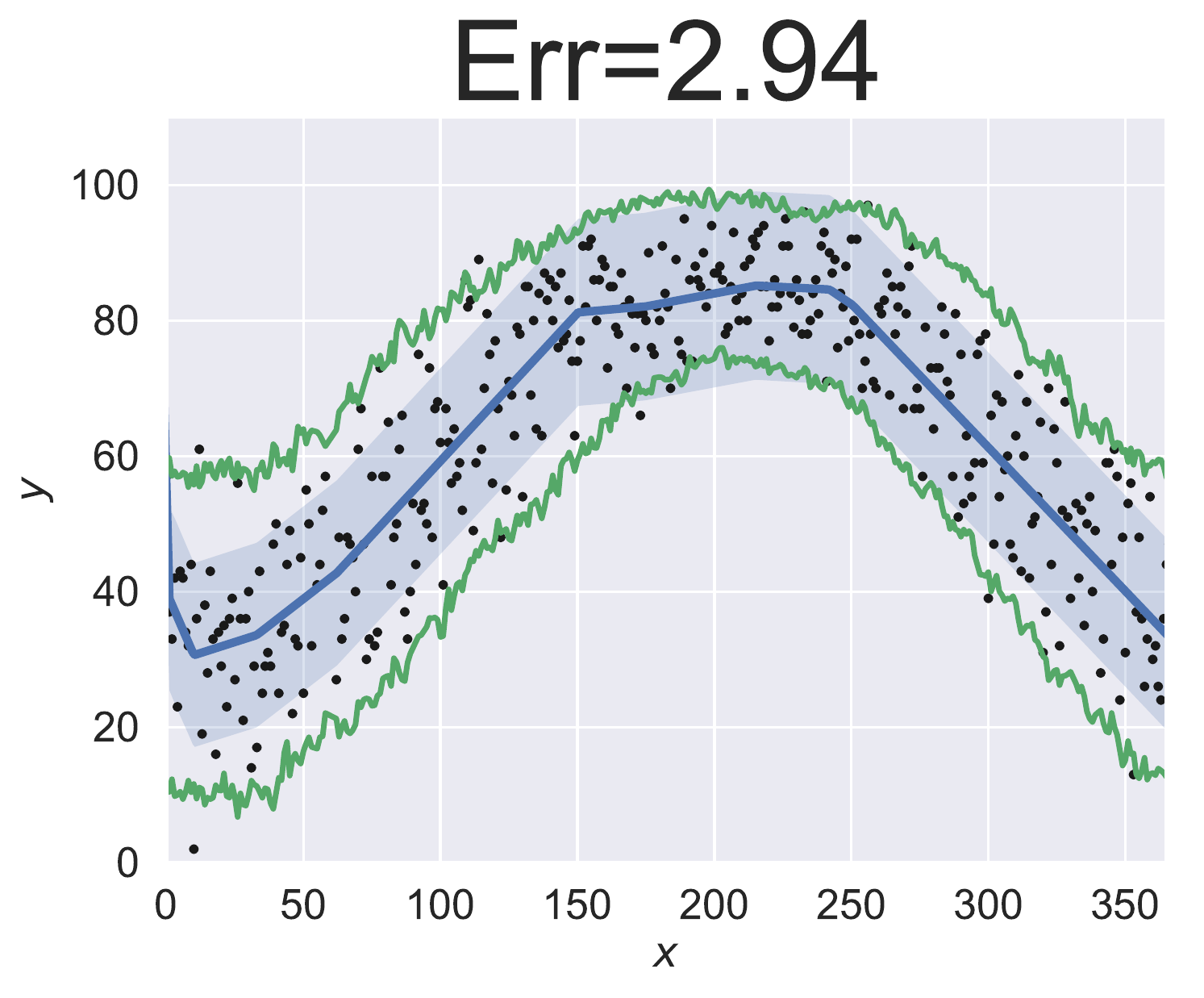}}
\subfloat[BNN]{\includegraphics[width=0.17\textwidth, trim=2cm 2cm 0cm 0cm, clip]{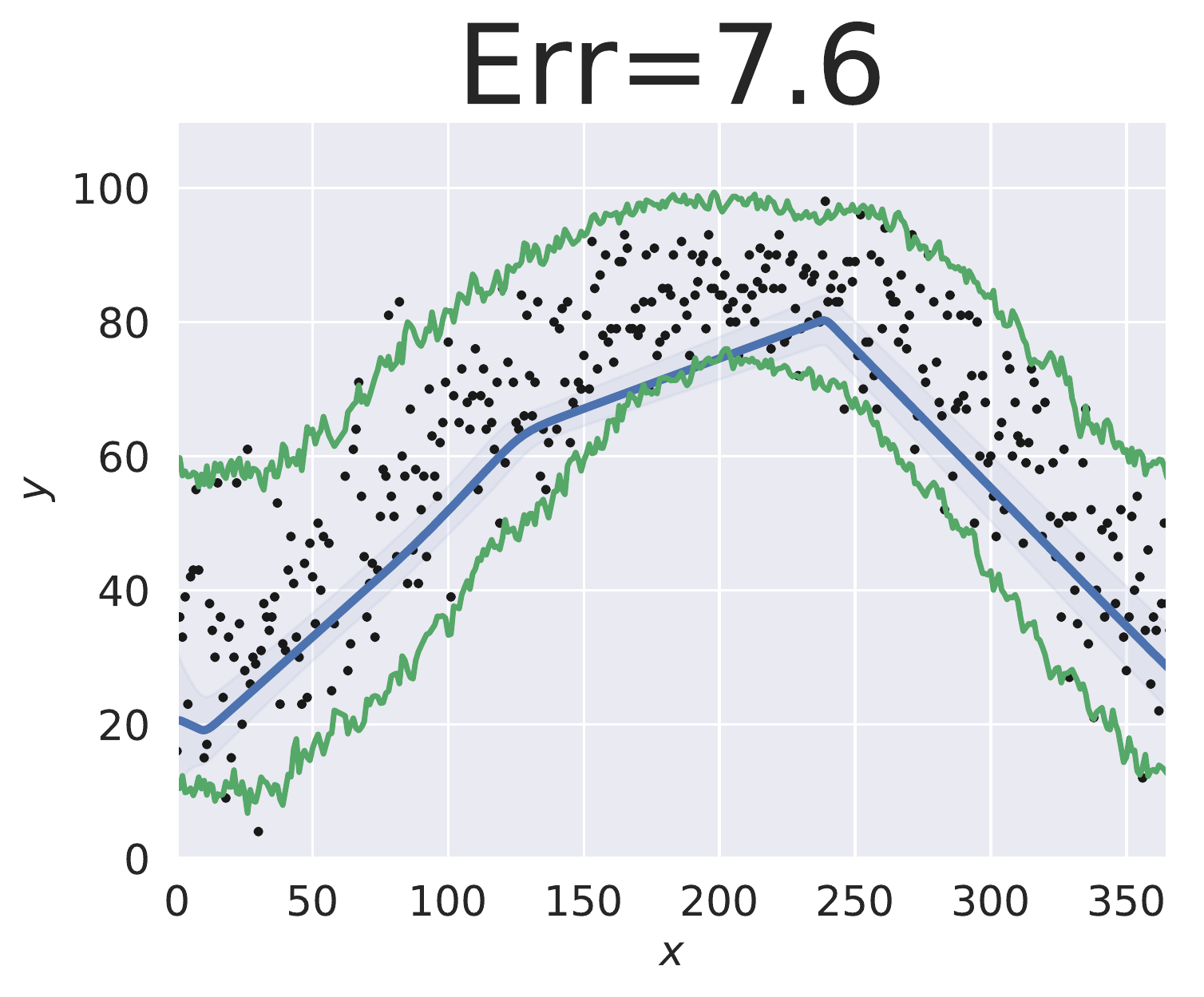}}
\subfloat[MC-Dropout]{\includegraphics[width=0.17\textwidth, trim=2cm 2cm 0cm 0cm, clip]{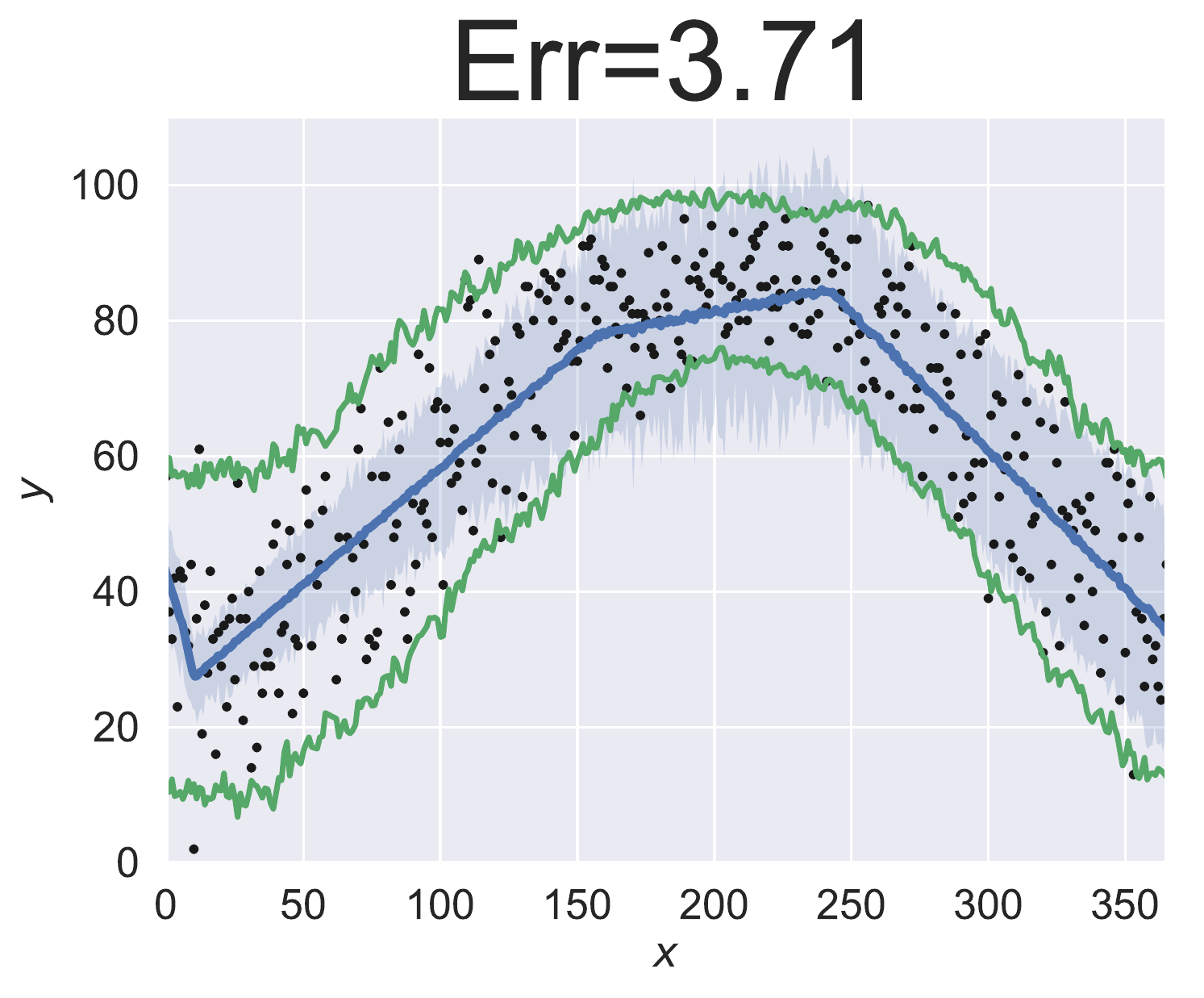}}
\subfloat[Ens-NN]{\includegraphics[width=0.17\textwidth, trim=2cm 2cm 0cm 0cm, clip]{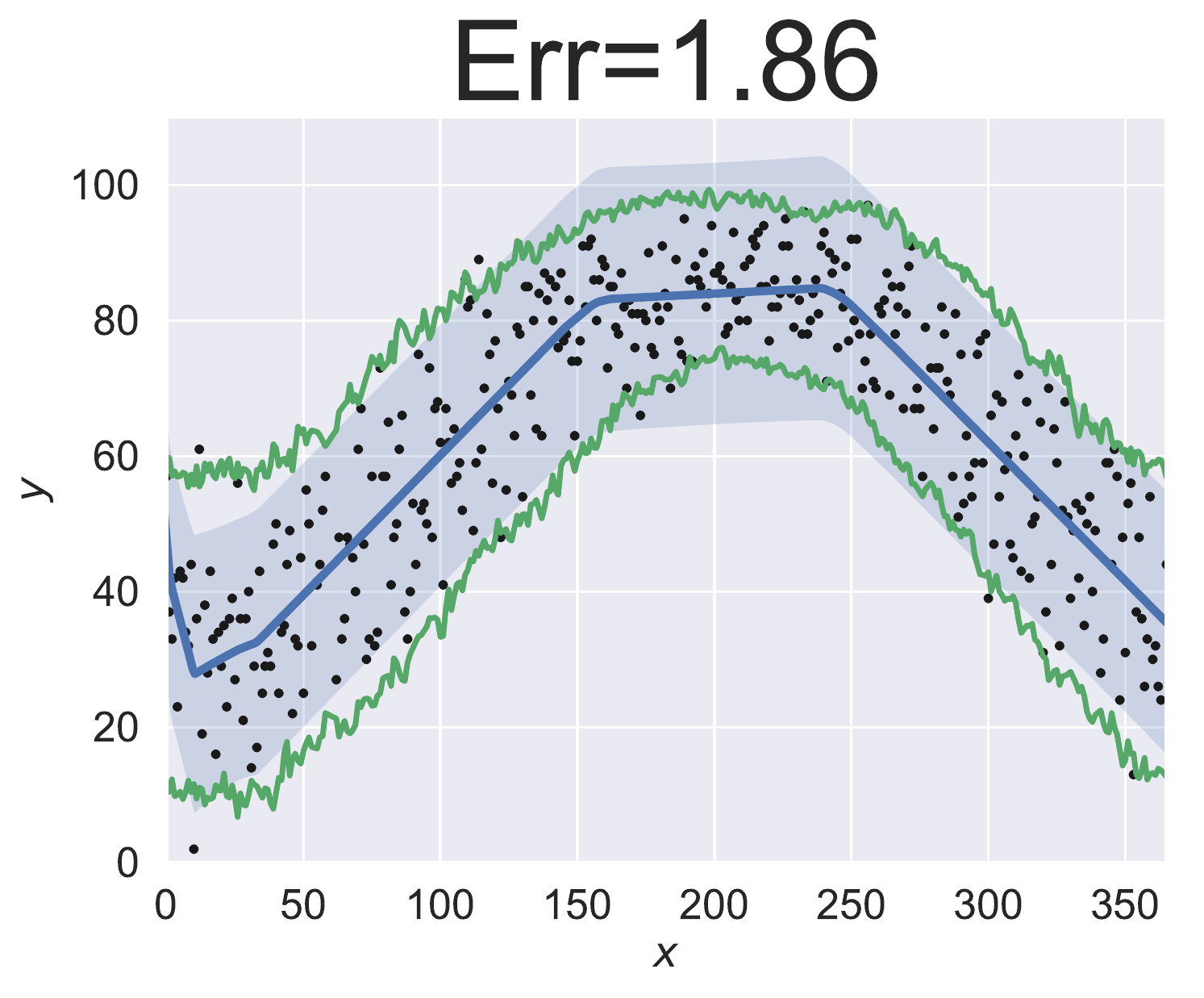}} 
\subfloat[Combined]{\includegraphics[width=0.17\textwidth, trim=2cm 2cm 0cm 0cm, clip]{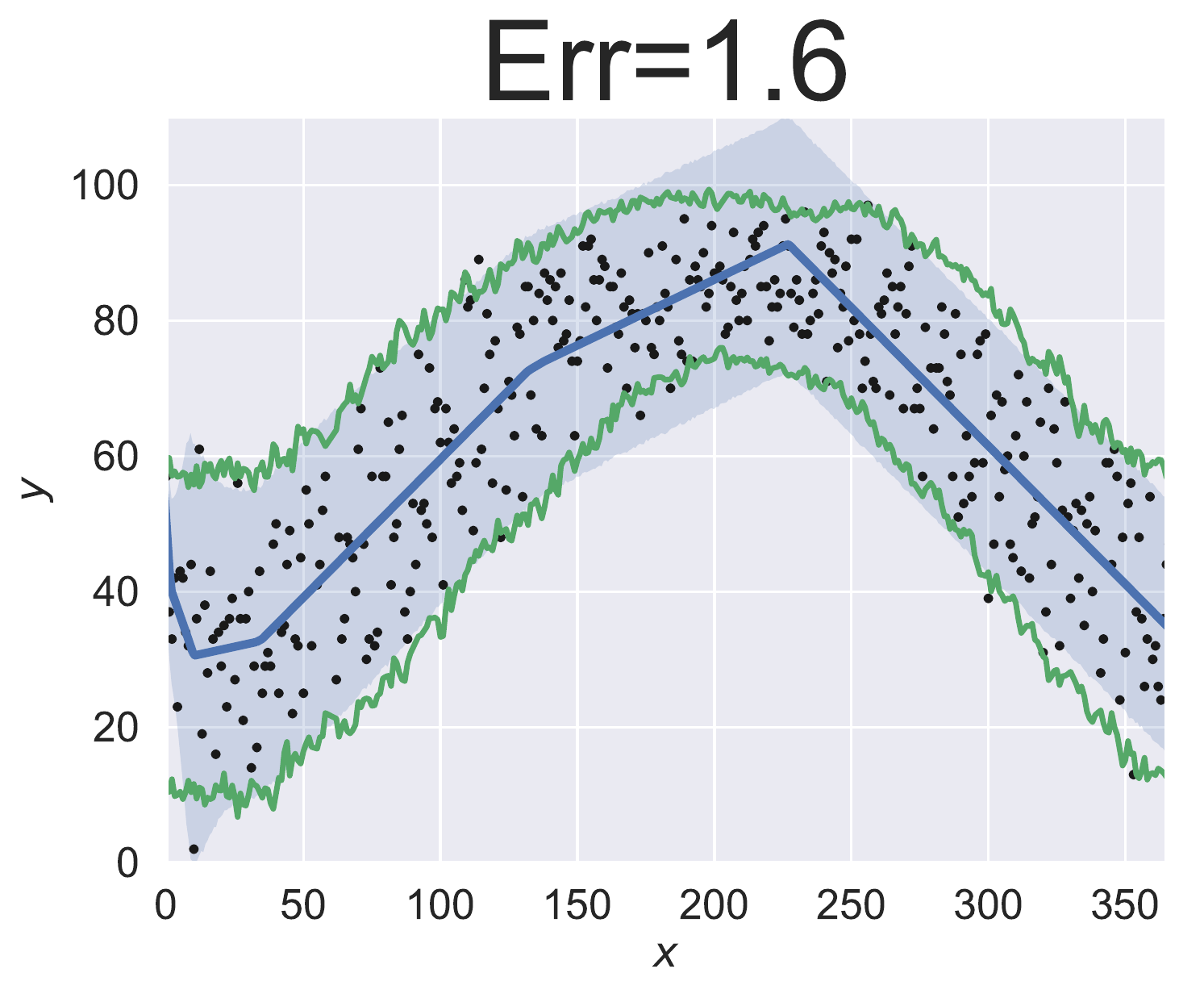}}
\caption{Weather data with uncertainties. Dots are datapoints, green lines are the true uncertainty, blue curves are mean predictions and the blue shaded areas are the estimated uncertainties.}
\label{fig:weather}
\end{figure}

\paragraph{Variance calibration.}
To our knowledge, no benchmark for quantifying variance estimation exists. We propose a simple dataset with known uncertainty information. More precisely, we consider weather data from over 130 years.\footnote{\url{https://mrcc.illinois.edu/CLIMATE/Station/Daily/StnDyBTD2.jsp}} Each day the maximum temperature is measured, and the uncertainty is then given as the variance in temperature over the 130 years. The fitted models can be seen in \FIG\ref{fig:weather}. Here we measure performance by calculating the mean error in uncertainty: $\text{Err} = \frac{1}{N} \sum_{i=1}^N |\sigma^2_\text{true}(\x_i) - \sigma^2_\text{est}(\x_i)|$. The numbers are reported above each fit. We observe that our Combined model achieves the lowest error of all the models, closely followed by Ens-NN and GP. Both NN, BNN and MC-Dropout all severely underestimate the uncertainty. 

\begin{figure}[h!]
	\subfloat{\includegraphics[width=0.325\textwidth, trim=0cm 0cm 0cm 0cm, clip]{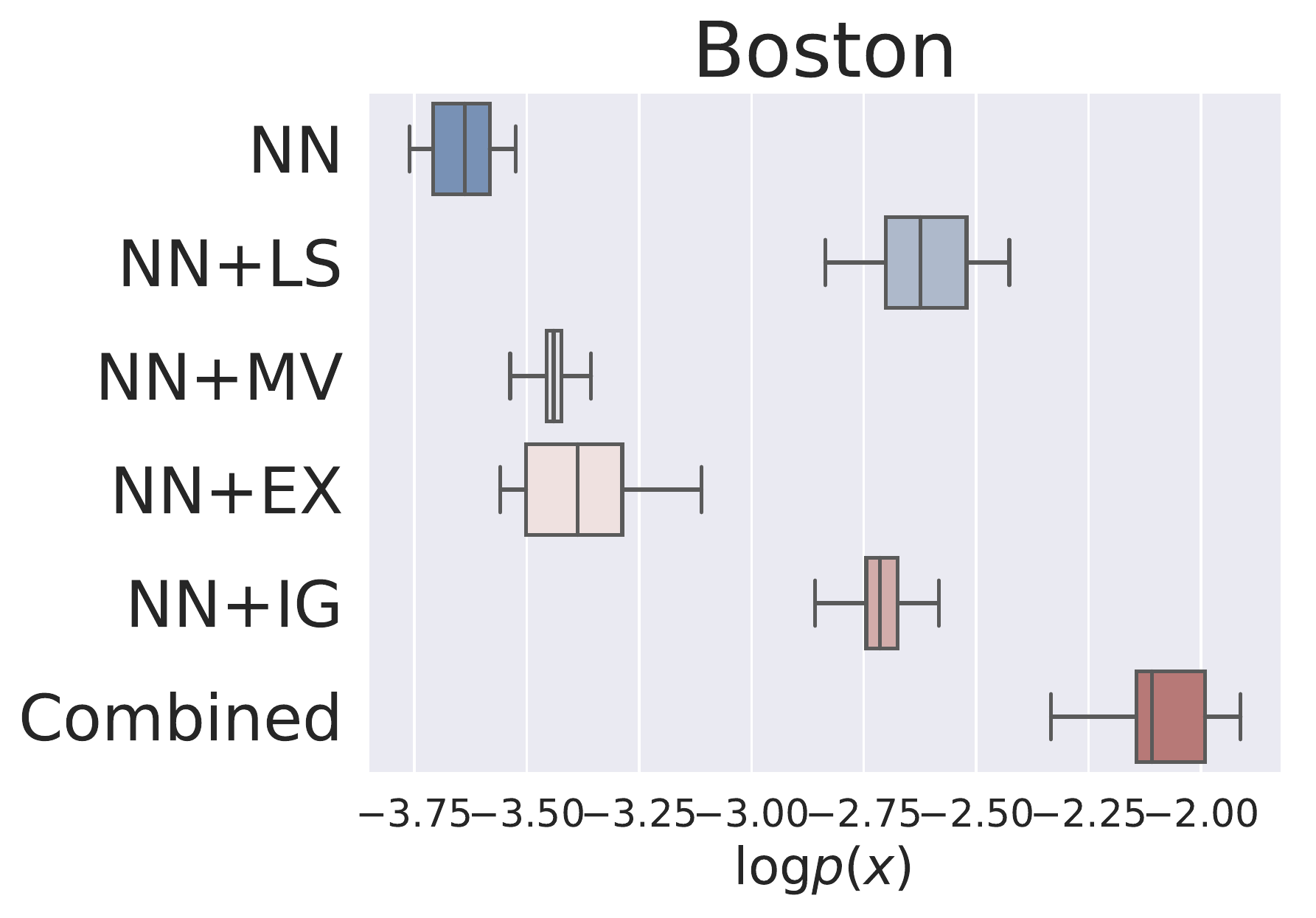}}
	\subfloat{\includegraphics[width=0.23\textwidth, trim=5.5cm 0cm 0cm 0cm, clip]{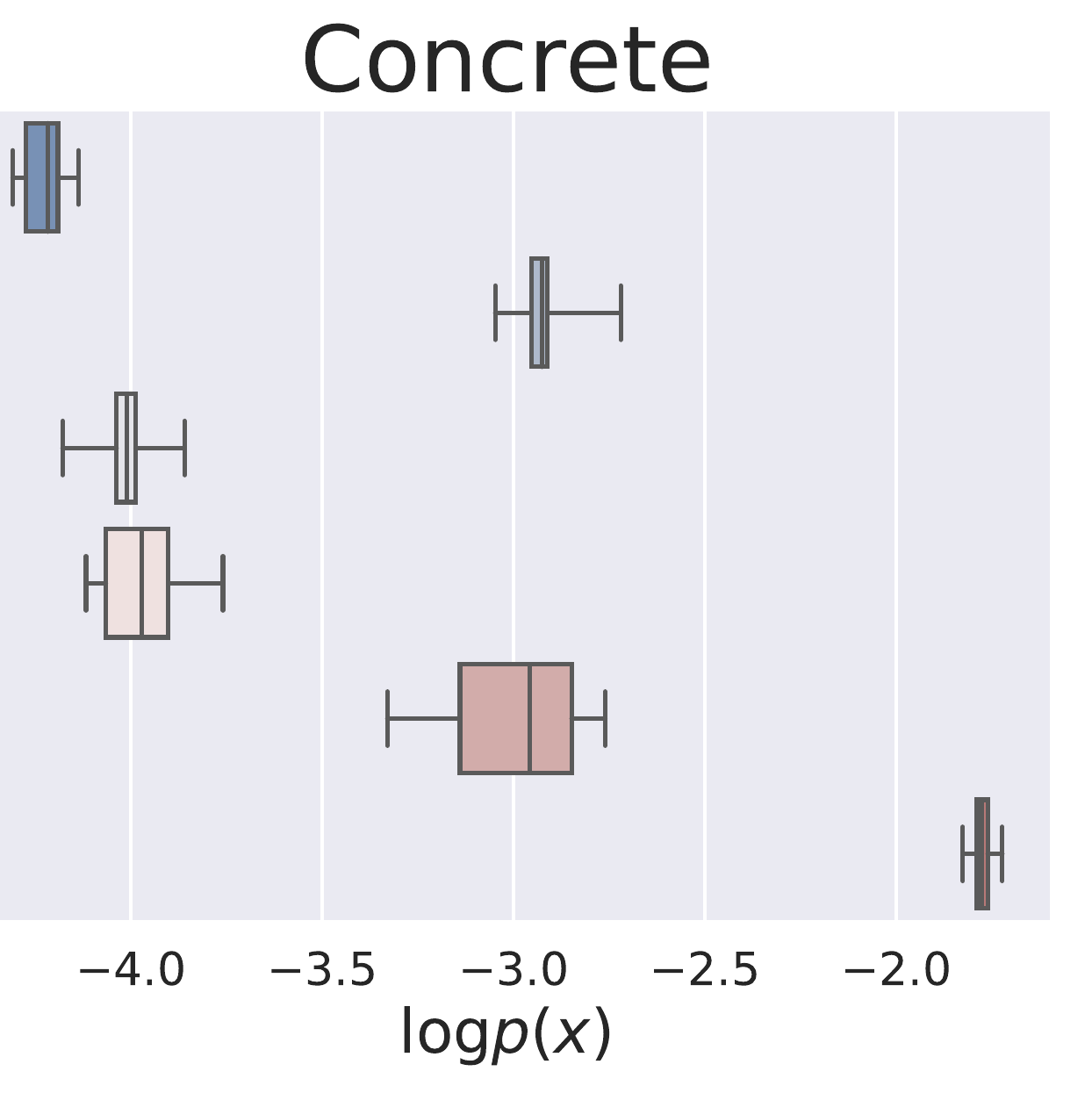}}
	\subfloat{\includegraphics[width=0.23\textwidth, trim=5.5cm 0cm 0cm 0cm, clip]{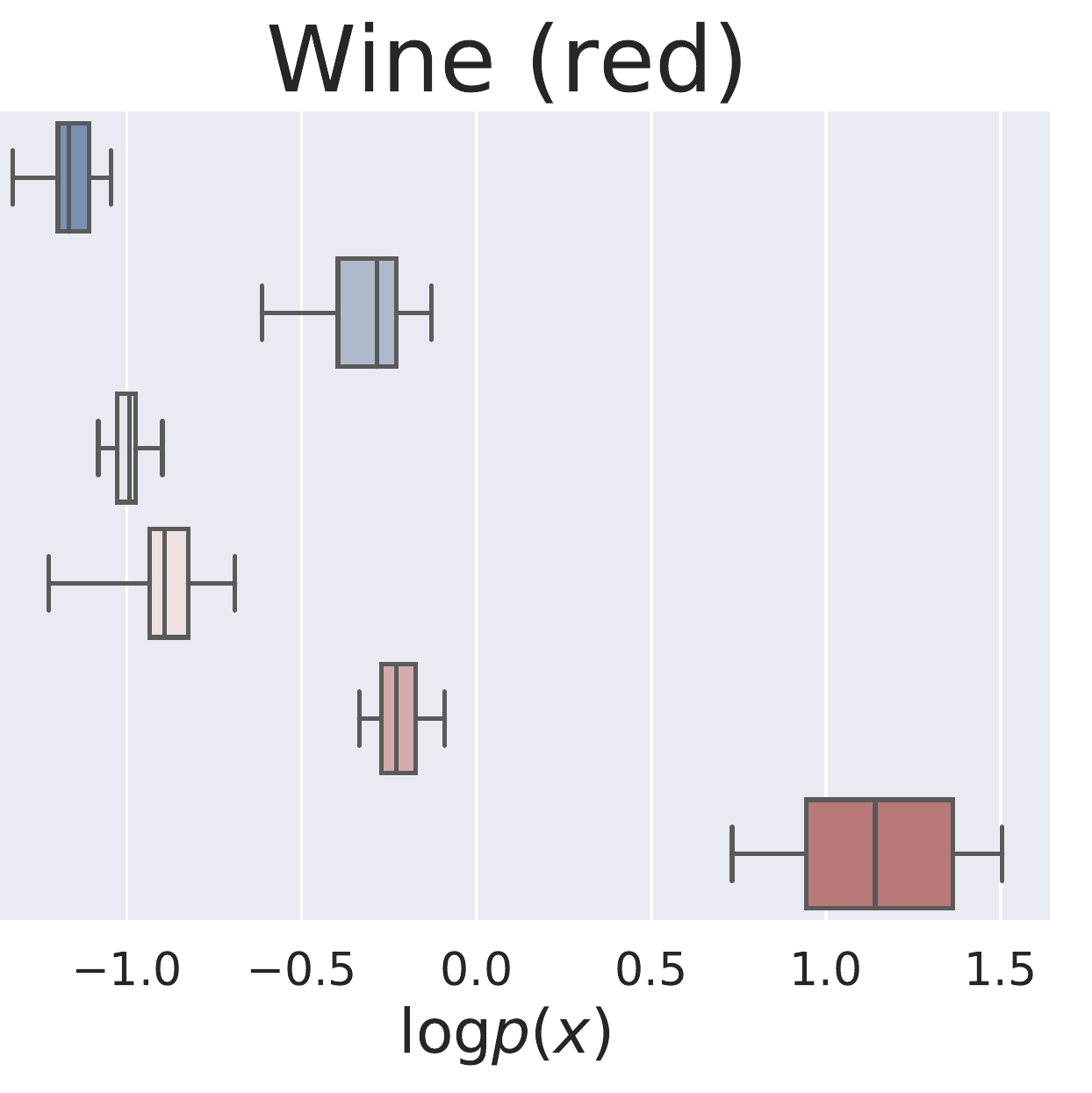}}
	\subfloat{\includegraphics[width=0.23\textwidth, trim=5.5cm 0cm 0cm 0cm, clip]{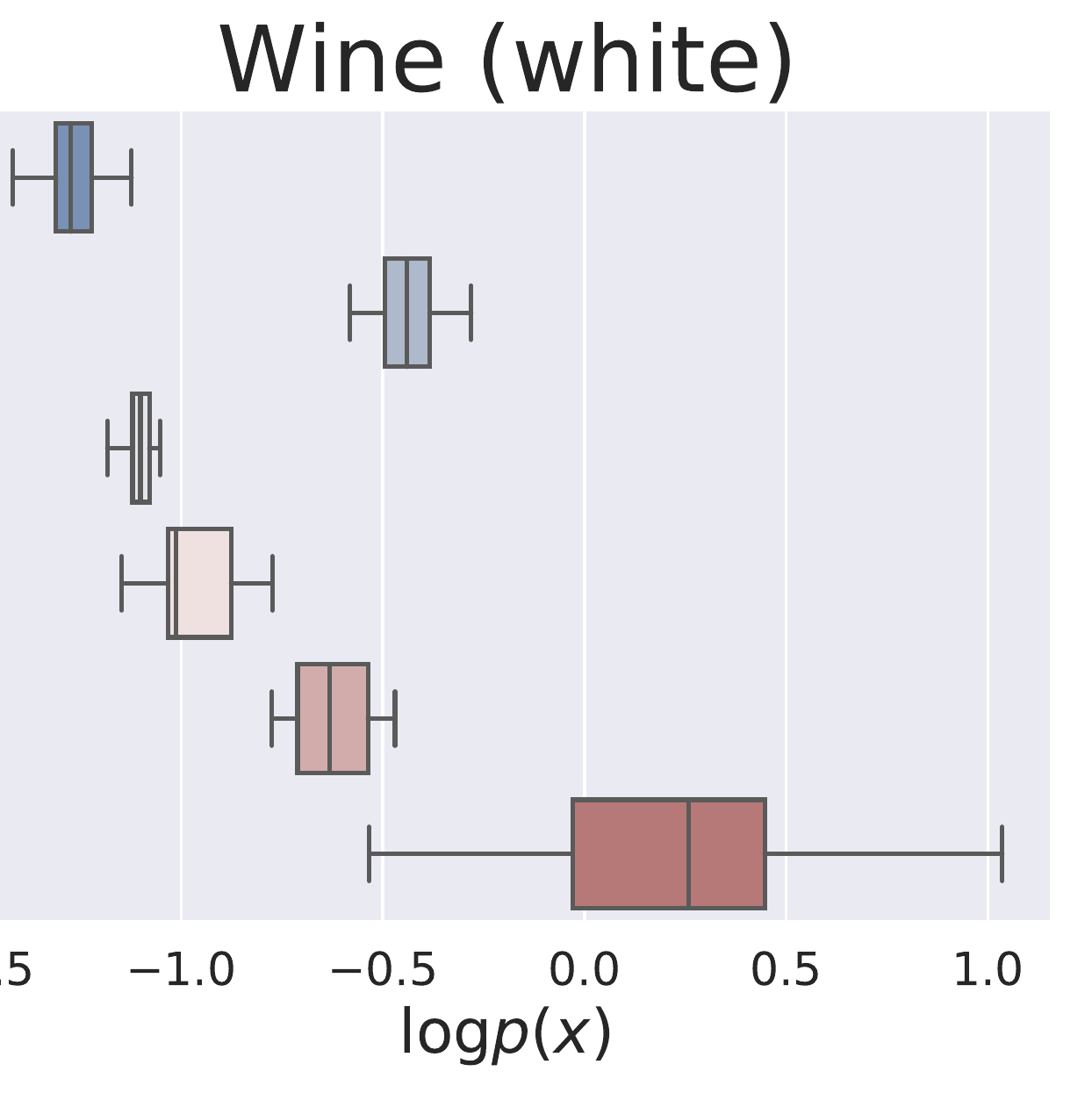}}
	\caption{The complementary methodologies from Sec.~\ref{sec:meth} evaluated on UCI benchmark datasets.}
	\label{fig:contrib}
\end{figure}

\paragraph{Ablation study.}
To determine the influence of each methodology from Sec.~\ref{sec:meth}, we experimented with four UCI regression datasets (Fig.~\ref{fig:contrib}). We split our contributions in four: the locality sampler (LS), the mean-variance split (MV), the inverse-gamma prior (IG) and the extrapolating architecture (EX). The combined model includes all four tricks. The results clearly shows that LS and IG methodologies has the most impact on test set log likelihood, but none of the methodologies perform worse than the baseline model. Combined they further improves the results, indicating that the proposed methodologies are complementary.

\paragraph{UCI benchmark.}
We now follow the experimental setup from \citet{Hernandez2015backprob}, by evaluating models on a number of regression datasets from the UCI machine learning database. Additional to the standard benchmark, we have added 4 datasets. Test set log-likelihood can be seen in \TAB\ref{tab:uci_benchmark_ll}, and the corresponding RMSE scores can be found in the \SUPMAT. 

\begin{table}[]
	\resizebox{\textwidth}{!}{
	\setlength\tabcolsep{1pt}
	\begin{tabular}{lll|rrrrrrr}
		& $N$    & $D$ & \multicolumn{1}{c}{GP}   & \multicolumn{1}{c}{SGP} & \multicolumn{1}{c}{NN} & \multicolumn{1}{c}{BNN} & \multicolumn{1}{c}{MC-Dropout} & \multicolumn{1}{c}{Ens-NN} & \multicolumn{1}{c}{Combined}   \\
		\rowcolor[HTML]{EFEFEF} 
		Boston       & 506    & 13  & $\mathbf{-1.76\pm 0.3}$  & $-1.85\pm 0.25$         & $-3.64\pm 0.09$        & $-2.59\pm 0.11$         & $-2.51\pm 0.31$                & $-2.45\pm 0.25$             & $-2.09\pm 0.09$         \\
		Carbon       & 10721  & 7   & -                      & $3.74\pm 0.53$          & $-2.03\pm 0.14$       & $-1.1\pm 1.76$          & $-1.08\pm 0.05$                & $-0.44\pm 7.28$            & $\mathbf{4.35\pm 0.16}$  \\
		\rowcolor[HTML]{EFEFEF} 
		Concrete     & 1030   & 8   & $-2.13\pm 0.14$          & $-2.29 \pm 0.12$        & $-4.23\pm 0.07$        & $-3.31\pm 0.05$         & $-3.11\pm 0.12$                & $-3.06\pm 0.32$            & $\mathbf{-1.78\pm 0.04}$ \\
		Energy       & 768    & 8   & $-1.85\pm 0.34$			& $-2.22\pm 0.15$         & $-3.78\pm 0.04$        & $-2.07\pm 0.08$         & $-2.01\pm 0.11$                & $\mathbf{-1.48\pm 0.31}$   & $-1.68\pm 0.13$ \\
		\rowcolor[HTML]{EFEFEF} 
		Kin8nm       & 8192   & 8   & -                      & $2.01\pm 0.02$          & $-0.08\pm 0.02$         & $0.95\pm 0.08$          & $0.95\pm 0.15$                 & $1.18\pm 0.03$             & $\mathbf{2.49\pm 0.07}$  \\
		Naval        & 11934  & 16  & -                      & -                     & $3.47\pm 0.21$         & $3.71 \pm 0.05$         & $3.80\pm 0.09$                 & $5.55\pm 0.05$            & $\mathbf{7.27\pm 0.13}$  \\
		\rowcolor[HTML]{EFEFEF} 
		Power plant  & 9568   & 4   & -                      & $-1.9 \pm 0.03$         & $-4.26\pm 0.14$        & $-2.89\pm 0.01$         & $-2.89\pm 0.14$                & $-2.77\pm 0.04$            & $\mathbf{-1.19\pm 0.03}$ \\
		Protein      & 45730  & 9   & -                      & -                     & $-2.95\pm 0.09$        & $-2.91\pm 0.00$         & $-2.93\pm 0.14$                & $\mathbf{-2.80\pm 0.02}$   & $\mathbf{-2.83\pm 0.05}$          \\
		\rowcolor[HTML]{EFEFEF} 
		Superconduct & 21263  & 81  & -                      & $-4.07\pm 0.01$         & $-4.92\pm 0.10$        & $-3.06\pm 0.14$         & $-2.91\pm 0.19$                & $-3.01\pm 0.05$            & $\mathbf{-2.43\pm 0.05}$ \\
		Wine (red)   & 1599   & 11  & $0.96\pm 0.18$           & $-0.08\pm 0.01$         & $-1.19\pm 0.11$     & $-0.98\pm 0.01$         & $-0.94\pm 0.01$                	& $-0.93\pm 0.09$            	& $\mathbf{1.21\pm 0.23}$  \\
		\rowcolor[HTML]{EFEFEF} 
		Wine (white) & 4898   & 11  & -                      & $-0.14\pm 0.05$         & $-1.29\pm 0.09$        & $-1.41\pm 0.17$         & $-1.26\pm 0.01$                & $-0.99\pm 0.06$            	& $\mathbf{0.40\pm 0.42}$  \\
		Yacht        & 308    & 7   & $\mathbf{0.16\pm 1.22}$  & $-0.38\pm 0.32$         & $-4.12\pm 0.17$        & $-1.65\pm 0.05$         & $-1.55\pm 0.12$                & $-1.18\pm 0.21$             & $\mathbf{-0.07\pm 0.05}$ \\
		\rowcolor[HTML]{EFEFEF} 
		Year         & 515345 & 90  & -                      & -                     & $-5.21\pm 0.87$       & $-3.97\pm 0.34$         & $-3.78\pm 0.01$                & $-3.42\pm 0.02$             & $\mathbf{-3.01\pm 0.14}$
	\end{tabular}}
    \vspace{0.1cm}
	\caption{Dataset characteristics and tests set log-likelihoods for the different methods. A - indicates the model was infeasible to train. Bold highlights the best results.} 
	\label{tab:uci_benchmark_ll}
\end{table}

Our \emph{Combined} model performs best on 10 of the 13 datasets. For the small \emph{Boston} and \emph{Yacht} datasets, the standard GP performs the best, which is in line with the experience that GPs perform well when data is scarce. On these datasets our model is the best-performing neural network. On the \emph{Energy} and \emph{Protein} datasets Ens-NN perform the best, closely followed by our Combined model. One clear advantage of our model compared to Ens-NN is that we only need to train one model, whereas Ens-NN need to train 5+ (see the supplementary material for training times for each model). The worst performing model in all cases is the baseline NN model, which clearly indicates that the usual tools for \emph{mean} estimation does not carry over to \emph{variance} estimation.

\paragraph{Active learning.}
The performance of active learning depends on predictive uncertainty \citep{Settles10activelearning},
so we use this to demonstrate the improvements induced by our method.
%
%
We use the same network architectures and datasets as in the UCI benchmark. Each dataset is split into: 20\% train, 60\% pool and 20\% test. For each active learning iteration, we first train a model, evaluate the performance on the test set and then estimate uncertainty for all datapoints in the pool. We then select the $n$ points with highest variance (corresponding to highest entropy \citep{Houlsby2012}) and add these to the training set. 
We set $n=1\%$ of the initial pool size. This is repeated 10 times, such that the last model is trained on 30\%. We repeat this on 10 random training-test splits to compute standard errors. 

\begin{figure}[h!]
\centering
\includegraphics[width=\textwidth]{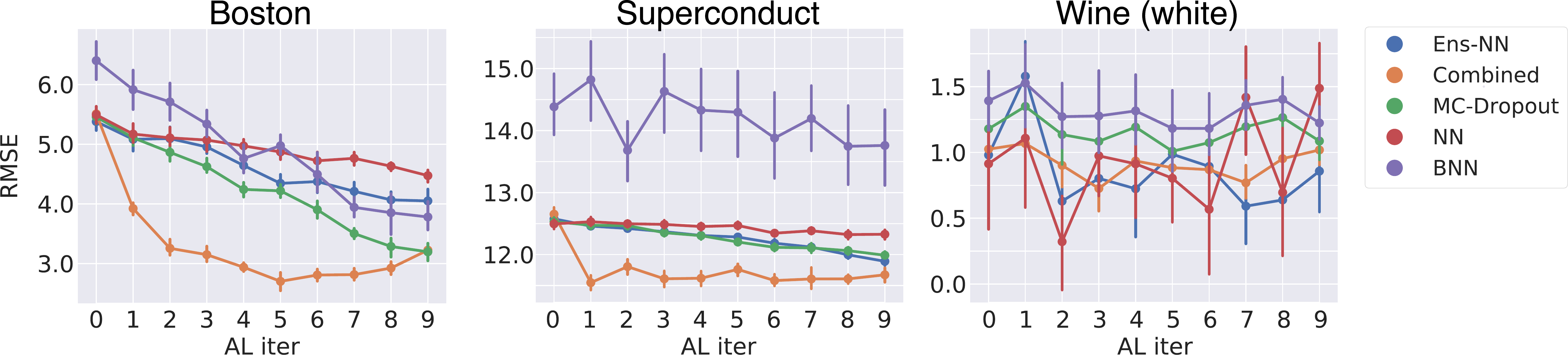}
\caption{Average test set RMSE and standard errors in active learning. The remaining datasets are shown in the supplementary material.}
\label{fig:rmse_active_learning}
\end{figure}

\FIG \ref{fig:rmse_active_learning},show the evolution of average RMSE for each method during the data collection process for the \emph{Boston}, \emph{Superconduct} and \emph{Wine (white)} datasets (all remaining UCI datasets are visualized in the \SUPMAT). In general, we observe two trends. For some datasets we observe that our \emph{Combined} model outperforms all other models, achieving significantly faster learning. This indicates that our model is better at predicting the uncertainty of the data in the pool set. On datasets where the sampling process does not increase performance, we are on par with other models. 

\subsection{Generative models}
To show a broader application of our approach, we also explore it in the context of generative modeling. We focus on variational autoencoders (VAEs) \citep{kingma:iclr:2014, rezende2014stochastic} that are popular deep generative models. A VAE model the generative process:
\begin{equation}
p(\x) \!=\! \int p_\theta(\x | \z) p(\z) \mathrm{d}\z, 
\quad p_\theta(\x | \z) \!=\! \Ncal\big(\x | \mu_\theta(\z) , \sigma_\theta^2 (\z)\big) 
\text{\quad or\quad}  p_\theta(\x | \z) \!=\! \mathcal{B}\big(\x | \mu_\theta(\z)\big),
\label{eq:vae}
\end{equation}
where $p(\z) = \Ncal(\textbf{0}, \mathbb{I}_d)$. This is trained by introducing a variational approximation $q_\phi(\z | \x)=\Ncal(\z | \mu_\phi(\x) , \sigma_\phi^2 (\x))$ and then jointly training $p_\theta$ and $q_\phi$. For our purposes, it suffcient to note that a VAE estimates both a mean and a variance function. Thus using standard training methods, the same problems arise as in the regression setting. \cite{Mattei2018} have recently shown that estimating a VAE is ill-posed unless the variance is bounded from below.
In the literature, we often find that\looseness=-1


\textbf{1.} Variance networks are avoided by using a Bernoulli distribution, even if data is not binary.

\textbf{2.} Optimizing VAEs with a Gaussian posterior is considerably harder than the Bernoulli case. To overcome this, the variance is often set to a constant \eg $\sigma^2(\z)=1$. The consequence is that the log-likelihood reconstruction term in the ELBO collapses into an L2 reconstruction term. 

\textbf{3.} Even though the generative process is given by \EQN\ref{eq:vae}, samples shown in the literature are often reduced to $\tilde{\x} = \mu(\z), \z \sim \Ncal(\textbf{0}, \mathbb{I})$. This is probably due to the wrong/meaningless variance term. 

We aim to fix this by training the posterior variance $\sigma_\theta^2(\z)$ with our Combined method. We do not change the encoder variance $\sigma_\phi^2(\x)$ and leave this to future study. 

\begin{figure}
\centering
\begin{minipage}{0.72\textwidth}\centering
\vspace{-6mm}
\captionsetup[subfigure]{labelformat=empty}
\subfloat{\includegraphics[width=0.33\textwidth, trim=0.77in 5.4in 7.27in 2.53in, clip]{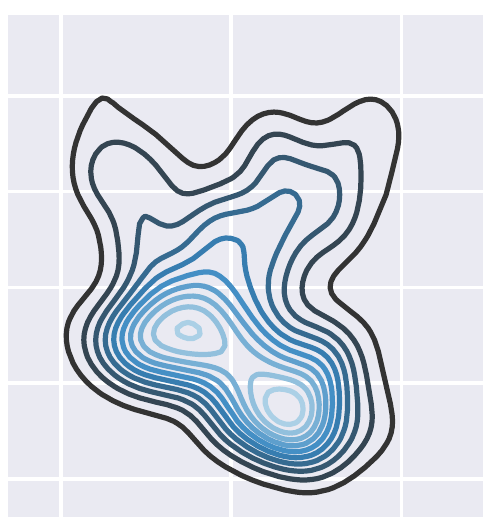}}
\subfloat{\includegraphics[width=0.33\textwidth, trim=0.77in 5.4in 7.27in 2.53in, clip]{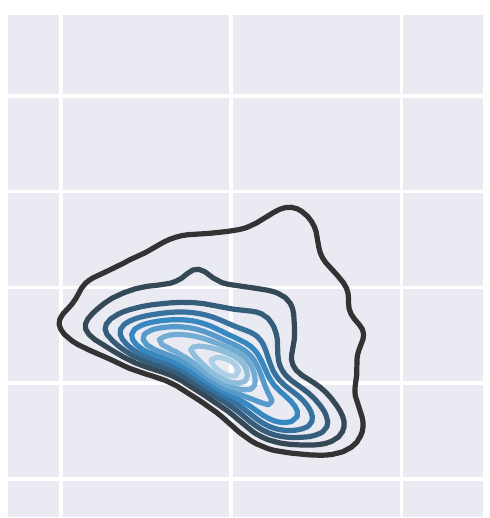}}
\subfloat{\includegraphics[width=0.33\textwidth, trim=0.77in 5.4in 7.27in 2.53in, clip]{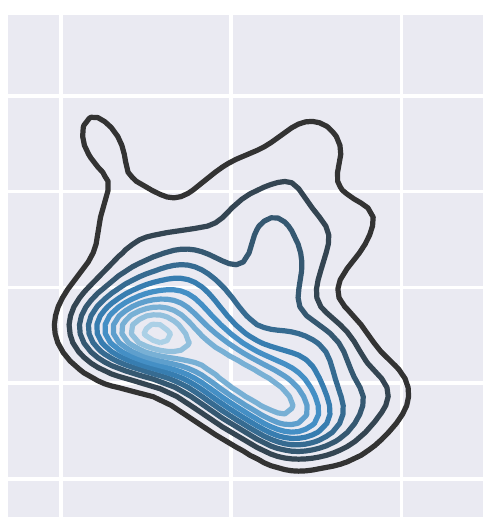}} \hspace{-1mm}
\subfloat[Ground truth]{\includegraphics[width=0.33\textwidth, trim=3.06in 0.7in 4.95in 7.23in, clip]{ground_true.pdf}}
\subfloat[VAE]{\includegraphics[width=0.33\textwidth, trim=3.06in 0.7in 4.95in 7.23in, clip]{vae.pdf}}
\subfloat[Comb-VAE]{\includegraphics[width=0.33\textwidth, trim=3.06in 0.7in 4.95in 7.23in, clip]{comb_vae.pdf}}
\caption{The ground truth and generated distributions. \\\emph{Top}: $x_1$ vs. $x_2$. \emph{Bottom}: $x_2$ vs $x_3$.}
\label{fig:vae_toy}
\end{minipage}
\begin{minipage}{0.27\textwidth}\centering
\subfloat{\includegraphics[width=0.9\textwidth, trim=1.08in 0.5in 0.84in 0.57in, clip]{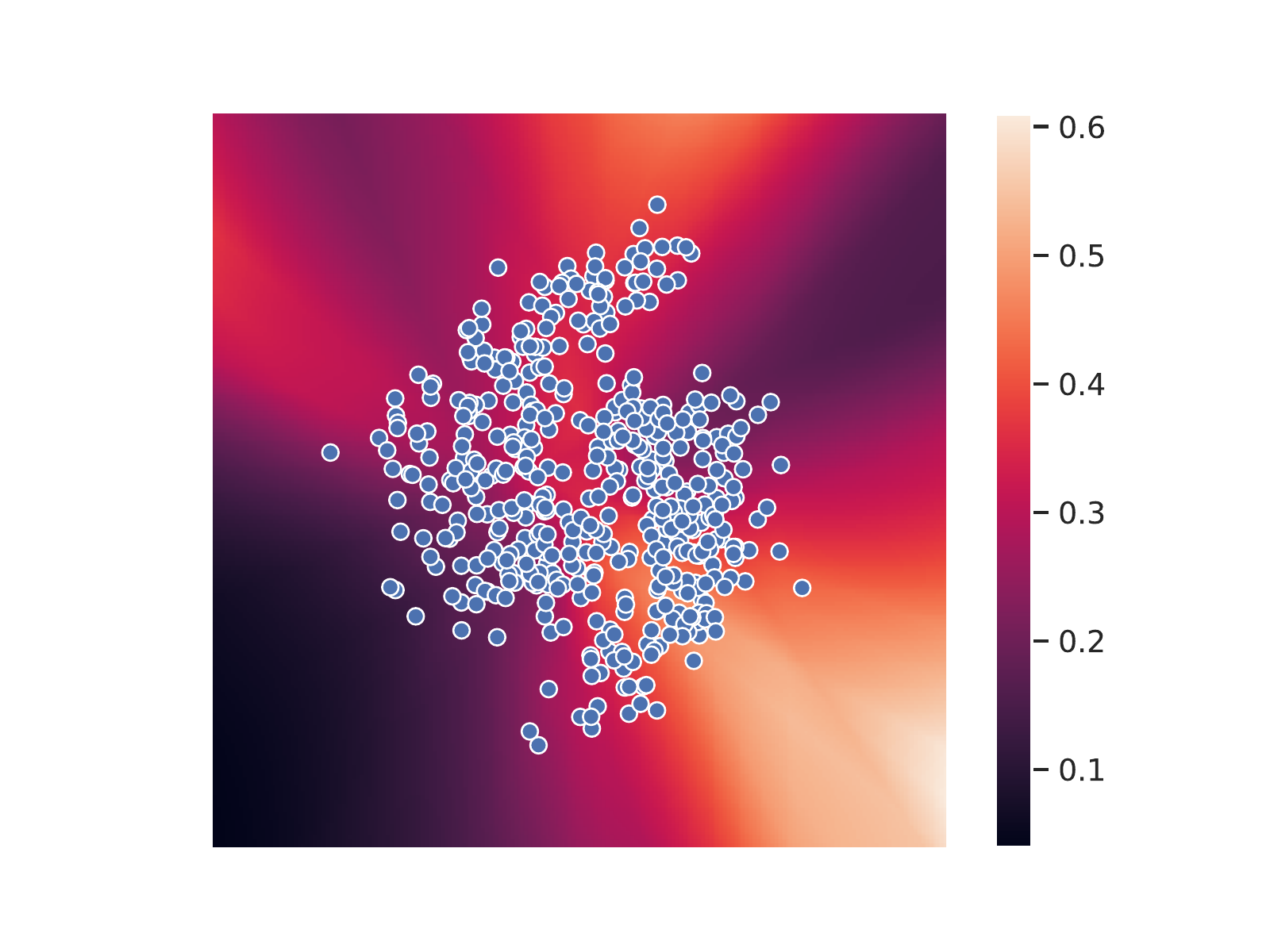}}
\hspace{0.5cm}
\subfloat{\includegraphics[width=0.9\textwidth, trim=1.08in 0.5in 0.84in 0.57in, clip]{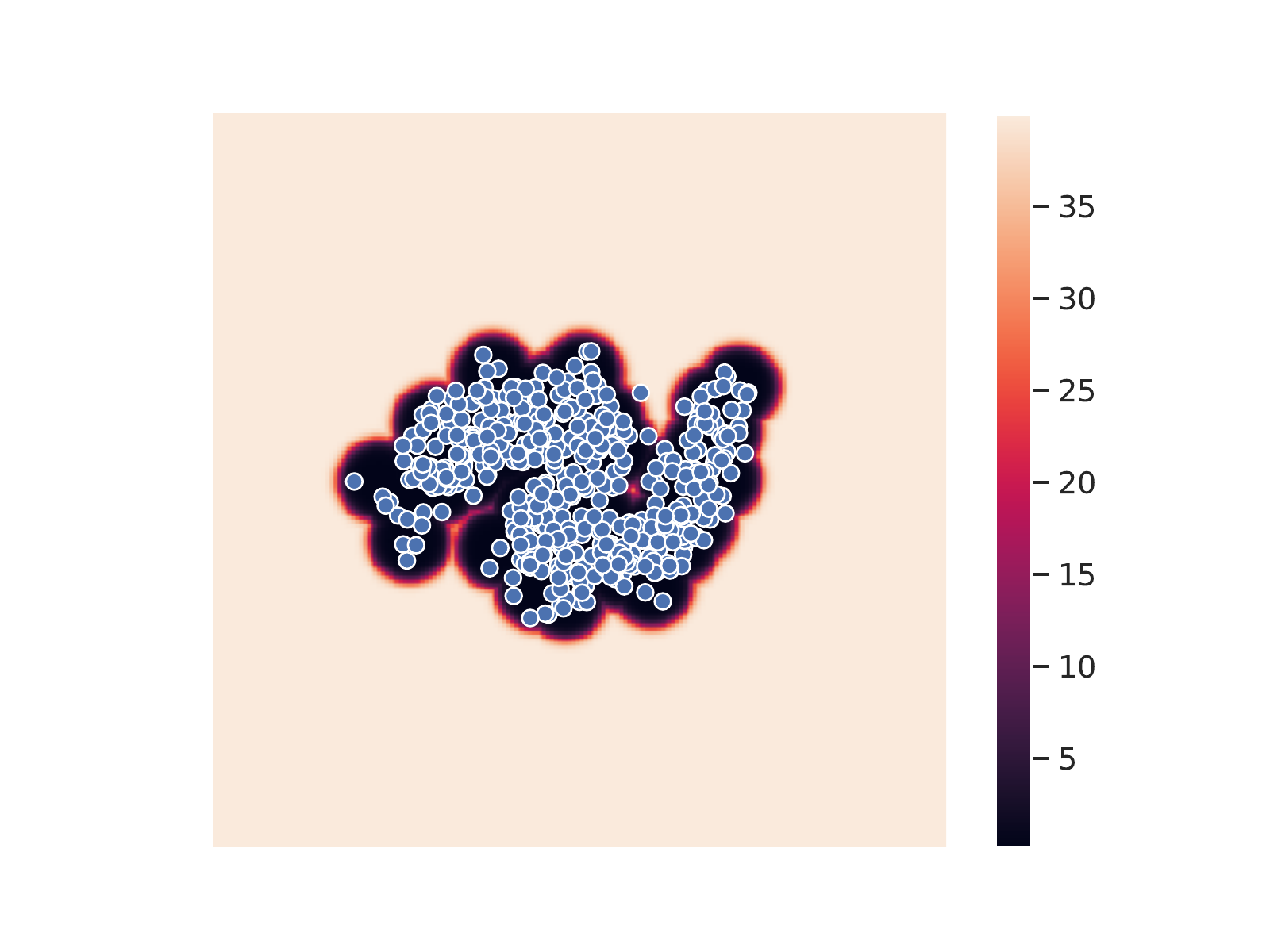}}
\captionof{figure}{Variance estimates in latent space for standard VAE (top) and our Comb-VAE (bottom). Blue points are the encoded training data.}
\label{fig:latent_var}
\end{minipage}
\end{figure}


\paragraph{Artificial data.}
We first evaluate the benefits of more reliable variance networks in VAEs on artificial data. We generate data inspired by the two moon dataset\footnote{\url{https://scikit-learn.org/stable/modules/generated/sklearn.datasets.make_moons.html}}, which we map into four dimensions. The mapping is thoroughly described in the \SUPMAT, and we emphasize that we have deliberately used mappings that MLP's struggle to learn, thus with a low capacity network the only way to compensate is to learn a meaningful variance function. 

In \FIG\ref{fig:vae_toy} we plot pairs of output dimensions using 5000 generated samples. 
For all pairwise combinations we refer to the \SUPMAT. We observe that samples from our Comb-VAE capture the data distribution in more detail than a standard VAE. For VAE the variance seems to be underestimated, which is similar to the results from regression. The poor sample quality of a standard VAE can partially be explained by the arbitrariness of decoder variance function $\sigma^2(\z)$ away from data. In \FIG\ref{fig:latent_var}, we calculated the accumulated variance $\sum_{j=1}^D \sigma^2_j(\z)$ over a grid of latent points. We clearly see that for the standard VAE, the variance is low where we have data and arbitrary away from data. However, our method produces low-variance region where the two half moons are and a high variance region away from data. We note that \citet{arvanitidis2018oddity} also dealt with the problem of arbitrariness of the decoder variance. However their method relies on post-fitting of the variance, whereas ours is fitted during training. Additionally, we note that \citep{takahashi2018studentvae} also successfully modeled the posterior of a VAE as a Student t-distribution similar to our proposed method, but without the extrapolation and different training procedure.\looseness=-1

\begin{table}[]
\resizebox{\textwidth}{!}{
\begin{tabular}{ll|llll}
\multicolumn{1}{l}{}   &						& MNIST    				& FashionMNIST			& CIFAR10 				& SVHN \\ \hline
\multirow{2}{*}{ELBO}  				& VAE      	& 2053.01 $\pm$ 1.60	& 1506.31 $\pm$ 2.71 	& 1980.84 $\pm$ 3.32	& 3696.35 $\pm$ 2.94  \\
                       				& Comb-VAE 	& \textbf{2152.31 $\pm$ 3.32}   & \textbf{1621.29 $\pm$ 7.23}    & \textbf{2057.32 $\pm$ 8.13}     & 3701.41 $\pm$ 5.84     \\ \hline
\multirow{2}{*}{$\log p(x)$} 		& VAE      	& 1914.77 $\pm$ 2.15	& 1481.38 $\pm$ 3.68 	& 1809.43 $\pm$ 10.32 	& 3606.28 $\pm$ 2.75     \\
                       				& Comb-VAE 	& \textbf{2018.37 $\pm$ 4.35}   & \textbf{1567.23 $\pm$ 4.82}    &  \textbf{1891.39 $\pm$ 20.21}   &  3614.39 $\pm$ 7.91   
\end{tabular}
}
\vspace{0.1cm}
\caption{Generative modeling of 4 datasets. For each dataset we report training ELBO and test set log-likelihood. The standard errors are calculated over 3 trained models with random initialization.}
\label{tab:generative_res}
\end{table}

\begin{figure}
\centering
\subfloat{\includegraphics[width=0.49\textwidth]{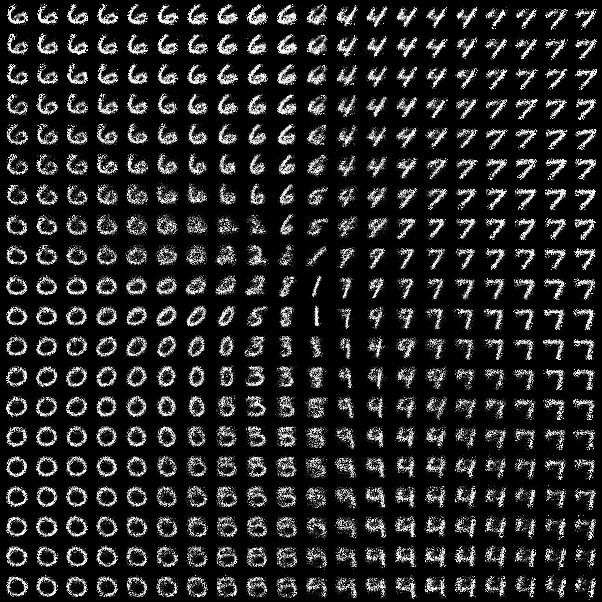}}
\hspace{1mm}
\subfloat{\includegraphics[width=0.49\textwidth]{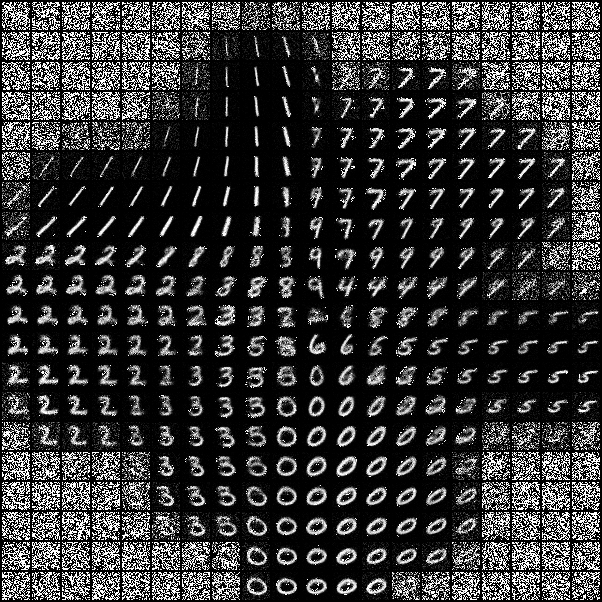}}
\caption{Generated MNIST images on a grid in latent space using the standard variance network (left) and proposed variance network (right).}
\label{fig:johnmnist}
\end{figure}

\paragraph{Image data.}
For our last set of experiments we fitted a standard VAE and our Comb-VAE to four datasets: MNIST, FashionMNIST, CIFAR10, SVHN. We want to measure whether there is an improvement to generative modeling by getting better variance estimation. The details about network architecture and training can be found in the \SUPMAT. Training set ELBO and test set log-likelihoods can be viewed in Table \ref{tab:generative_res}. We observe on all datasets that, on average tighter bounds and higher log-likelihood are achieved, indicating that we better fit the data distribution. 
We quantitatively observe (see \FIG\ref{fig:johnmnist}) that variance has a more local structure for Comb-VAE and that the variance reflects the underlying latent structure. 

%% file: discussion.tex
While variance networks are commonly used for modeling the predictive uncertainty
in regression and in generative modeling, there have been no systematic studies
of how to fit these to data. We have demonstrated that tools developed for
fitting \emph{mean} networks to data are subpar when applied to \emph{variance}
estimation. The key underlying issue appears to be that 
it is not feasible to estimate both a mean and a variance at the same time, when data is scarce.

While it is beneficial to have separate estimates of both \emph{epistemic} and \emph{aleatoric} uncertainty, we have focused on \emph{predictive uncertainty}, which
combine the two. This is a lesser but more feasible goal.

We have proposed a new mini-batching scheme that samples locally to ensure that variances are better defined during model training. We have further argued that variance estimation is more meaningful when conditioned on the mean, which
implies a change to the usual training procedure of joint mean-variance estimation.
To cope with data scarcity we have proposed a more robust likelihood that model a distribution over the variance.
Finally, we have highlighted that variance networks need to extrapolate differently
from mean networks, which implies architectural differences between such networks.
We specifically propose a new architecture for variance networks that ensures
similar variance extrapolations to posterior Gaussian processes from stationary priors.

Our methodologies depend on algorithms that computes Euclidean distances. Since these often break down in high dimensions, this indicates that our proposed methods may not be suitable for high dimensional data. Since we mostly rely on nearest neighbor computations, that empirical are known to perform better in high dimensions, our methodologies may still work in this case. Interestingly, the very definition of variance is dependent on Euclidean distance and this may indicate that variance is inherently difficult to estimate for high dimensional data. This could possible be circumvented through a learned metric.

Experimentally, we have demonstrated that proposed methods are complementary and
provide significant improvements over state-of-the-art. In particular,
on benchmark data we have shown that our method improves upon the test set
log-likelihood without improving the RMSE, which demonstrate that the uncertainty
is a significant improvement over current methods.
Another indicator of improved uncertainty estimation is that our method speeds
up active learning tasks compared to state-of-the-art. Due to the similarities 
between active learning, Bayesian optimization, and reinforcement learning,
we expect that our approach carries significant value to these fields as well. 
Furthermore, we have demonstrated that variational autoencoders can be 
improved through better generative variance estimation.
Finally, we note that our approach is directly applicable alongside ensemble methods,
which may further improve results.

\newpage
\paragraph{Acknowledgements.} 
This project has received funding from the European Research Council (ERC) under the European Union's Horizon 2020 research and innovation programme (grant agreement n\textsuperscript{o} 757360). NSD, MJ and SH were supported in part by a research grant (15334) from VILLUM FONDEN. We gratefully acknowledge the support of NVIDIA Corporation with the donation of GPU hardware used for this research.

%% file: appendix.tex
\section{Further results}
\begin{figure}[h!]
	\centering
	\subfloat{\includegraphics[width=0.4\textwidth]{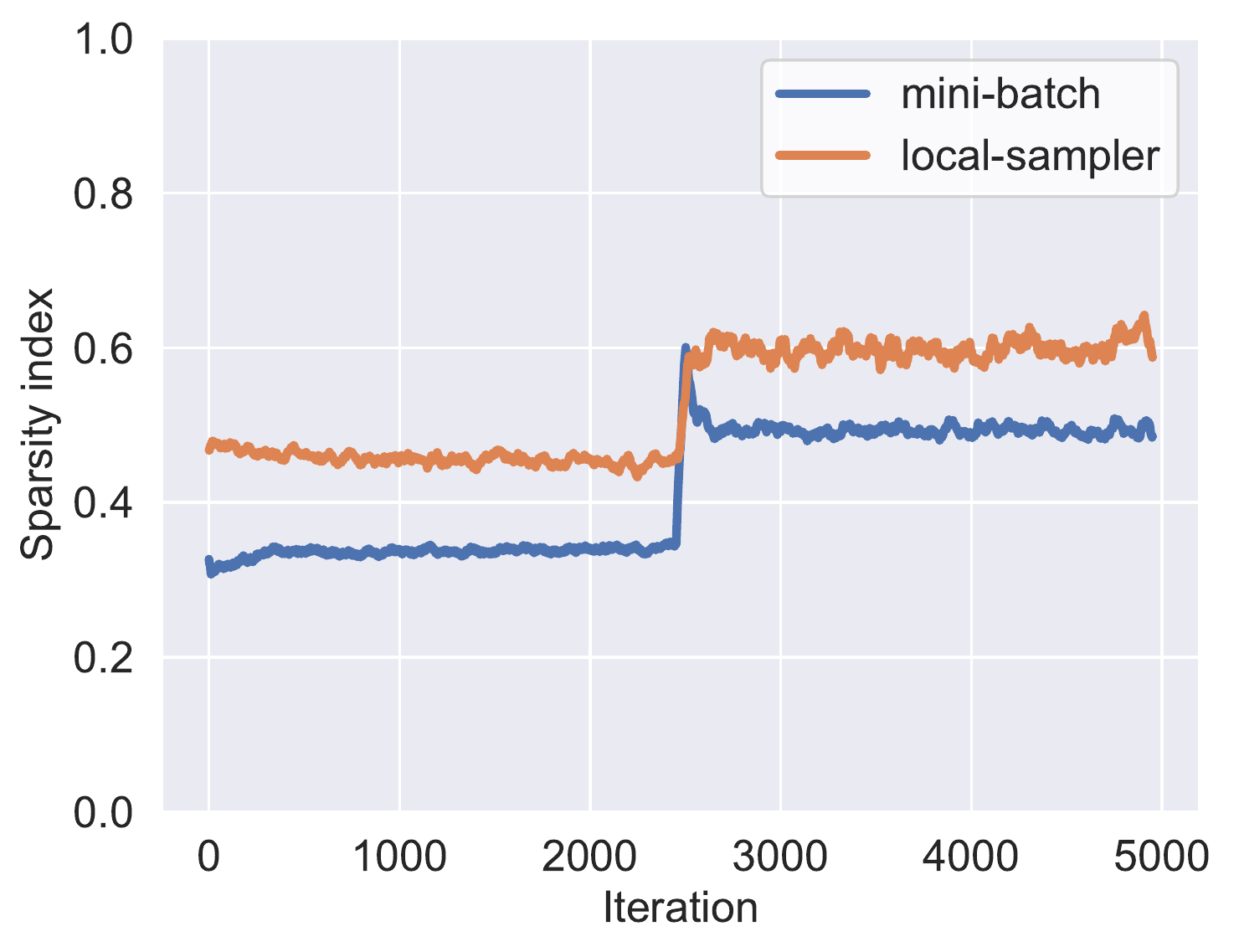}}
	\subfloat{\includegraphics[width=0.45\textwidth]{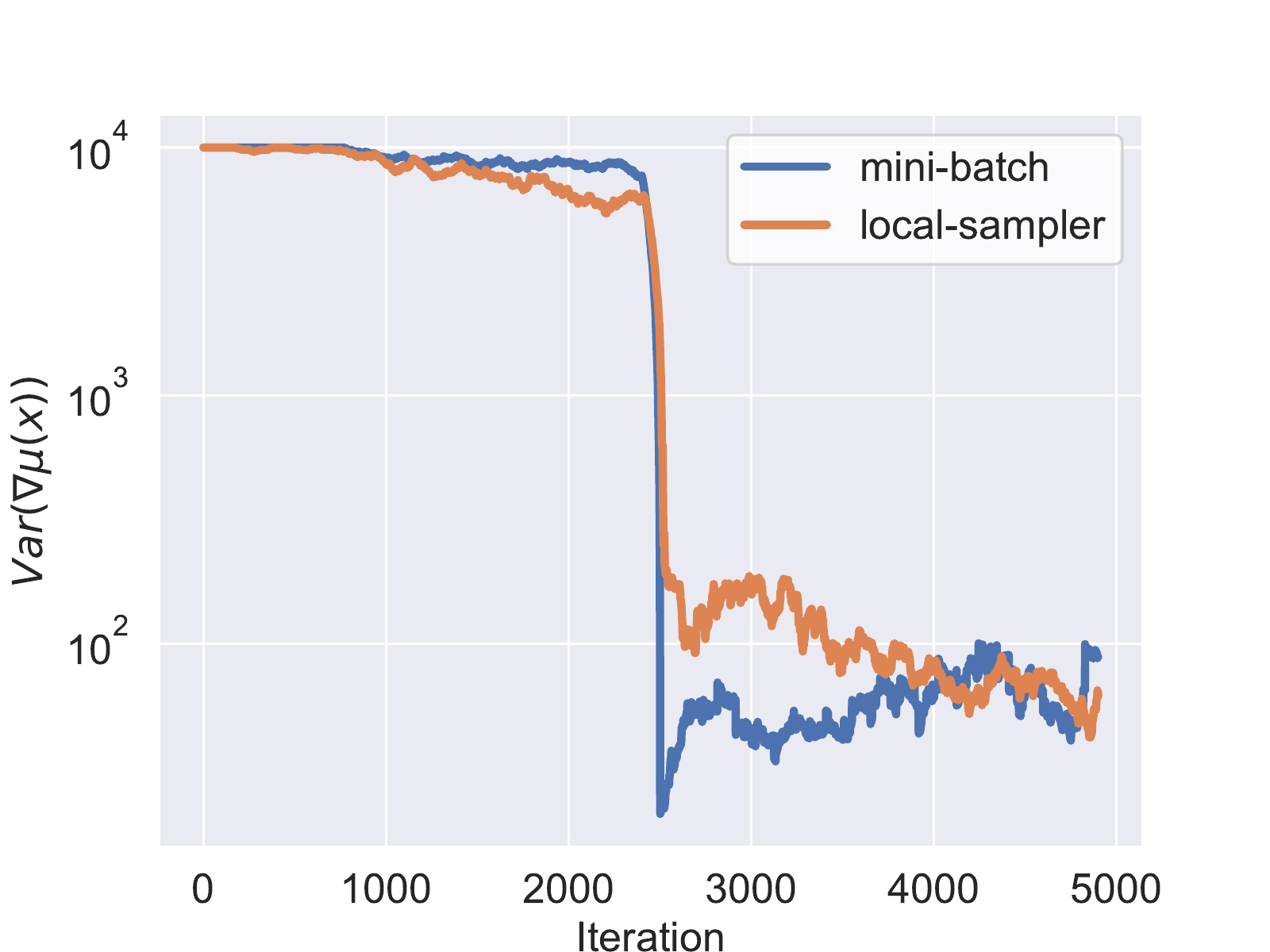}}
	\vspace{-0.4cm}
	\subfloat{\includegraphics[width=0.4\textwidth]{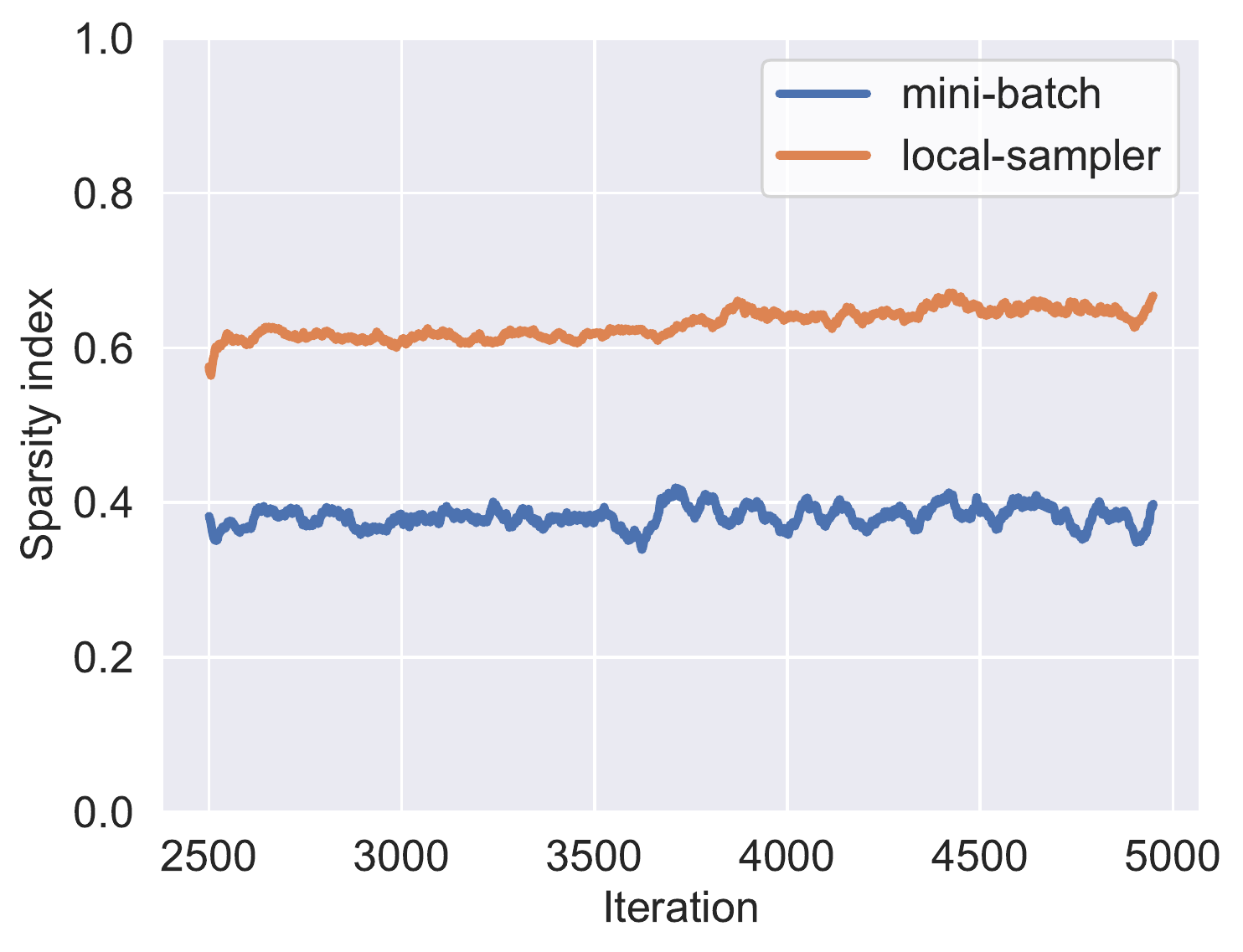}}
	\subfloat{\includegraphics[width=0.45\textwidth]{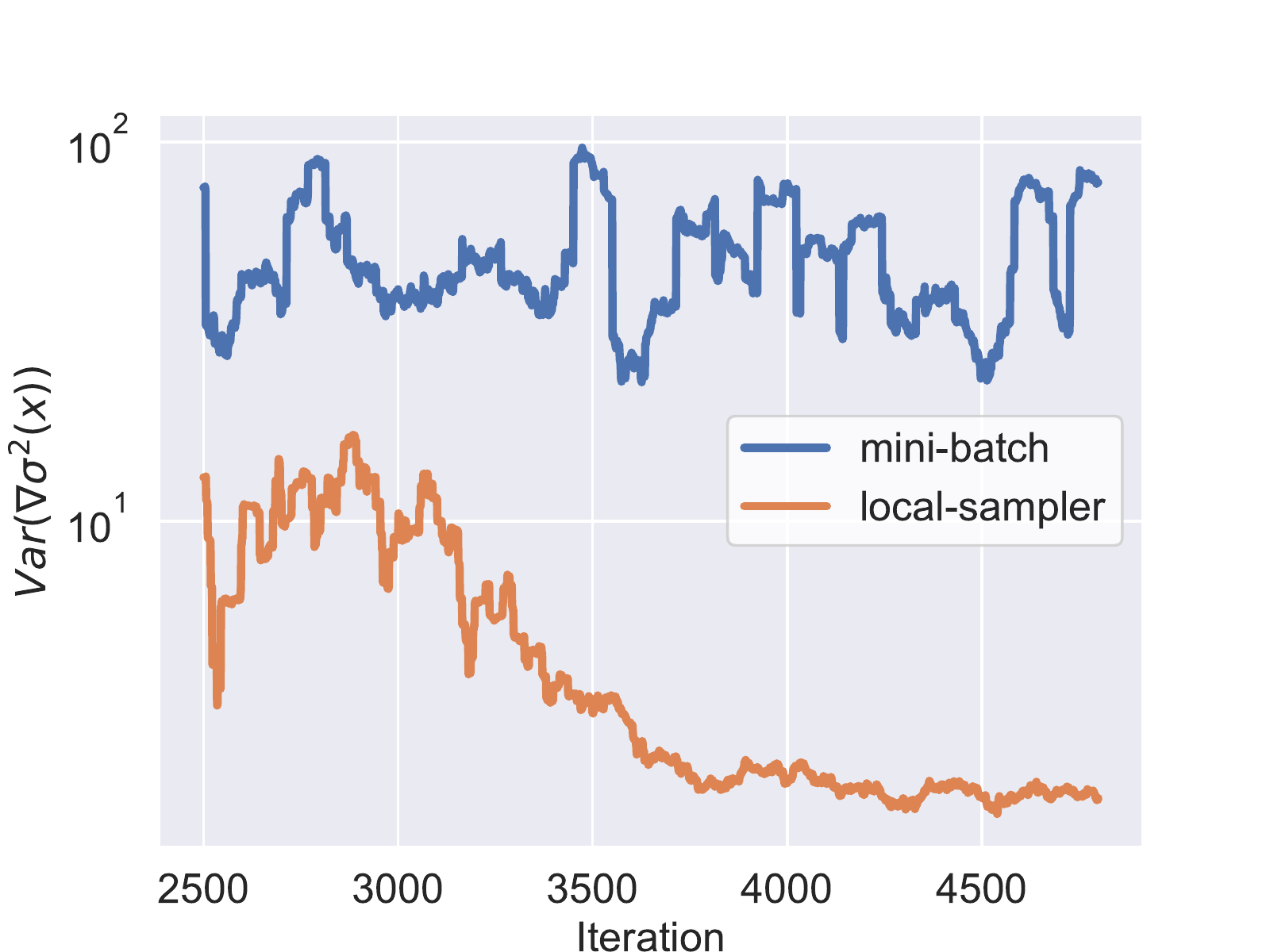}}

	\caption{\textit{Left}: Sparsity index for mean (top) and variance (bottom) network. \textit{Right}: Variance of gradient for mean (top) and variance (bottom) network. The variance network was disabled for the first 2500 iterations, to warm up the mean function to secure convergence.}
	\label{fig:sparsity_grad_var}
\end{figure}

\paragraph{Gradient experiments}
In \FIG\ref{fig:sparsity_grad_var} we have plotted the sparsity index and variance of gradient for both the mean (top row) and variance function (bottom row). We do this for both normal mini-batching and our proposed locality sampler. Sparsity is measured as $\ell_{0.001}^0(\nabla) = \{j, \nabla_j \leq 0.001 \}$ and the sparsity index is then given by $\text{SI}=\frac{\ell_{0.001}^0(\nabla)}{|\nabla|}$. We observe for the mean function, that the sparsity index and variance is similar for the two methods, indicating that our locality sampler does not improve on the fitting of the mean function, as expected. However for the variance function we see a clear gap in sparsity and variance, indicating that our locality sampler gives more local and stable updates to variance networks.

\paragraph{UCI benchmark (RMSE)}
In \TAB\ref{tab:uci_benchmark_rmse} the test set RMSE results for the UCI regression benchmark can be seen. We clearly observe that all neural network based methods achieve nearly identical RMSE for all datasets, indicating that the mean function is similarly trained for all the methods.

\begin{table}[h!]
\centering
\resizebox{\textwidth}{!}{
\begin{tabular}{lll|rrrrrrr}
					 & $N$    & $D$ & \multicolumn{1}{c}{GP}   & \multicolumn{1}{c}{SGP} & \multicolumn{1}{c}{NN} & \multicolumn{1}{c}{BNN} & \multicolumn{1}{c}{MC-Dropout} & \multicolumn{1}{c}{Ens-NN} & \multicolumn{1}{c}{Combined}   \\
		\rowcolor[HTML]{EFEFEF} 
		Boston       & 506    & 13  & 2.79 $\pm$ 0.52 &   2.98 $\pm$ 0.55 &  	4.45 $\pm$ 1.41 &    3.45 $\pm$ 0.87 &    3.01 $\pm$ 0.99 &  3.33 $\pm$ 1.33 &  3.11 $\pm$ 0.35 \\
		Carbon       & 10721  & 7   & - &   1.01 $\pm$ 0.01 &    0.41 $\pm$ 0.0 &     0.18 $\pm$ 0.1 &     0.29 $\pm$ 0.0 &    0.41 $\pm$ 0.0 &   0.35 $\pm$ 0.01 \\
		\rowcolor[HTML]{EFEFEF} 
		Concrete     & 1030   & 8   & 6.03 $\pm$ 0.59  &  6.45 $\pm$ 0.64  &  7.71 $\pm$ 1.32 &   5.78 $\pm$ 0.21 	 &   5.33 $\pm$ 0.65  &  5.65 $\pm$ 0.55 &  5.75 $\pm$ 0.41 \\
		Energy       & 768    & 8   & 1.98 $\pm$ 0.76 	&  2.12 $\pm$ 0.56 	&  1.67 $\pm$ 0.44 &   1.89 $\pm$ 0.04 &   1.69 $\pm$ 0.19 &  2.13 $\pm$ 0.46 &  1.70 $\pm$ 0.21 \\
		\rowcolor[HTML]{EFEFEF} 
		Kin8nm       & 8192   & 8   & - &    0.08 $\pm$ 0.0 &   0.21 $\pm$ 0.01 &    0.18 $\pm$ 0.02 &    0.12 $\pm$ 0.01 &   0.01 $\pm$ 0.01 &   0.12 $\pm$ 0.01 \\
		Naval        & 11934  & 16  & - &     - &    0.01 $\pm$ 0.0 &     0.01 $\pm$ 0.0 &      0.01 $\pm$ 0.0 &    0.01 $\pm$ 0.0 &    0.01 $\pm$ 0.0 \\
		\rowcolor[HTML]{EFEFEF} 
		Power plant  & 9568   & 4   & - &  4.65 $\pm$ 0.12 &  4.23 $\pm$ 0.33 &  4.12 $\pm$ 0.45 &    4.13 $\pm$ 0.13 &    4.11 $\pm$ 0.21 &  4.12 $\pm$ 0.13 \\
		Protein      & 45730  & 9   & - &     - &   4.38 $\pm$ 0.07 &    4.67 $\pm$ 0.94 &    4.19 $\pm$ 0.08 &   4.36 $\pm$ 0.07 &   4.52 $\pm$ 0.19 \\
		\rowcolor[HTML]{EFEFEF} 
		Superconduct & 21263  & 81  & - &  11.32 $\pm$ 0.38 &  11.73 $\pm$ 0.46 &   11.07 $\pm$ 1.7 &   11.44 $\pm$ 0.39 &  11.63 $\pm$ 0.49 &  11.65 $\pm$ 0.65 \\
		Wine (red)   & 1599   & 11  & 0.88 $\pm$ 0.06 &   0.65 $\pm$ 0.04 &   0.66 $\pm$ 0.06 &    0.69 $\pm$ 0.41 &    0.64 $\pm$ 0.06 &    0.67 $\pm$ 0.06 &   0.68 $\pm$ 0.11 \\
		\rowcolor[HTML]{EFEFEF} 
		Wine (white) & 4898   & 11  & - &   0.65 $\pm$ 0.03 &   0.67 $\pm$ 0.04 &    0.68 $\pm$ 0.32 &    0.71 $\pm$ 0.04 &   0.78 $\pm$ 0.04 &   0.72 $\pm$ 0.09 \\
		Yacht        & 308    & 7   & 0.42 $\pm$ 0.21 &   0.72 $\pm$ 0.21 &  1.63 $\pm$ 0.61 &    1.05 $\pm$ 0.11 &     1.11 $\pm$ 0.48 &  1.58 $\pm$ 0.58 &  1.27 $\pm$ 0.11 \\
		\rowcolor[HTML]{EFEFEF} 
		Year         & 515345 & 90  &  - &     - &  12.47 $\pm$ 0.96 &    9.01 $\pm$ 0.45 &  8.92 $\pm$ 0.23 &  8.88 $\pm$ 0.13 &  8.85 $\pm$ 0.05 \\
\end{tabular} 
}
\vspace{0.1cm}
\caption{Dataset characteristics and RMSE for the different methods. A - indicates the models was infeasible to train.} 
\label{tab:uci_benchmark_rmse}
\end{table}

\paragraph{Timings of models}
In \TAB\ref{tab:timings} we show the average computation time for each model. The experiments was conducted with an Intel Xeon E5-2620v4 CPU and Nvidia GTX TITAN X GPU. We note that our model suffers from long computations for very large datasets, mainly due to the computation of the neighborhood graph in the locality sampler. This could be reduced by using fast approximative method for k-nearest-neighbor and by reducing data dimensionality \eg PCA dimensionality reduction. 

\begin{table}[h!]
\centering
\resizebox{\textwidth}{!}{
\begin{tabular}{lll|rrrrrrr}
		& $N$    & $D$ & \multicolumn{1}{c}{GP}   & \multicolumn{1}{c}{SGP} & \multicolumn{1}{c}{NN} & \multicolumn{1}{c}{BNN} & \multicolumn{1}{c}{MC-Dropout} & \multicolumn{1}{c}{Ens-NN} & \multicolumn{1}{c}{Combined}   \\
\rowcolor[HTML]{EFEFEF} 
Boston &  506 & 13           &    8.37 +- 2.92 &     91.86 +- 30.12 &   94.04 +- 2.24 &   81.32 +- 1.83 &   98.08 +- 2.0 &   479.1 +- 12.49 &    93.39 +- 1.82 \\
Carbon  & 10721 & 7 &      - &    192.57 +- 72.28 &    90.05 +- 3.4 &   80.95 +- 2.04 &  98.62 +- 1.85 &  439.45 +- 23.01 &   123.61 +- 2.84 \\
\rowcolor[HTML]{EFEFEF} 
Concrete & 1030 & 8           &     7.48 +- 1.2 &      173.73 +- 4.0 &   92.91 +- 4.17 &   81.02 +- 2.04 &  97.94 +- 1.84 &  468.94 +- 10.81 &     97.65 +- 6.4 \\
Energy	& 768 & 8			 &   10.48 +- 4.12 &     121.39 +- 52.2 &   92.91 +- 2.14 &   80.92 +- 2.14 &  97.86 +- 2.13 &  475.04 +- 12.61 &    93.01 +- 1.29 \\
\rowcolor[HTML]{EFEFEF} 
Kin8nm & 8192 & 8             &      - &   1526.29 +- 20.28 &   92.22 +- 4.65 &   80.65 +- 2.11 &  97.85 +- 2.87 &  459.81 +- 27.66 &   123.15 +- 2.86 \\
Navel  & 11934 & 16             &      - &       9.79 +- 0.23 &    91.2 +- 3.28 &   82.97 +- 1.84 &  98.64 +- 1.89 &   482.74 +- 8.54 &   136.25 +- 3.29 \\
\rowcolor[HTML]{EFEFEF} 
Power & 9568 & 4        &      - &  1267.82 +- 783.28 &   92.23 +- 3.28 &   81.29 +- 1.98 &  98.26 +- 1.87 &   472.39 +- 9.73 &   118.26 +- 1.86 \\
Protein & 45730 & 9  &      - &         - &  138.49 +- 2.51 &  124.72 +- 1.86 &  140.73 +- 3.32 &  707.69 +- 11.02 &  658.63 +- 11.75 \\
\rowcolor[HTML]{EFEFEF} 
Superconduct & 21263 & 81       &      - &       313.9 +- 2.9 &   95.27 +- 3.43 &   90.71 +- 1.86 &  90.25 +- 1.67 &  477.57 +- 11.91 &    235.72 +- 3.8 \\
Wine (red) & 1599 & 11         &  35.33 +- 19.09 &    262.69 +- 10.14 &   93.51 +- 2.09 &   80.67 +- 2.04 &  97.78 +- 2.58 &   416.02 +- 23.9 &   130.61 +- 7.91 \\
\rowcolor[HTML]{EFEFEF} 
Wine (white) & 4898 & 11       &      - &      781.13 +- 8.2 &   91.97 +- 2.41 &   80.94 +- 1.75 &  97.61 +- 2.73 &  451.65 +- 19.04 &    99.44 +- 1.33 \\
Yacht & 308 & 7				 &    0.93 +- 0.32 &     22.74 +- 11.98 &   92.82 +- 3.47 &    81.0 +- 1.92 &  98.27 +- 1.68 &   422.9 +- 35.85 &   105.33 +- 2.32 \\
\rowcolor[HTML]{EFEFEF} 
Year  & 515345 & 90   			&      - &         - &  139.45 +- 6.15 &  643.96 +- 4.17 &  77.96 +- 2.09 &  725.27 +- 20.08 &  2453.62 +- 18.7 \\
\end{tabular}
}
\vspace{0.1cm}
\caption{Timings(s) for the different models evaluated on the UCI benchmark }
\label{tab:timings}
\end{table}

\paragraph{Ablation study (RMSE)}

\begin{figure}[h!]
	\subfloat{\includegraphics[width=0.325\textwidth, trim=0cm 0cm 0cm 0cm, clip]{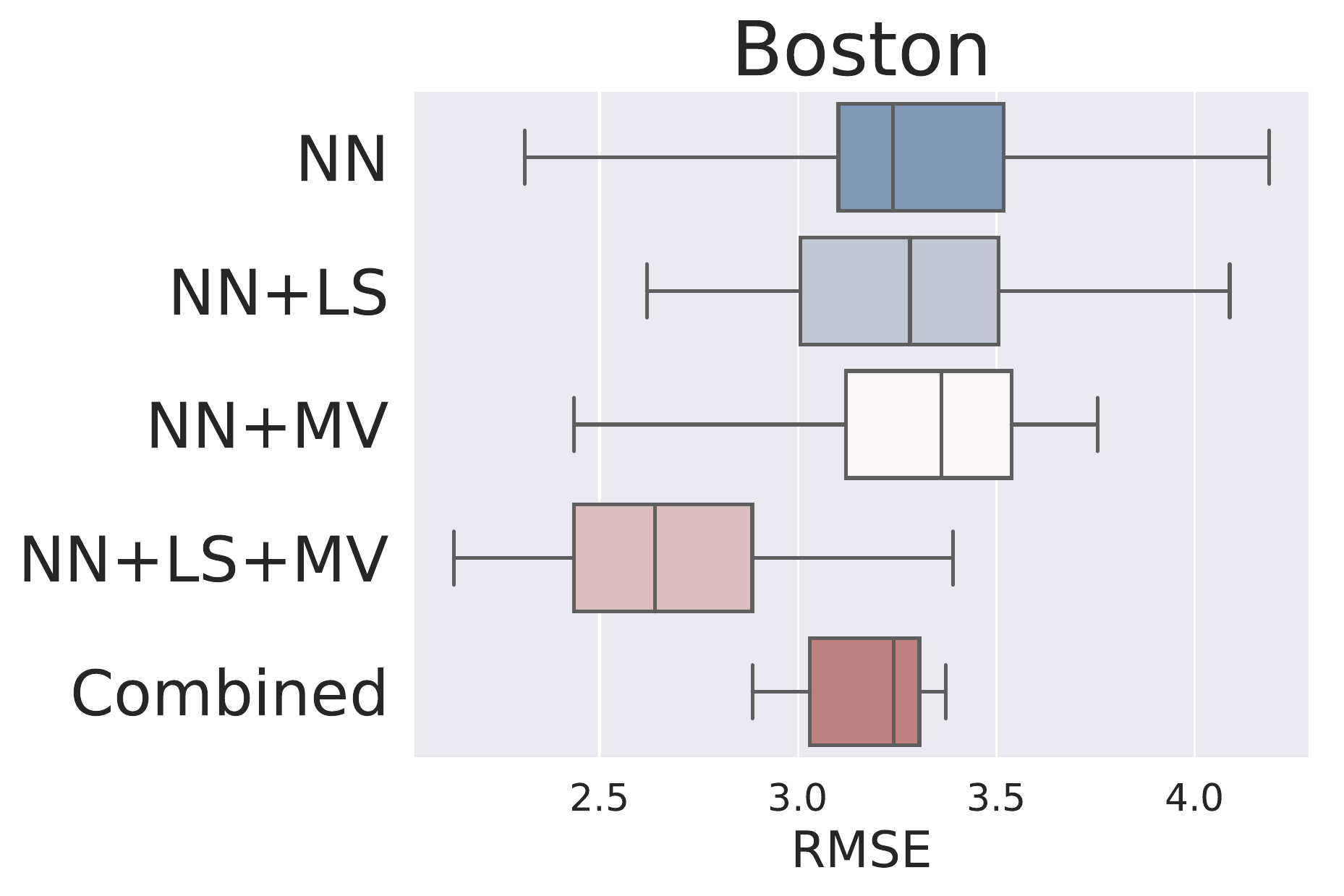}}
	\subfloat{\includegraphics[width=0.23\textwidth, trim=5.5cm 0cm 0cm 0cm, clip]{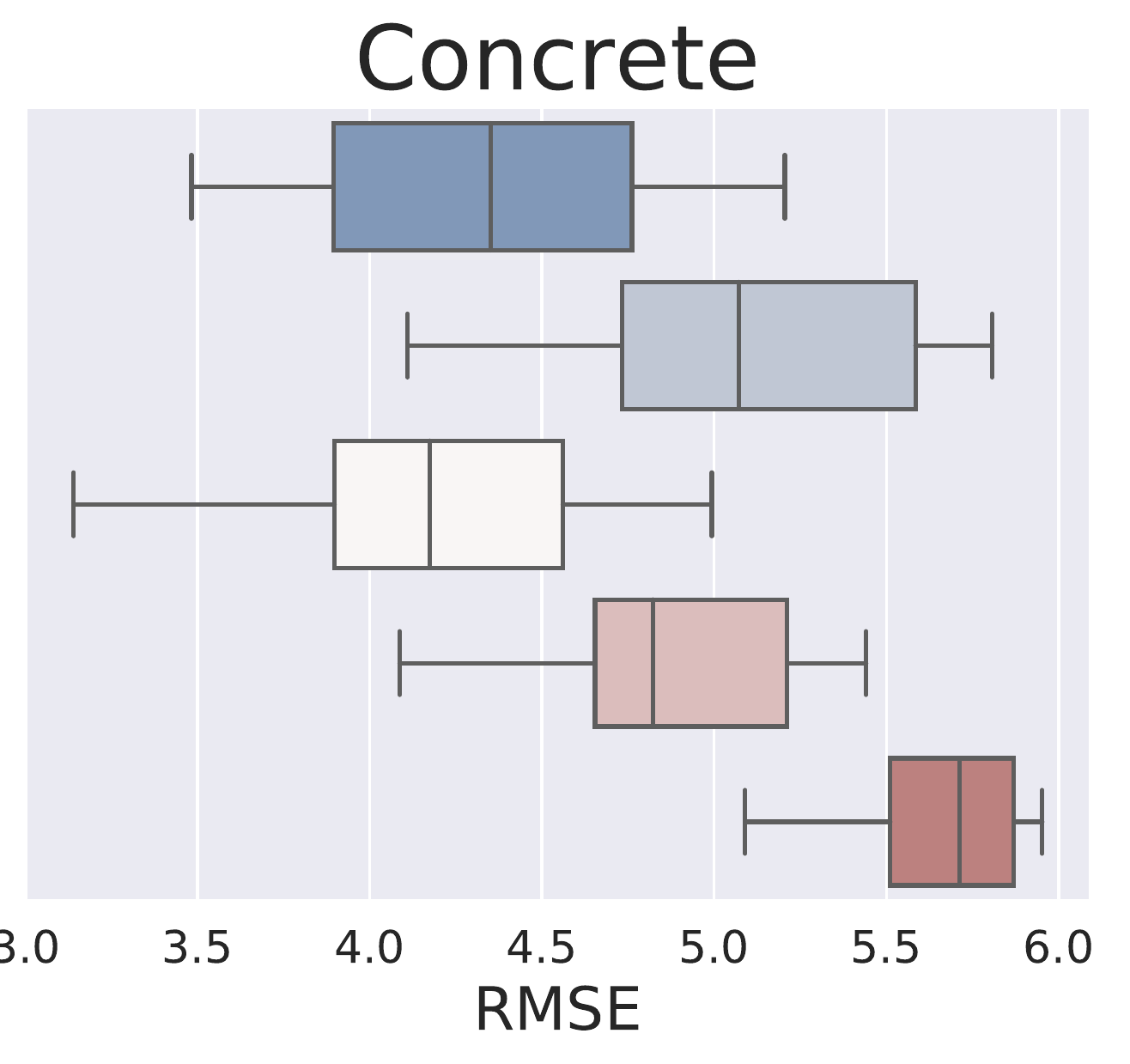}}
	\subfloat{\includegraphics[width=0.23\textwidth, trim=5.5cm 0cm 0cm 0cm, clip]{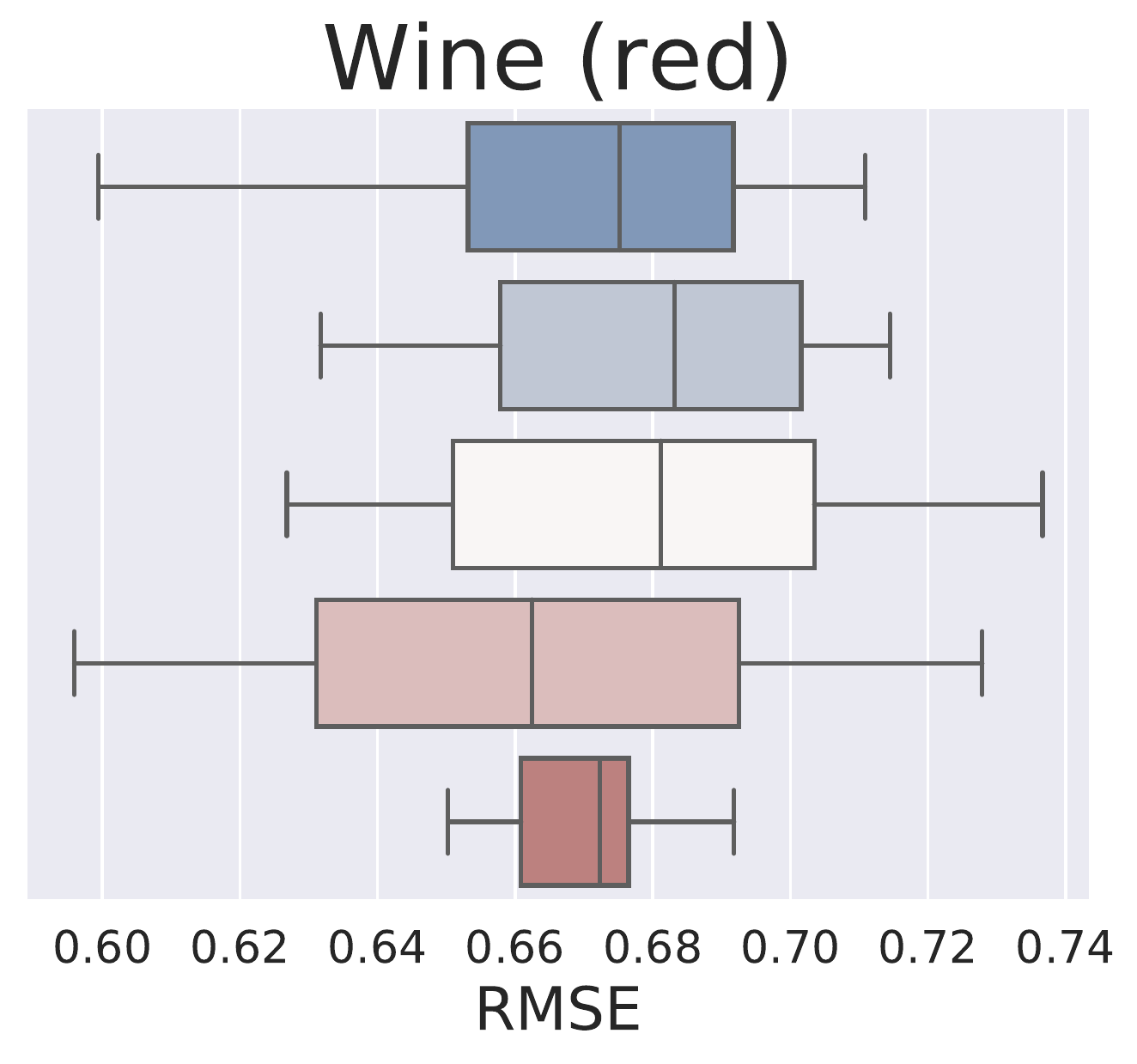}}
	\subfloat{\includegraphics[width=0.23\textwidth, trim=5.5cm 0cm 0cm 0cm, clip]{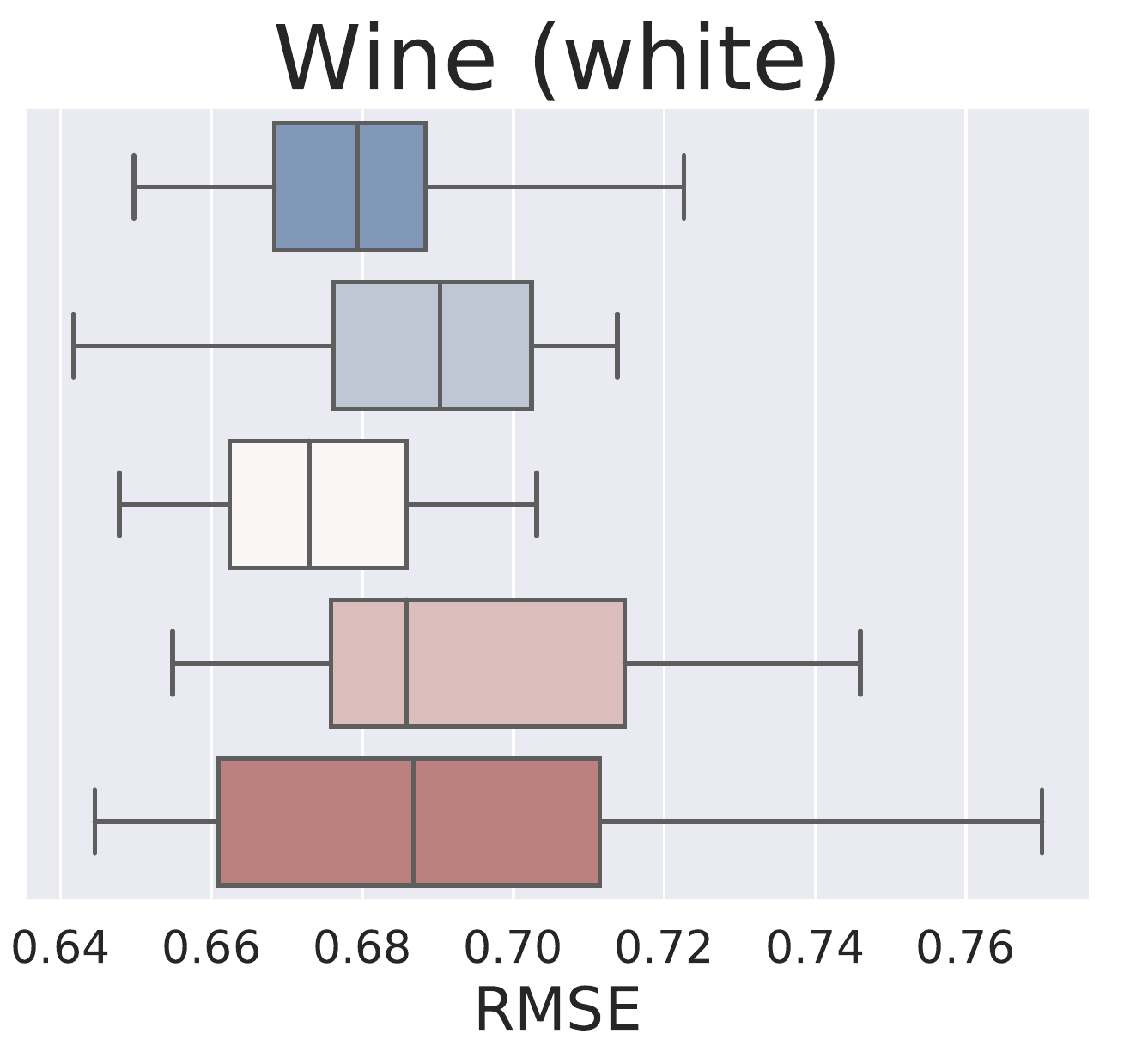}}
	\caption{Our contributions evaluated on four different UCI benchmark datasets.}
	\label{fig:contrib_rmse}
\end{figure}

In \FIG\ref{fig:contrib_rmse} we have plotted the RMSE for different combinations of our methodologies. Since the RMSE is only influenced by how well $\mu(x)$ is fitted, the difference in log likelihood that we observe between the models (see paper) must be explained by how well $\sigma^2(x)$ is fitted.

\paragraph{Active learning}

\begin{figure}[h!]
\subfloat{\includegraphics[width=0.14\textwidth]{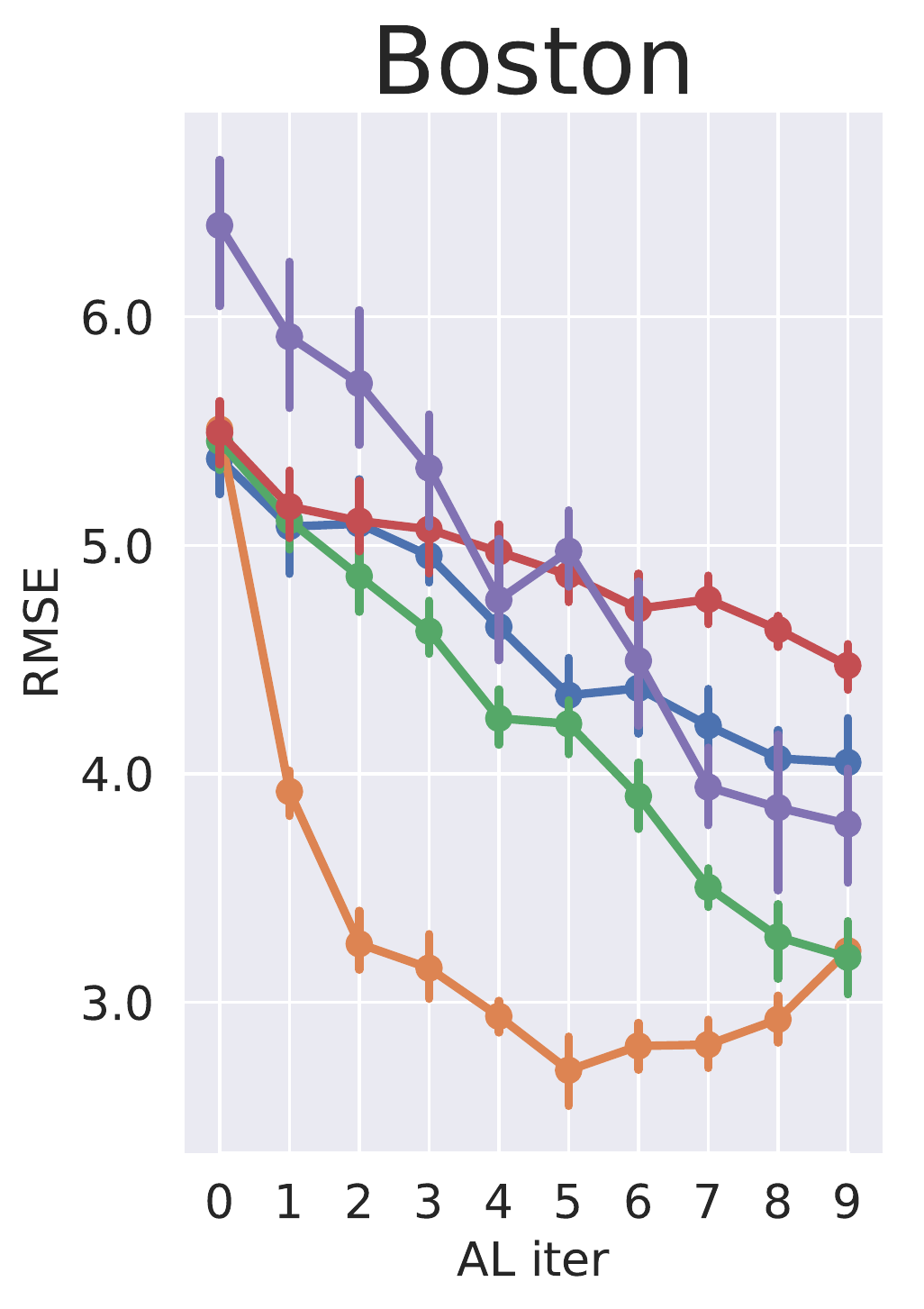}}
\subfloat{\includegraphics[width=0.14\textwidth]{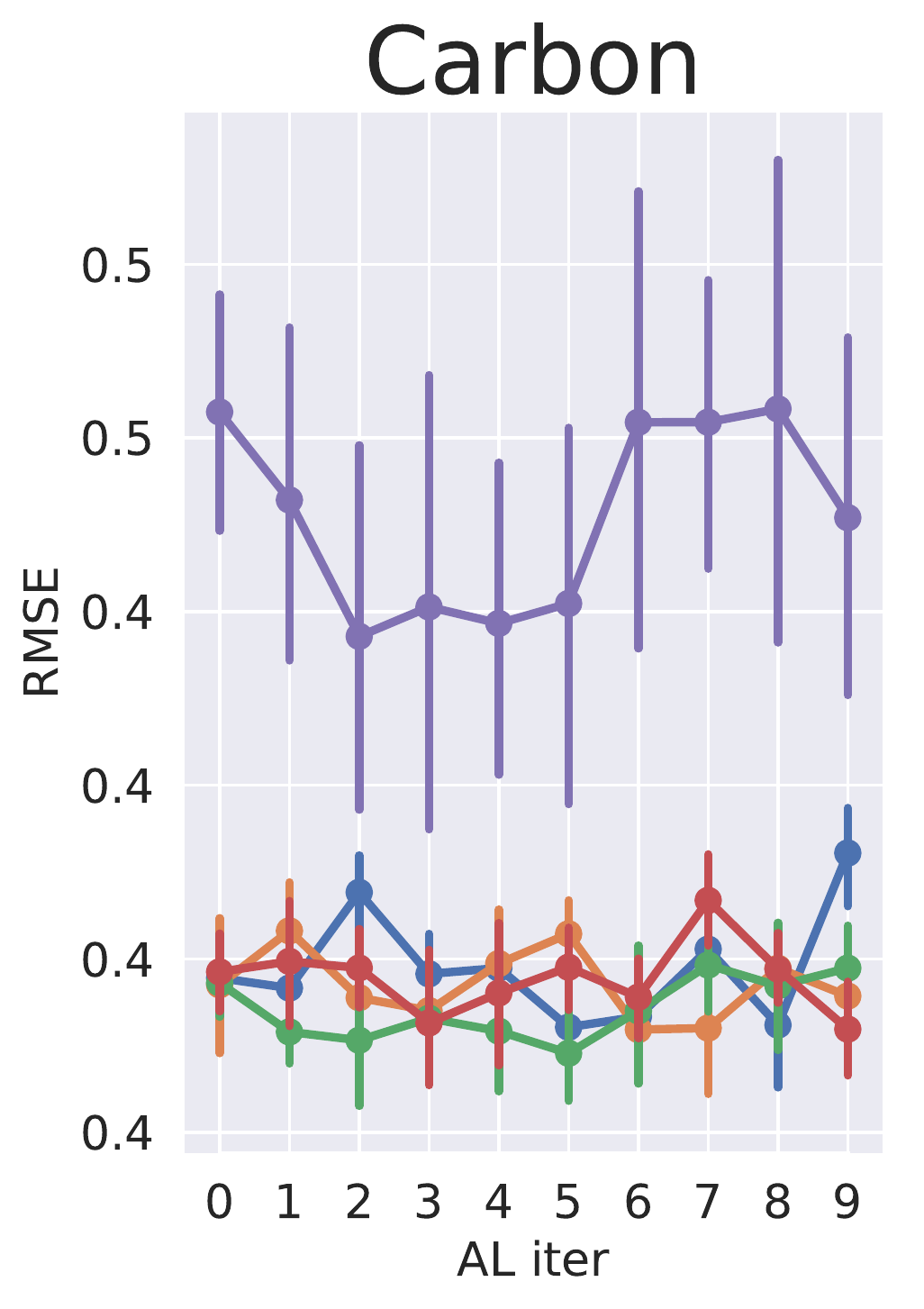}}
\subfloat{\includegraphics[width=0.14\textwidth]{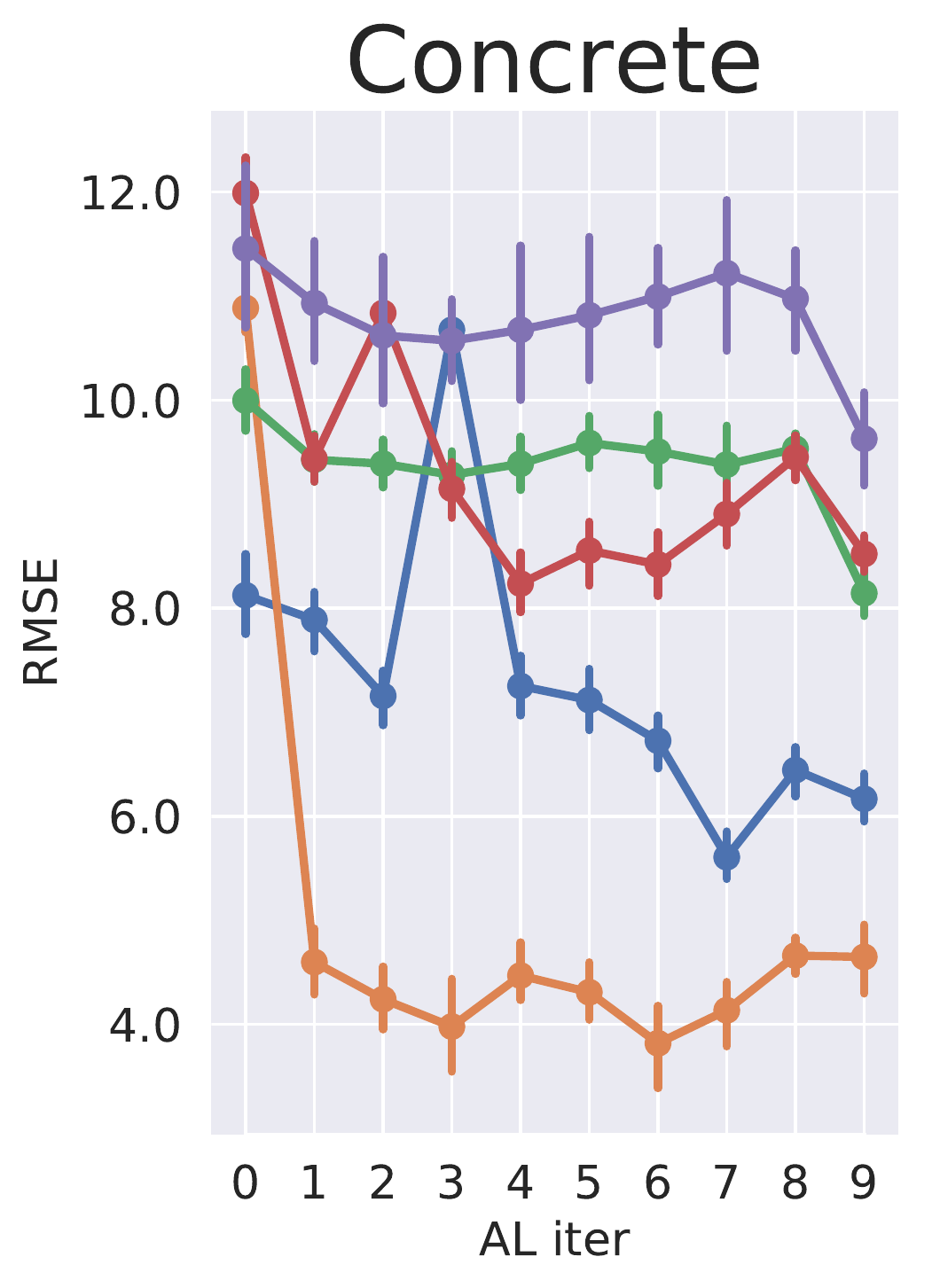}}
\subfloat{\includegraphics[width=0.14\textwidth]{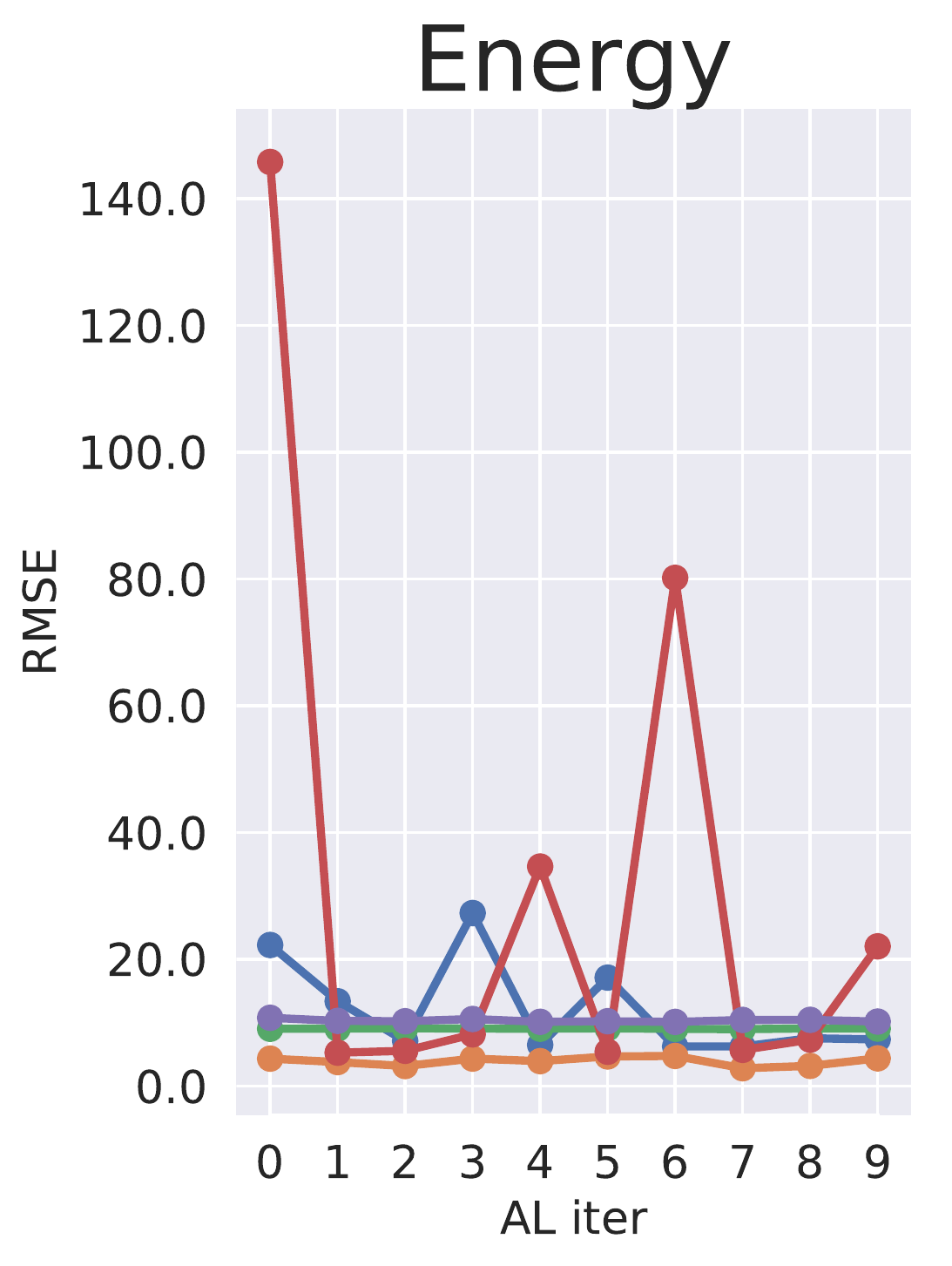}}
\subfloat{\includegraphics[width=0.14\textwidth]{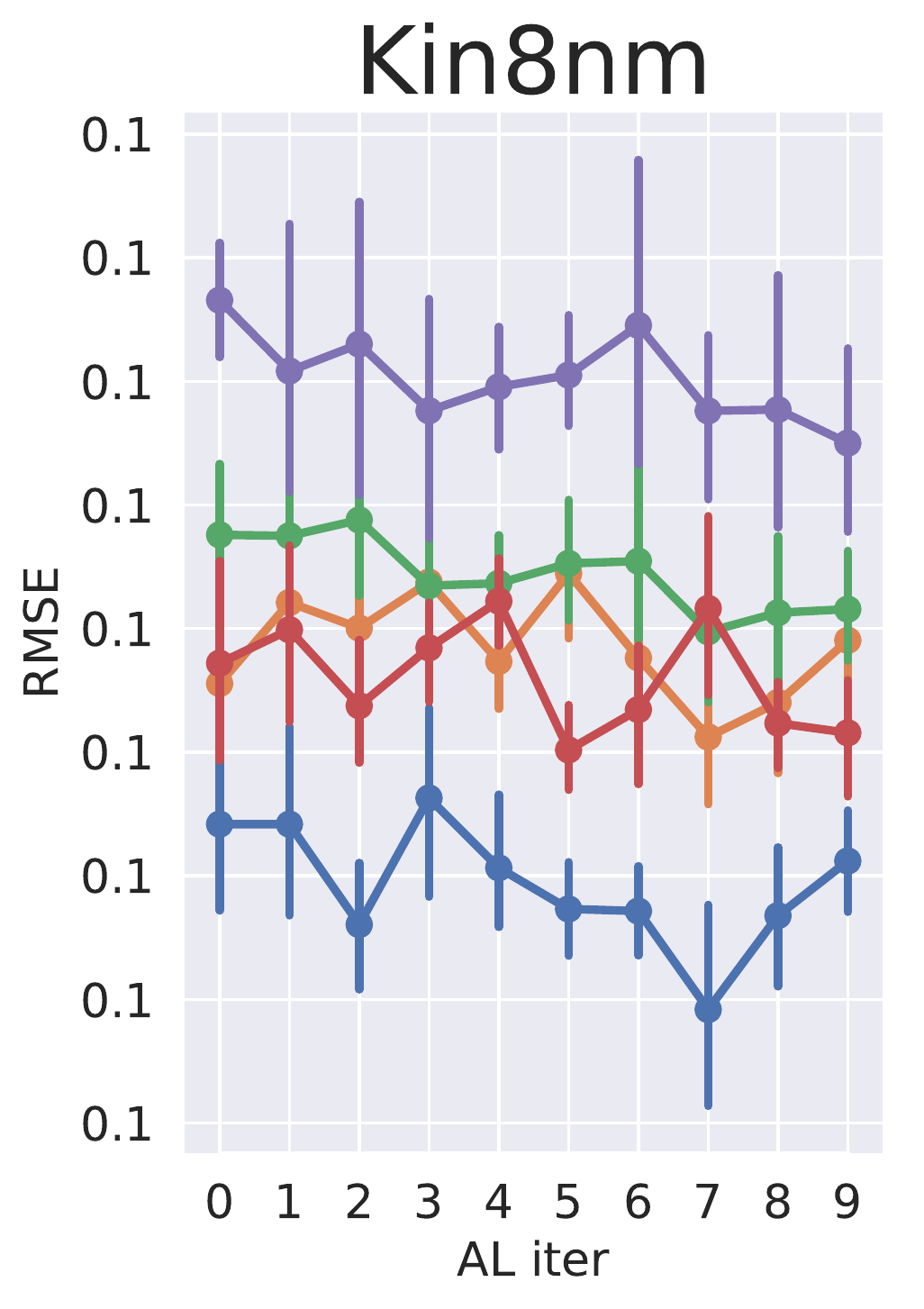}}
\subfloat{\includegraphics[width=0.14\textwidth]{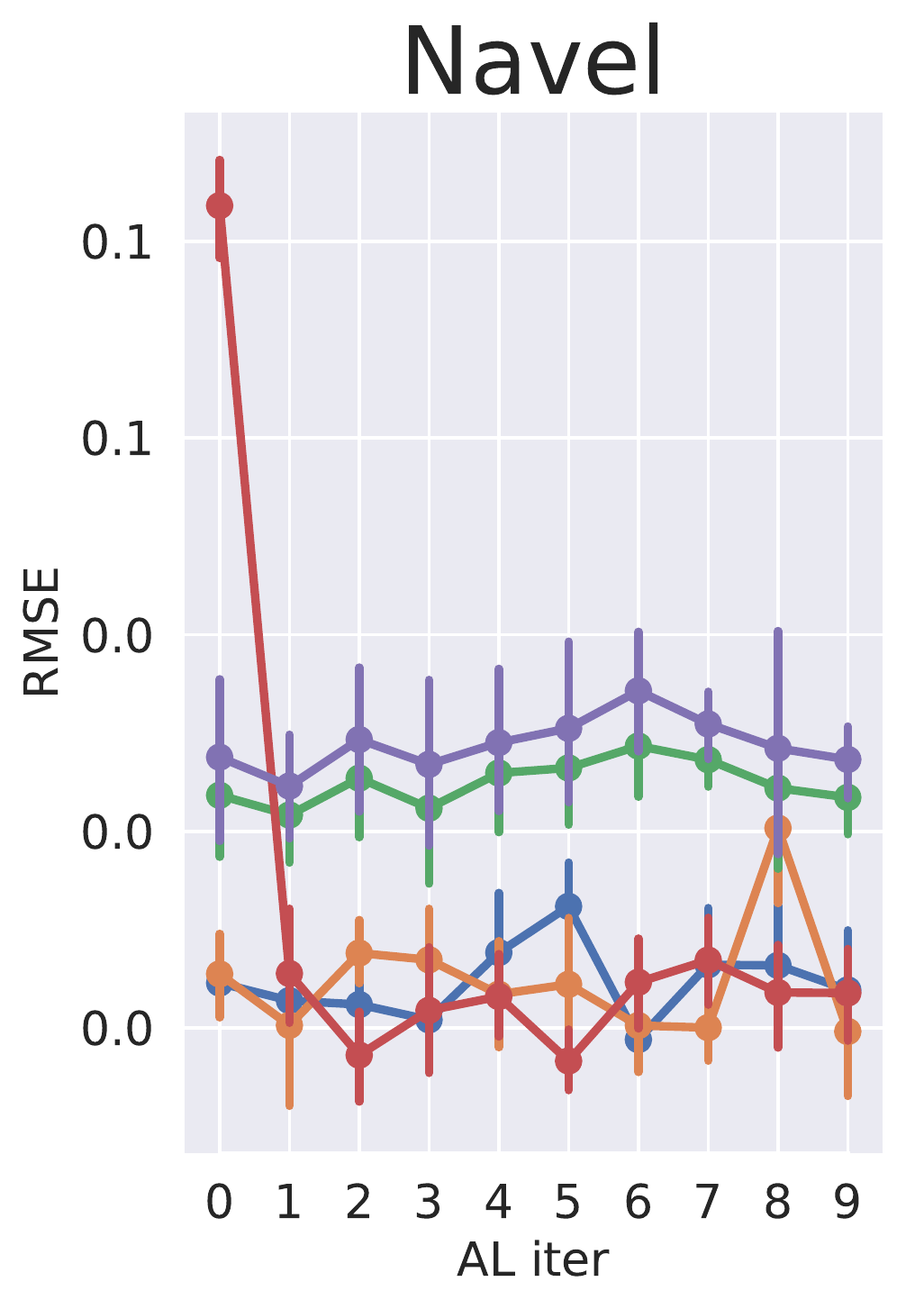}}
\subfloat{\includegraphics[width=0.14\textwidth]{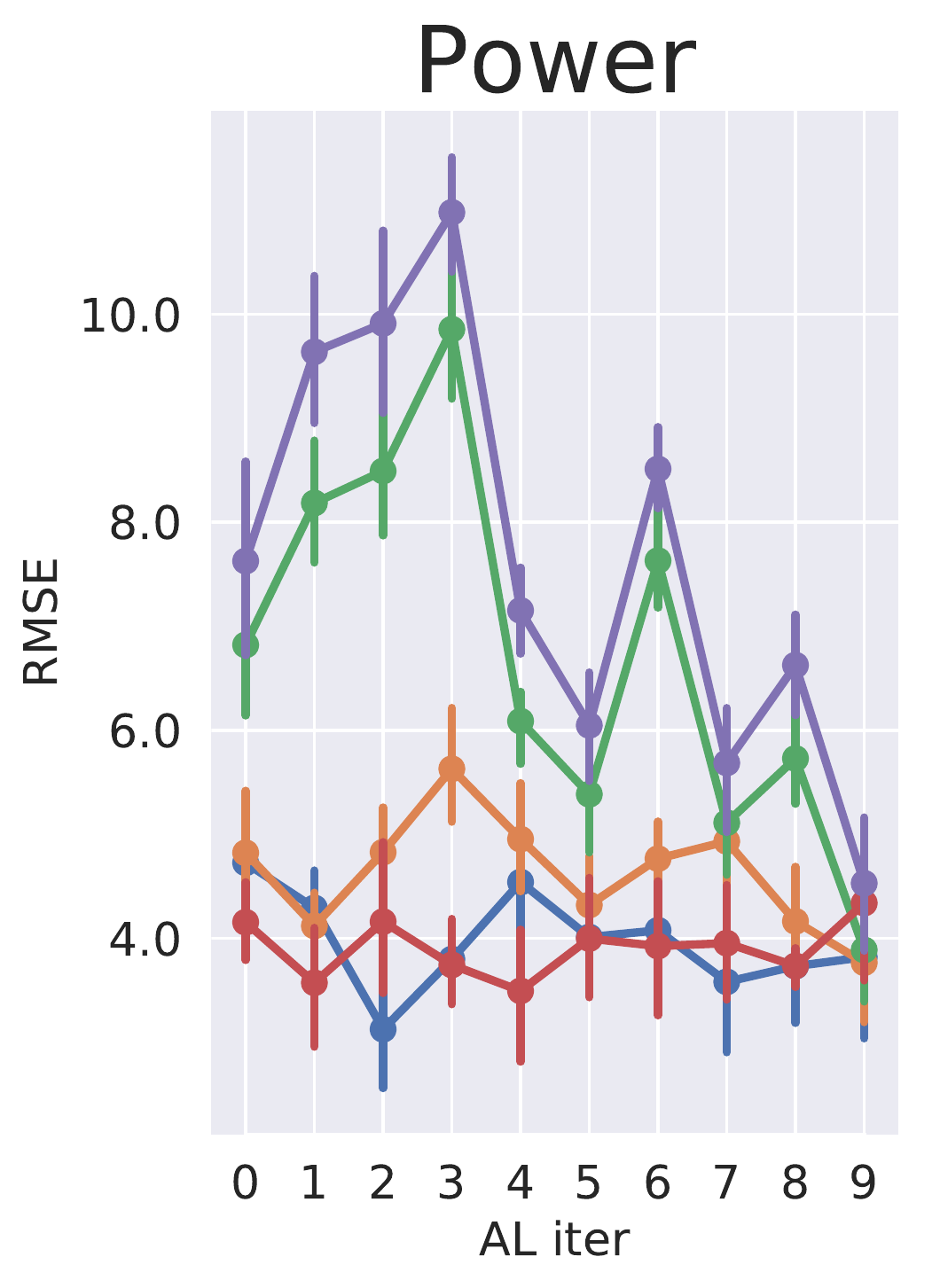}}

\subfloat{\includegraphics[width=0.14\textwidth]{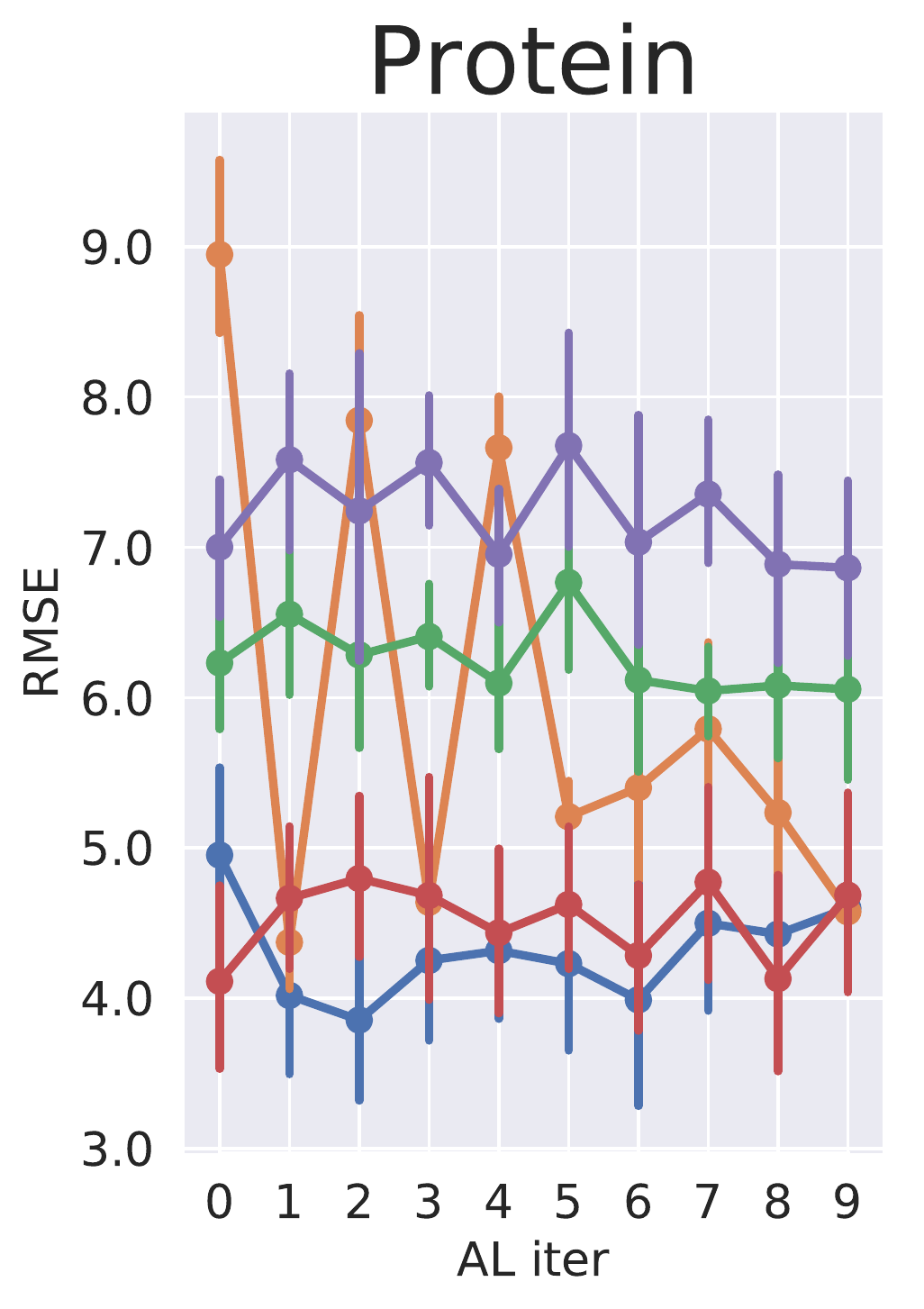}}
\subfloat{\includegraphics[width=0.14\textwidth]{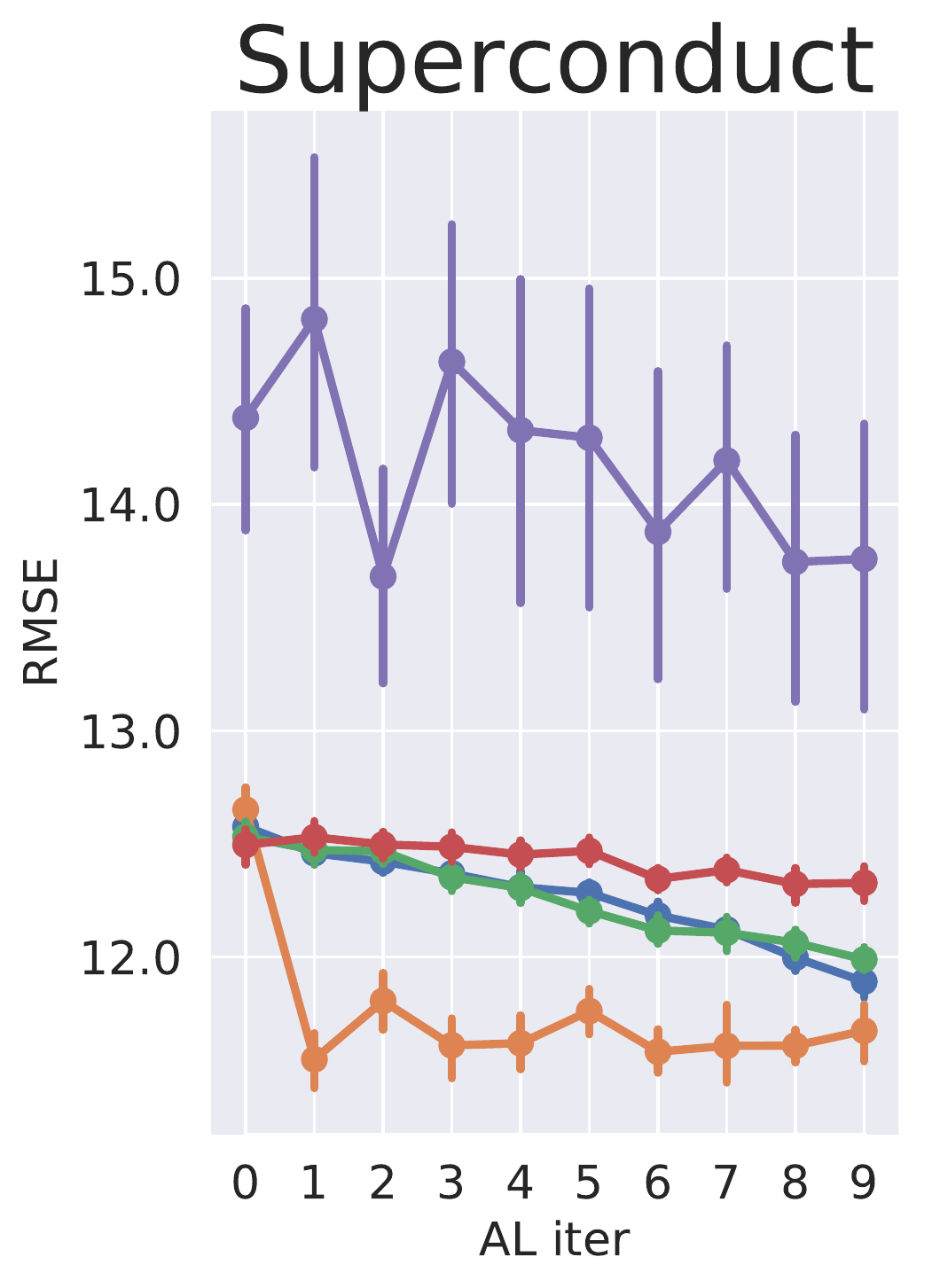}}
\subfloat{\includegraphics[width=0.14\textwidth]{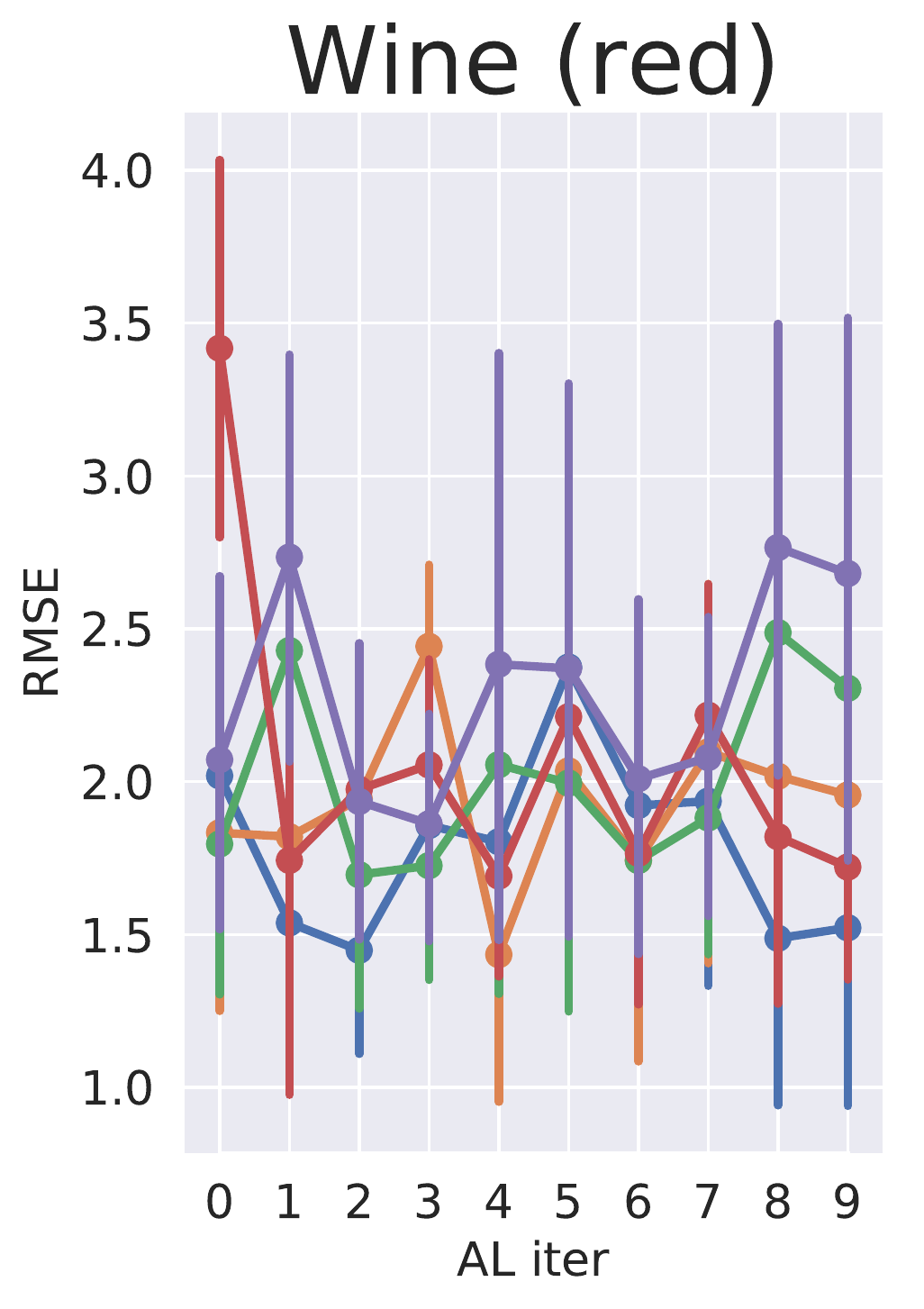}}
\subfloat{\includegraphics[width=0.14\textwidth]{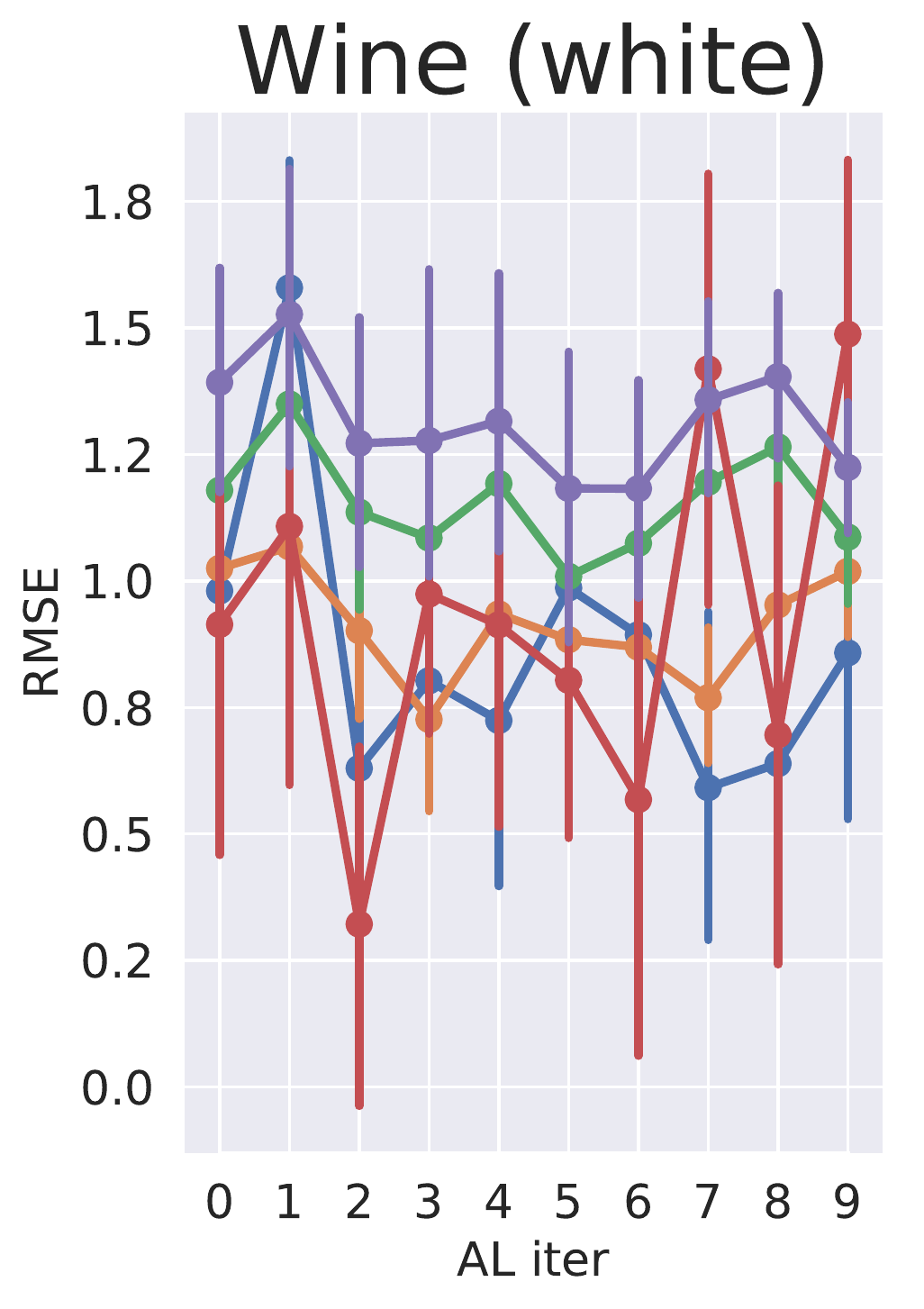}}
\subfloat{\includegraphics[width=0.14\textwidth]{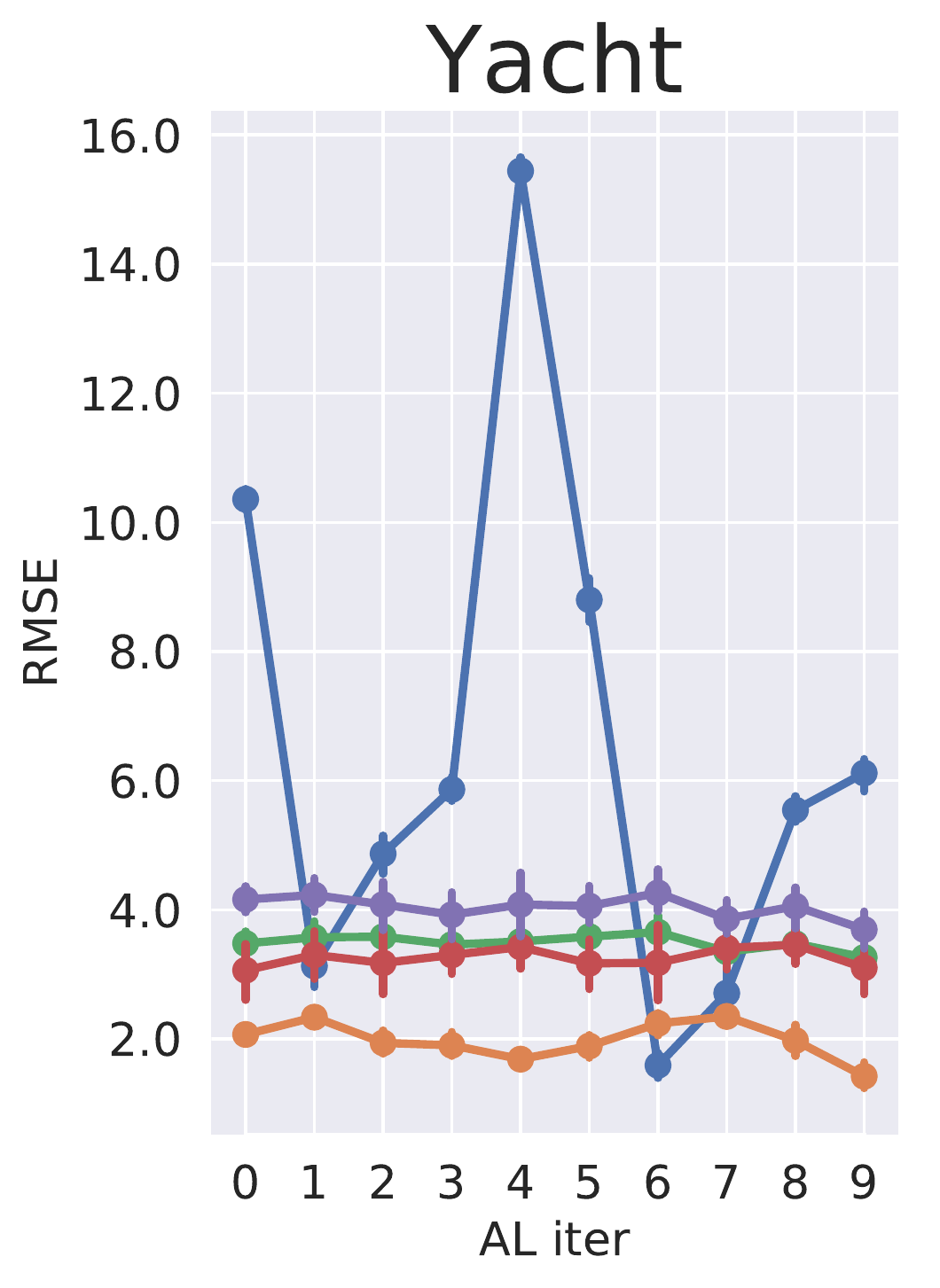}}
\subfloat{\includegraphics[width=0.14\textwidth]{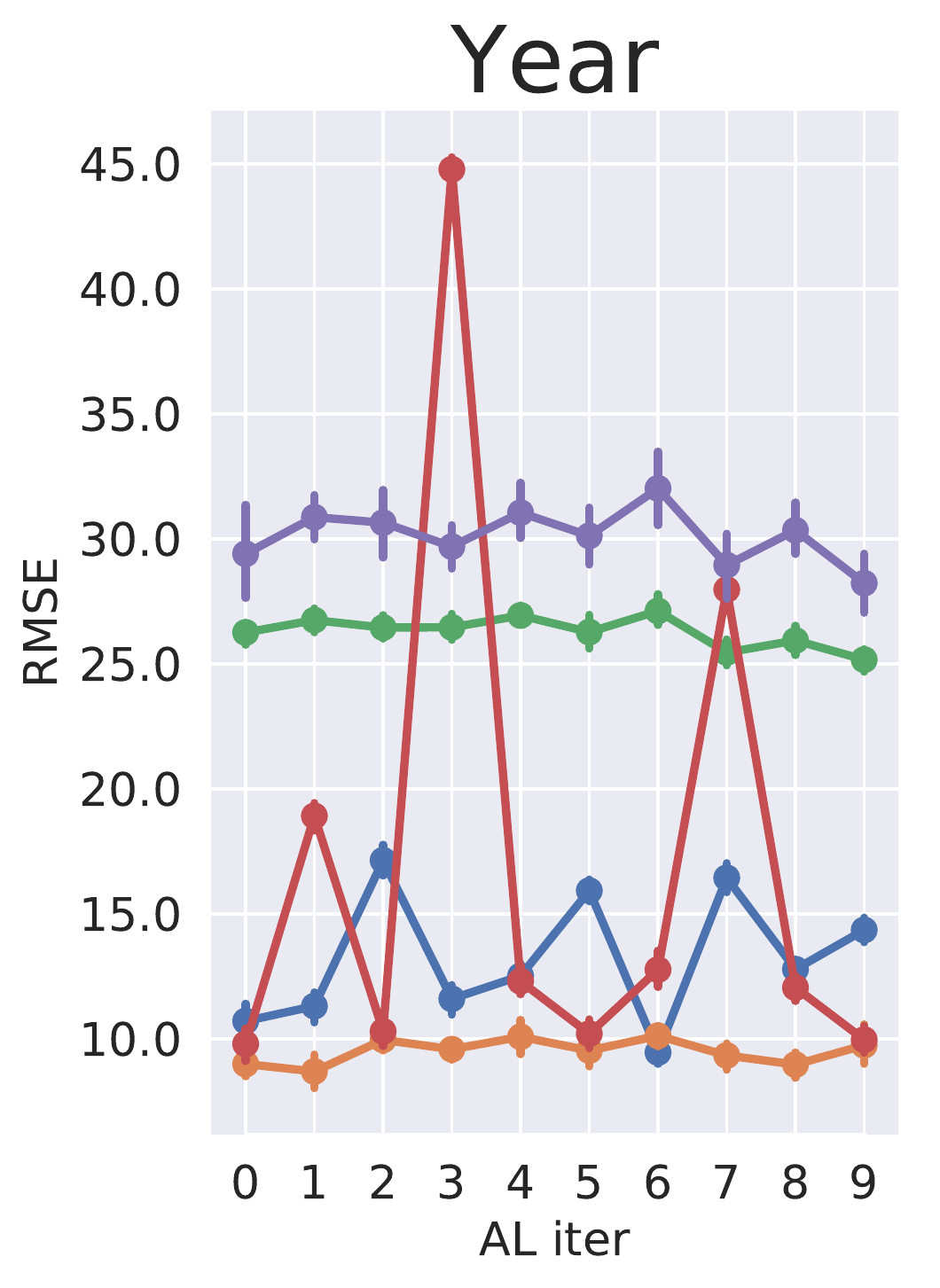}}
\raisebox{0.3\height}{\subfloat{\includegraphics[width=0.14\textwidth]{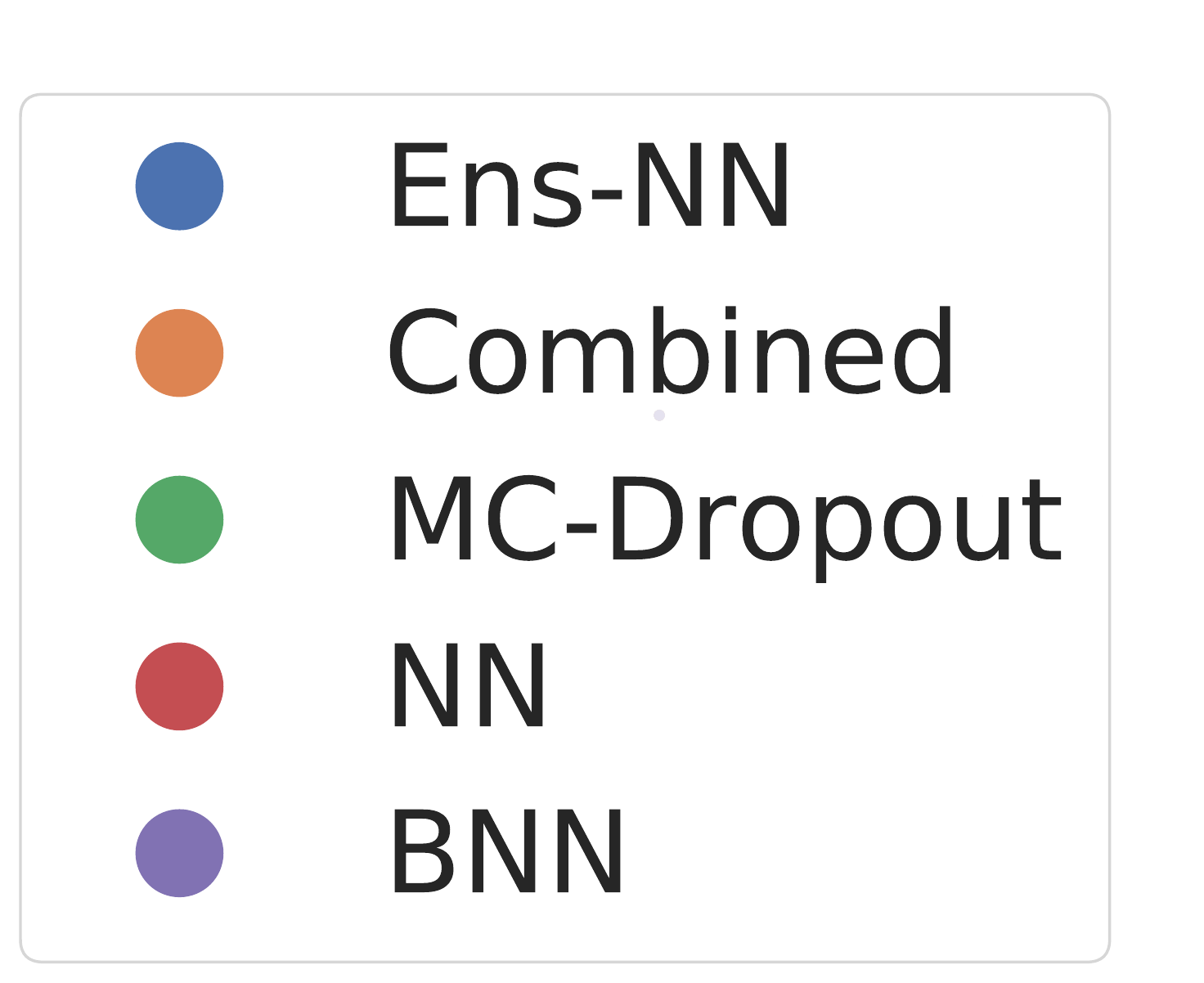}}}
\caption{Average test RMSE and standard errors in the active learning experiments for all datasets.}
\label{fig:al_rmse}
\end{figure}

\begin{figure}[h!]
\subfloat{\includegraphics[width=0.14\textwidth]{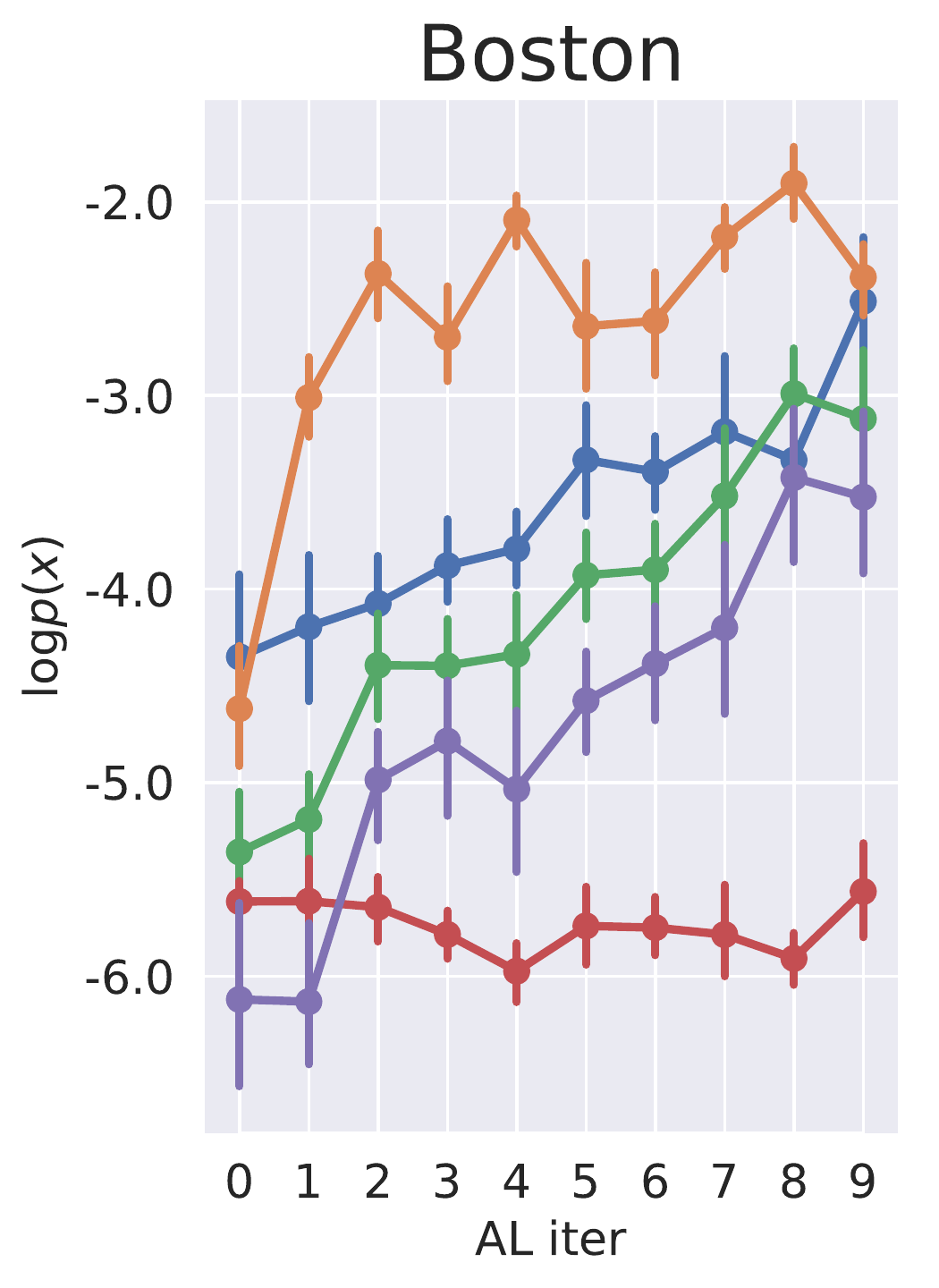}}
\subfloat{\includegraphics[width=0.14\textwidth]{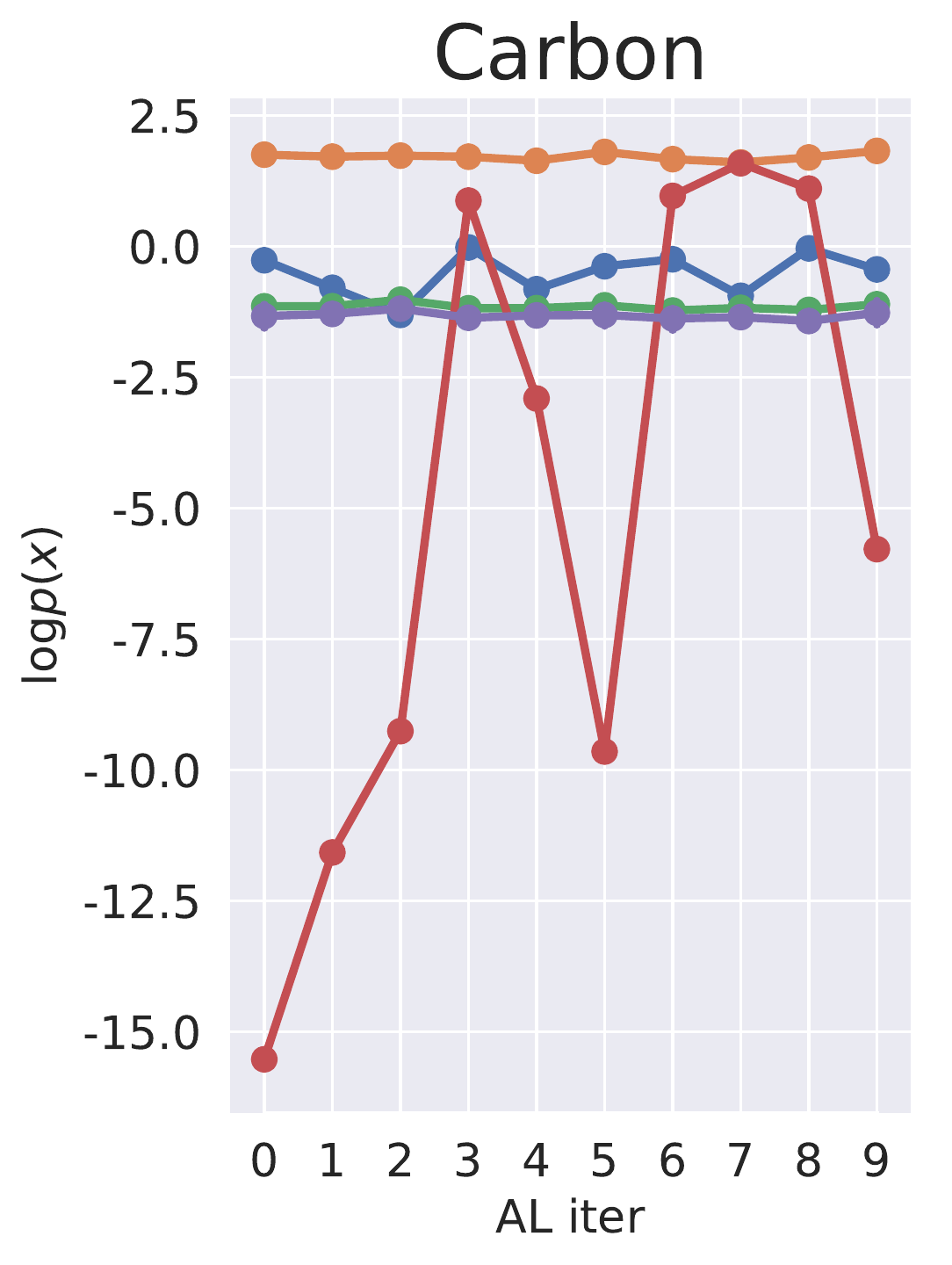}}
\subfloat{\includegraphics[width=0.14\textwidth]{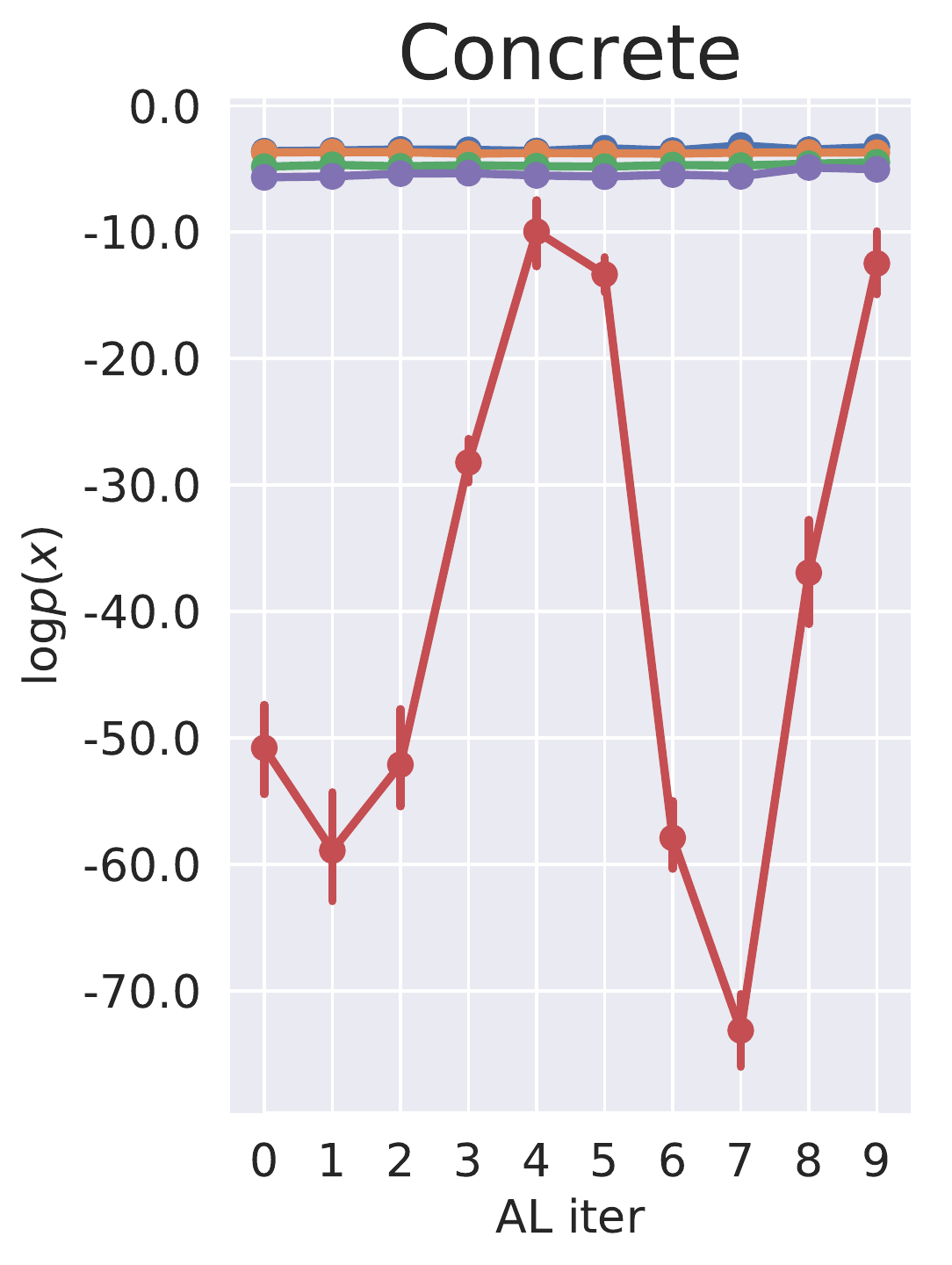}}
\subfloat{\includegraphics[width=0.14\textwidth]{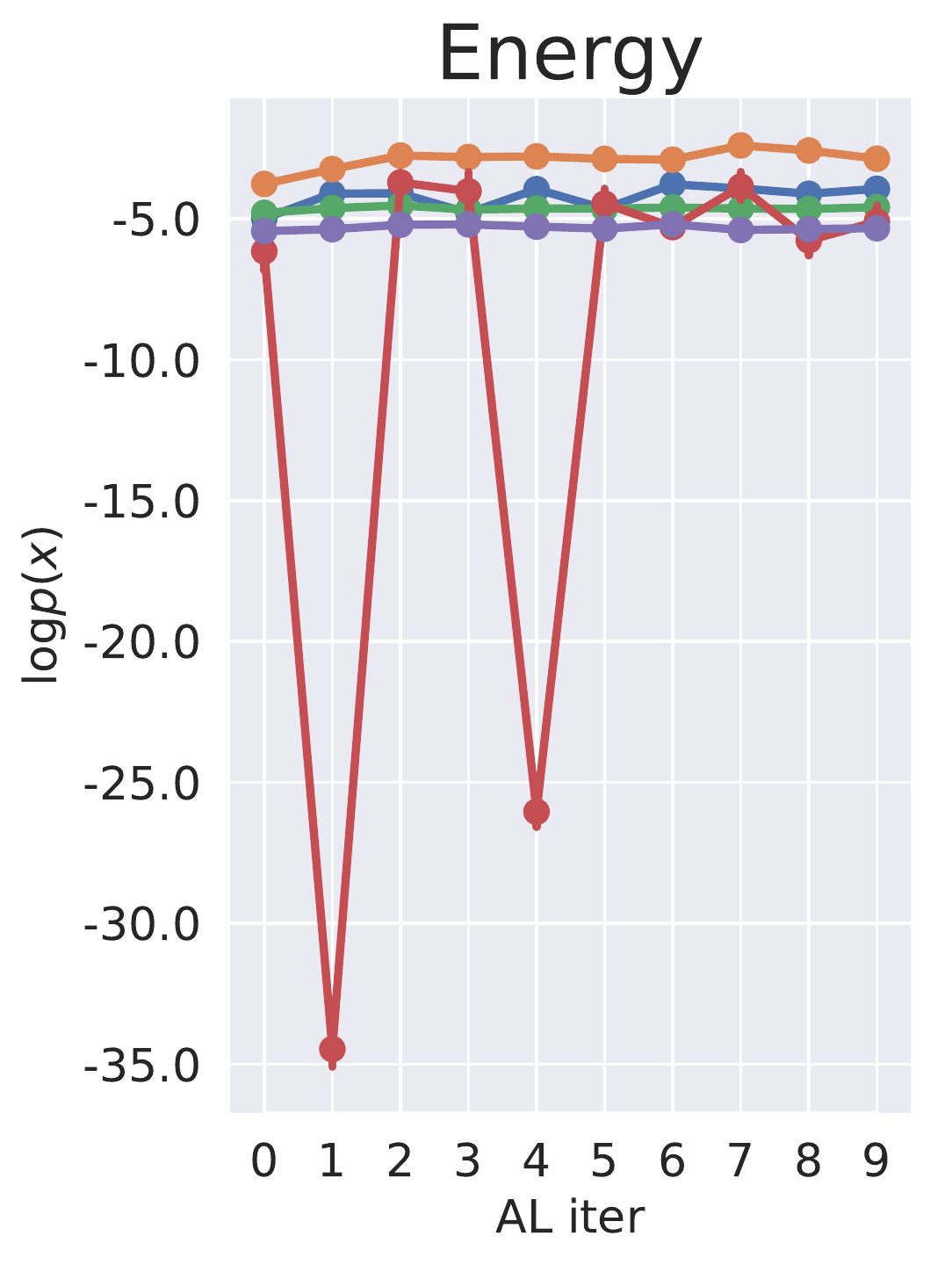}}
\subfloat{\includegraphics[width=0.14\textwidth]{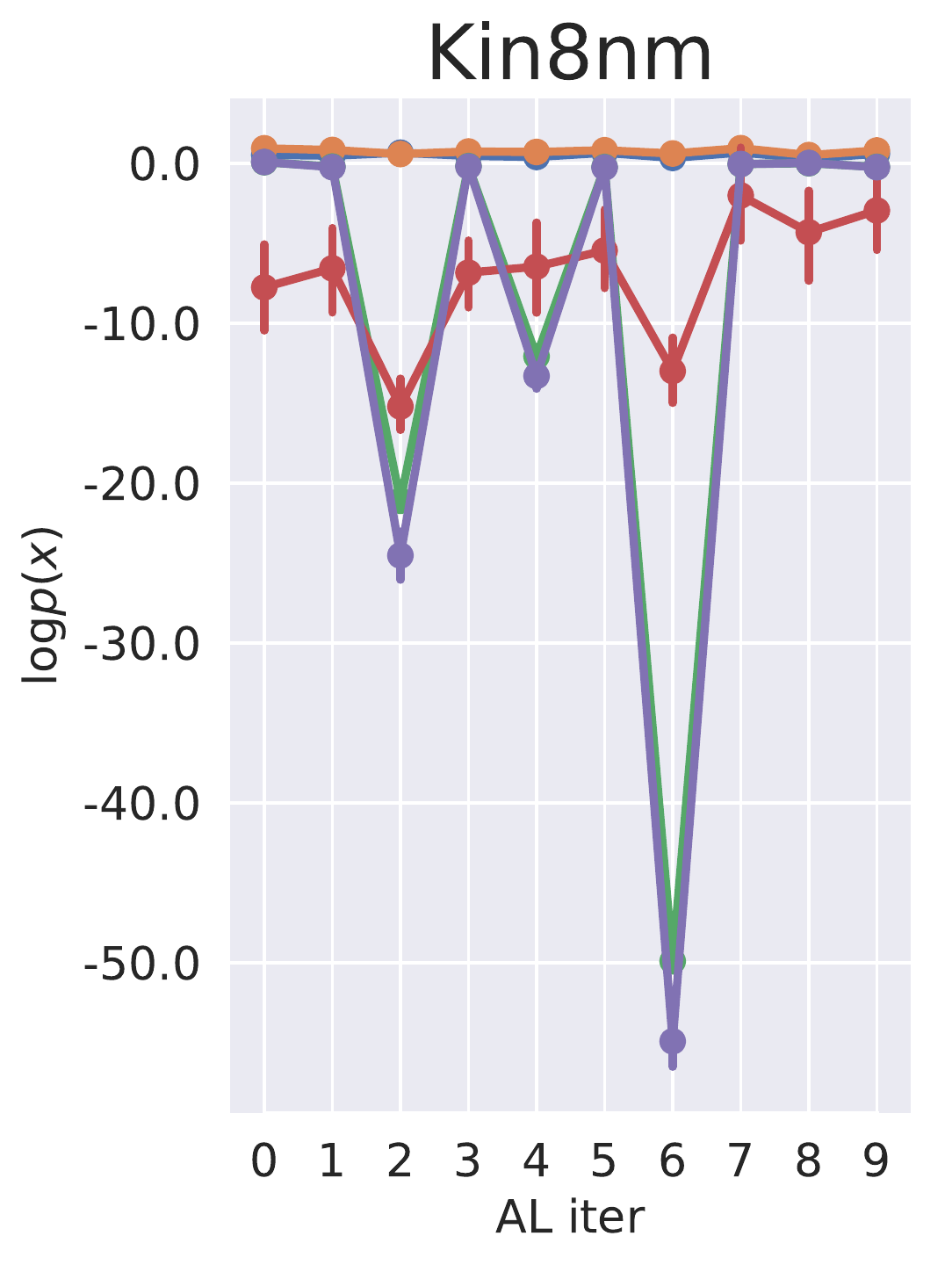}}
\subfloat{\includegraphics[width=0.14\textwidth]{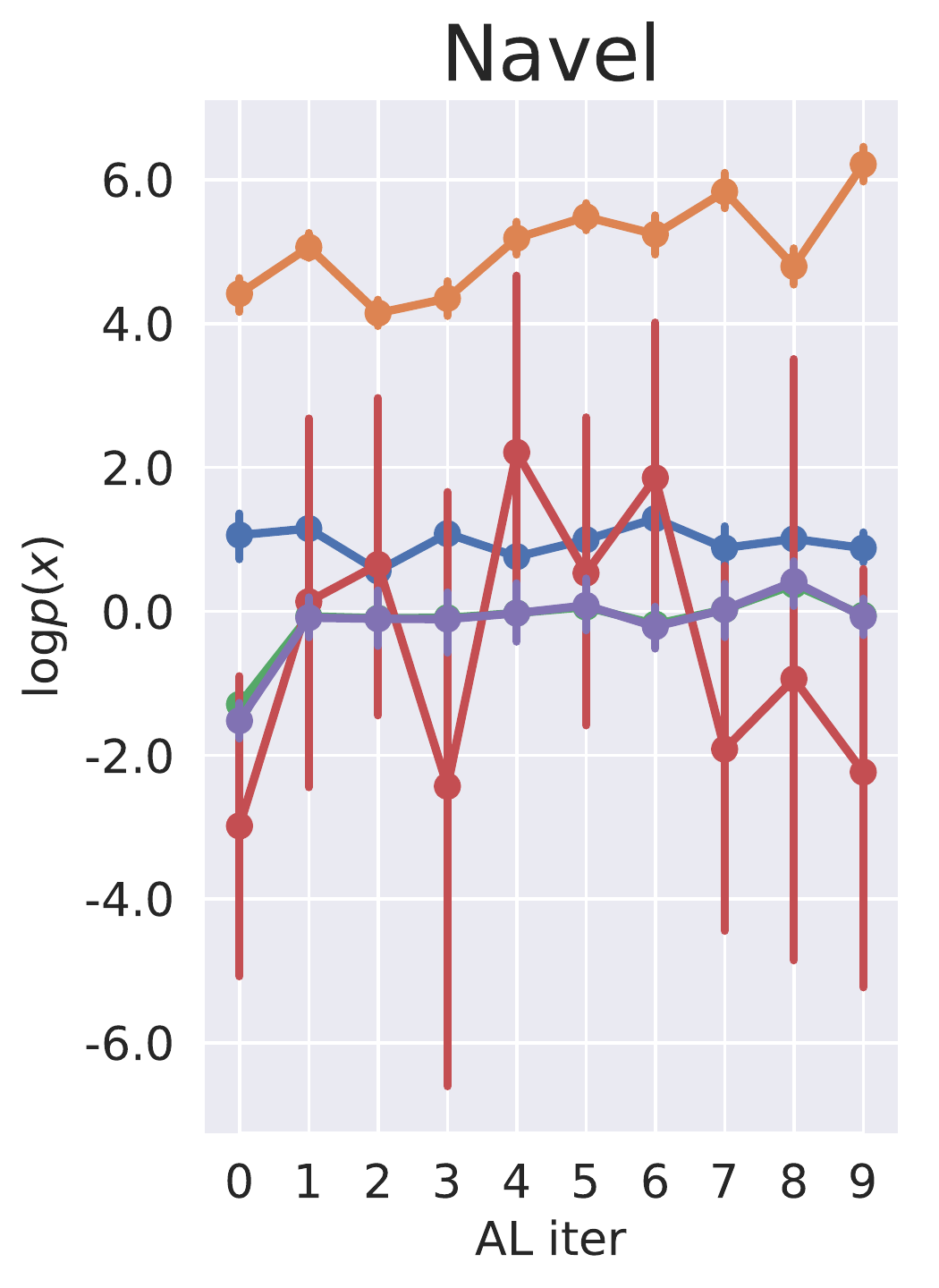}}
\subfloat{\includegraphics[width=0.14\textwidth]{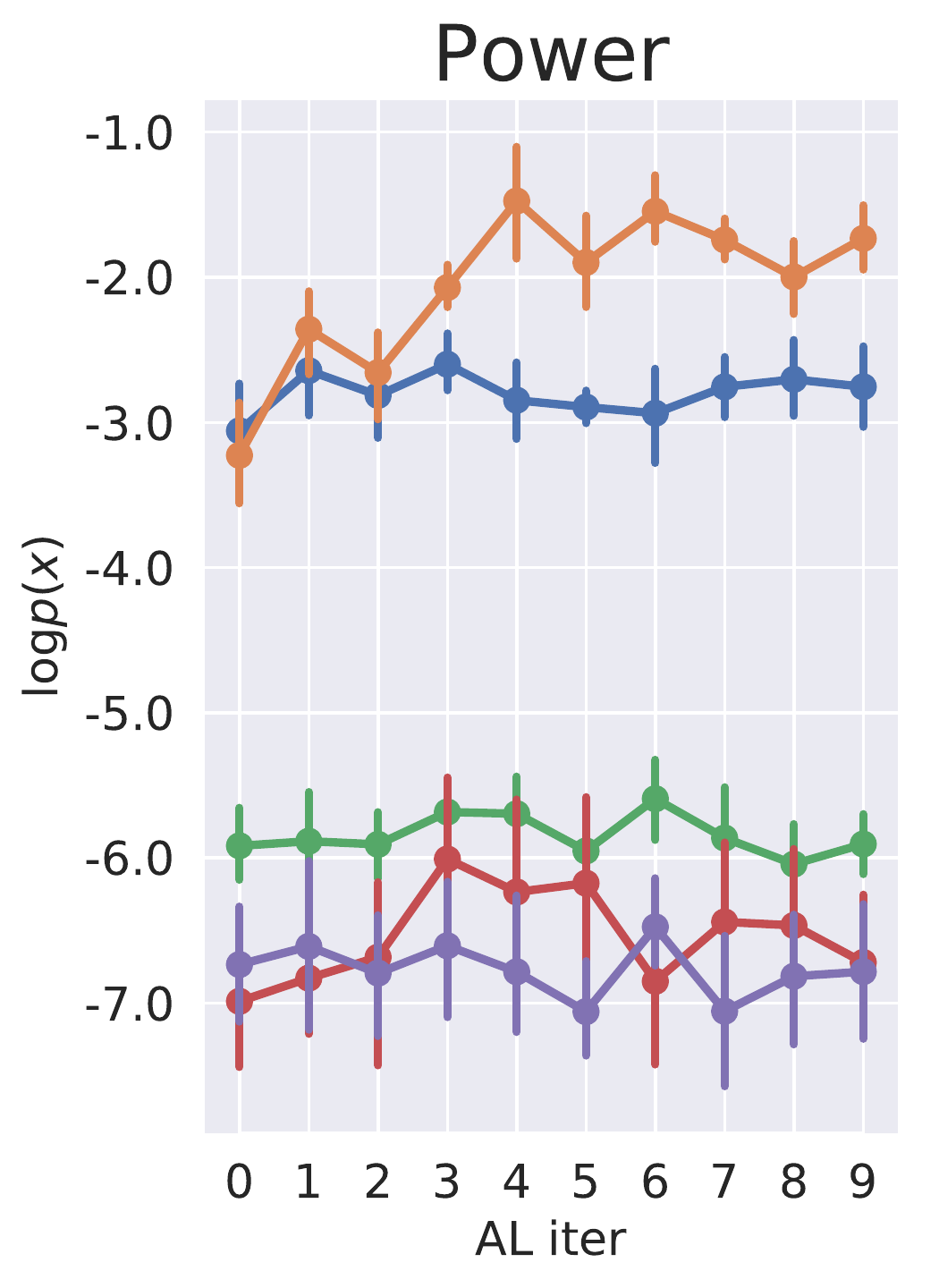}}

\subfloat{\includegraphics[width=0.14\textwidth]{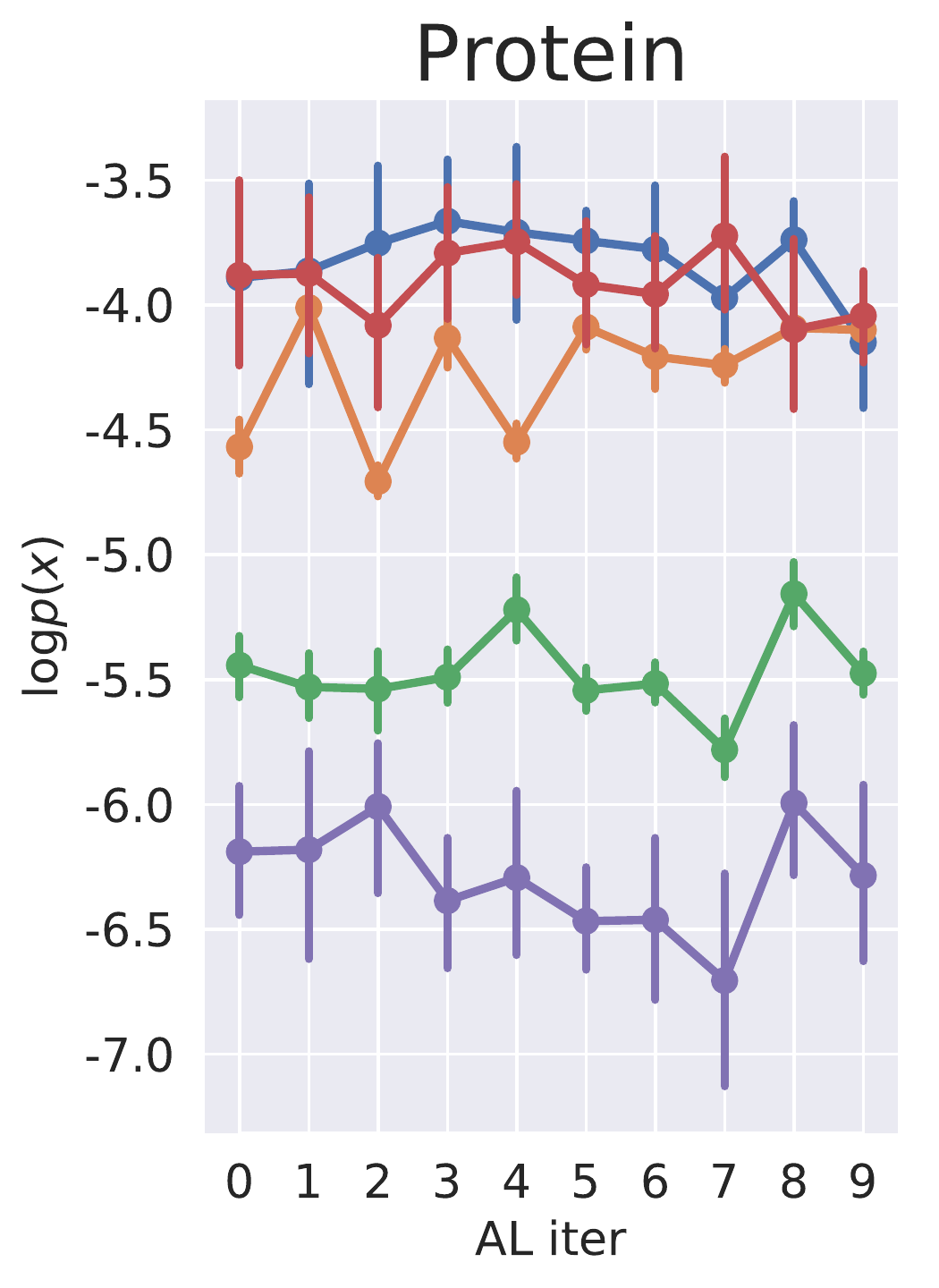}}
\subfloat{\includegraphics[width=0.14\textwidth]{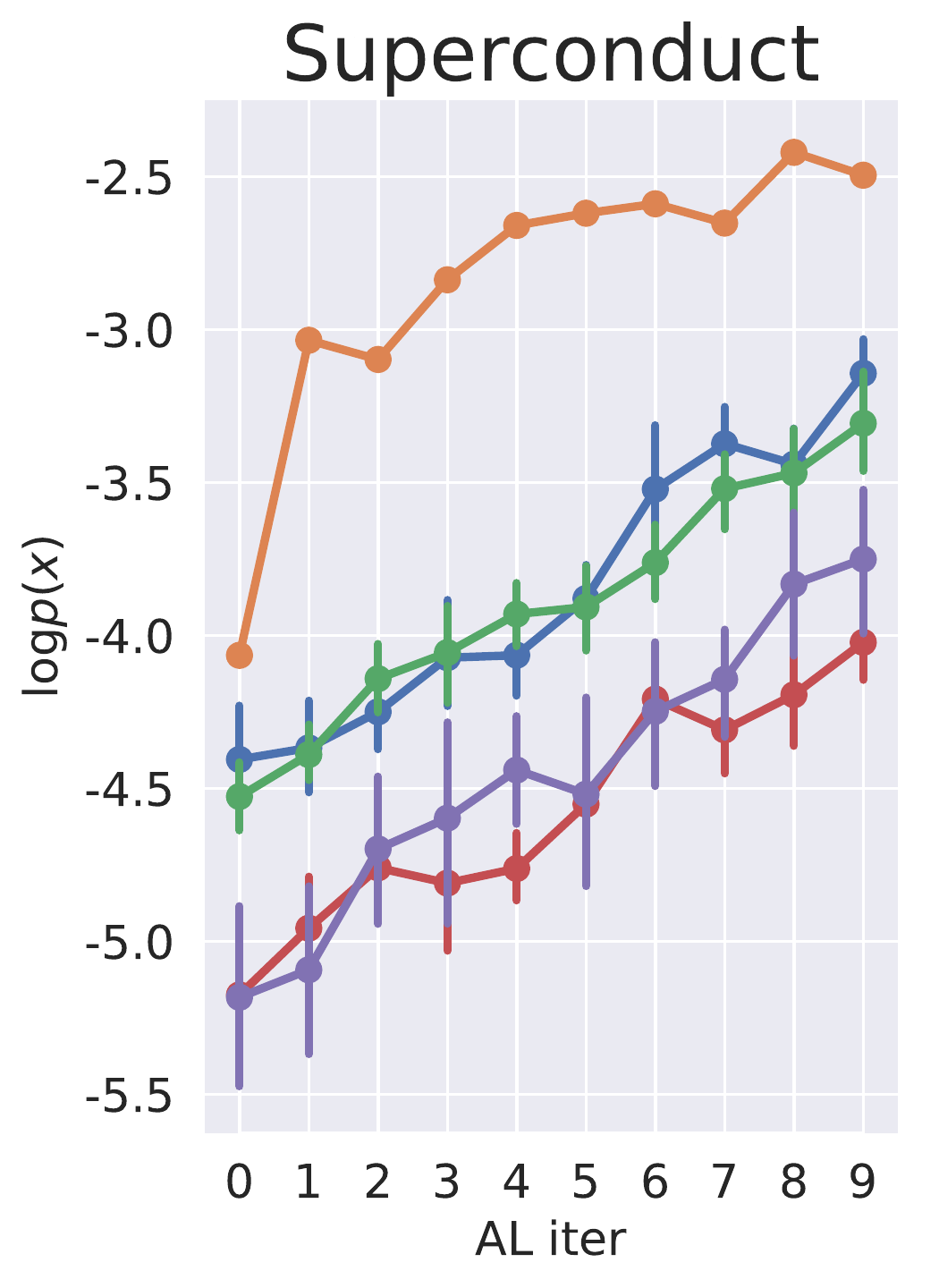}}
\subfloat{\includegraphics[width=0.14\textwidth]{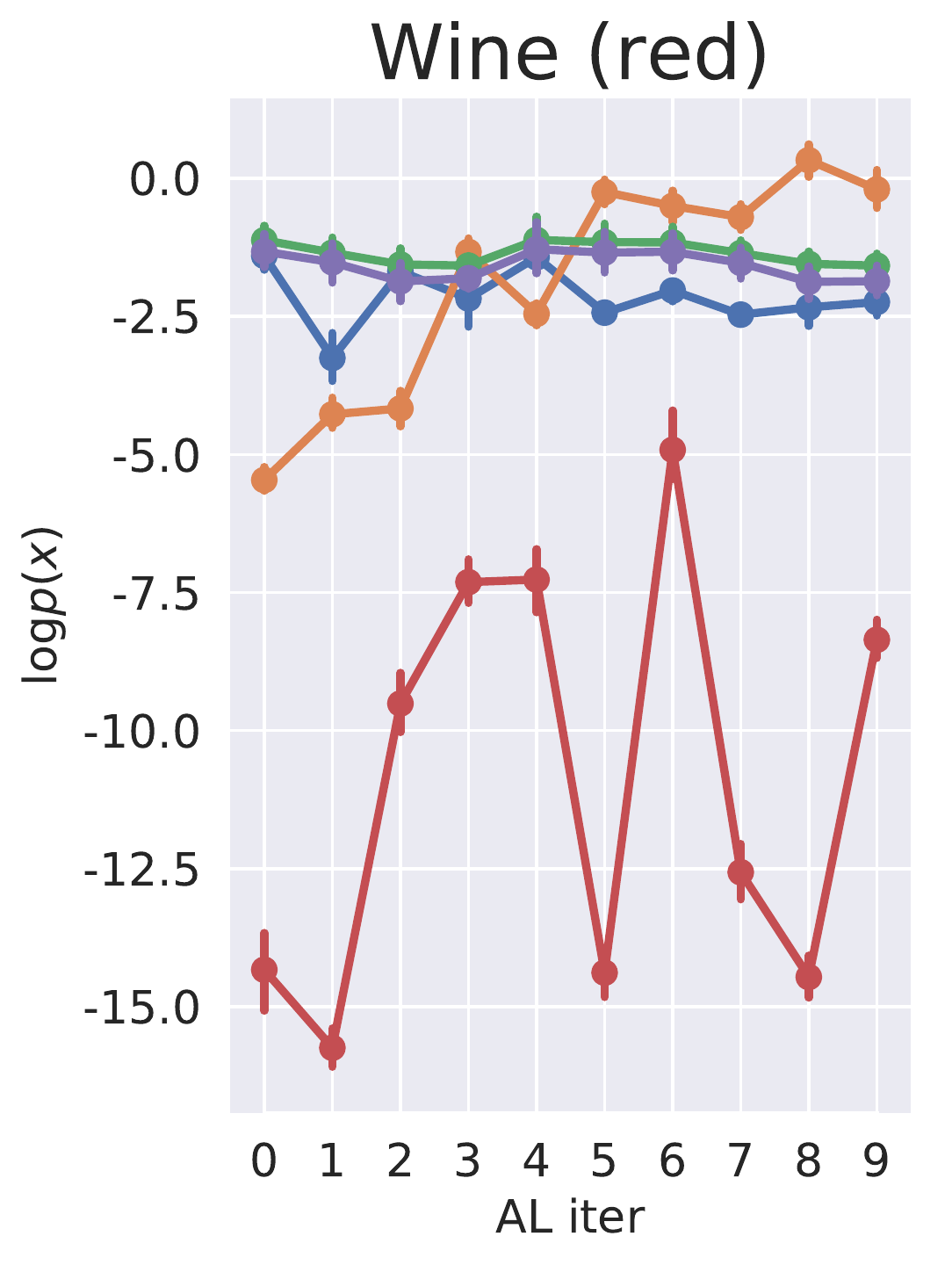}}
\subfloat{\includegraphics[width=0.14\textwidth]{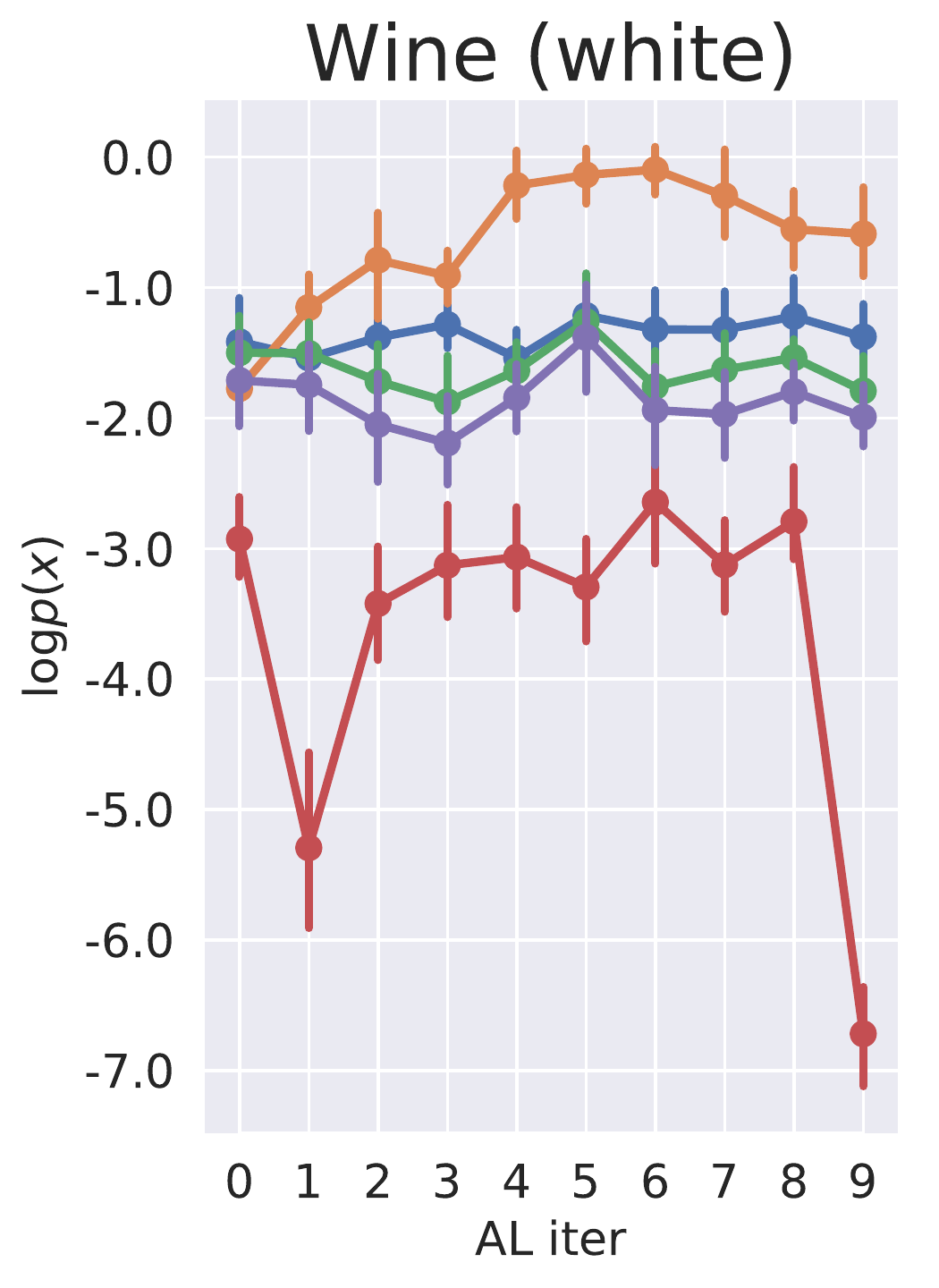}}
\subfloat{\includegraphics[width=0.14\textwidth]{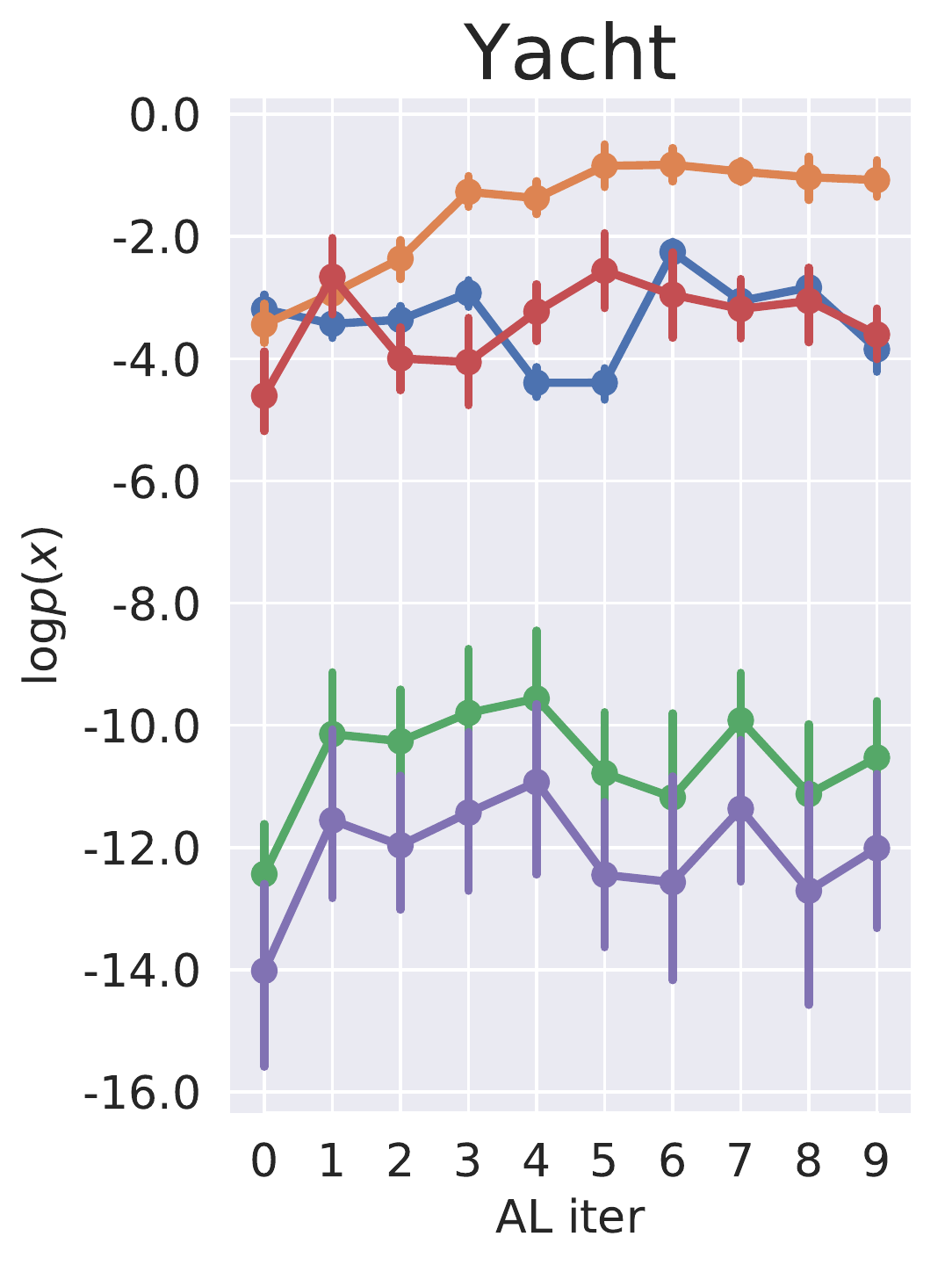}}
\subfloat{\includegraphics[width=0.14\textwidth]{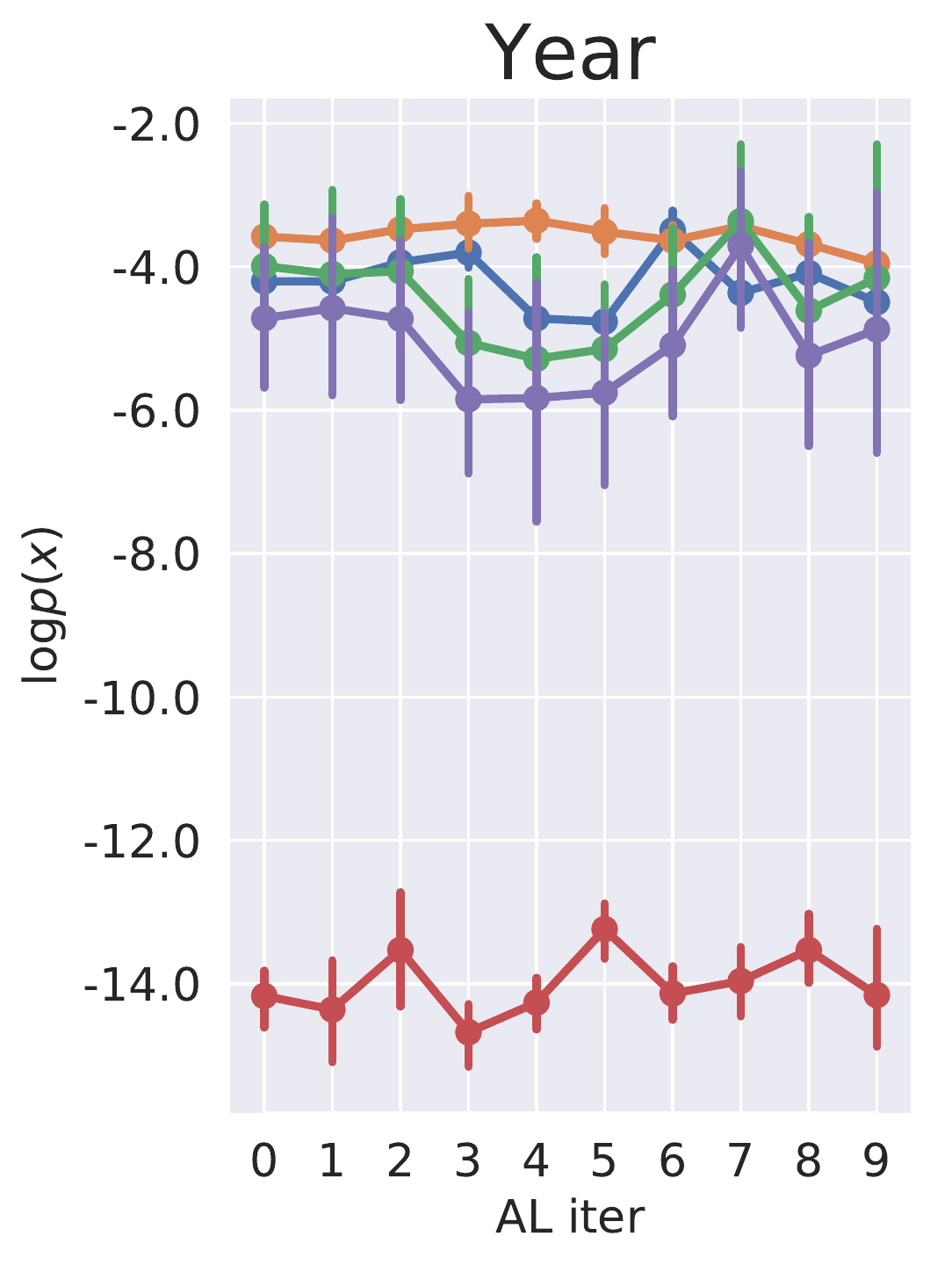}}
\raisebox{0.3\height}{\subfloat{\includegraphics[width=0.14\textwidth]{active_learning_labels.pdf}}}
\caption{Average test log likelihood and standard errors in the active learning experiments for all datasets.}
\label{fig:al_logpx}
\end{figure}

In \FIGS\ref{fig:al_rmse} and \ref{fig:al_logpx} we respective shows the progress of RMSE and log likelihood on all 13 dataset. We observe that for some of the datasets (Boston, Superconduct, Power) our proposed Combine model achieves faster learning than other methods. On all other datasets we are equally good as the best performing model. 

\paragraph{Generative modeling toy data}

\begin{figure}[]
\centering
\subfloat[Ground true]{\includegraphics[width=0.49\textwidth]{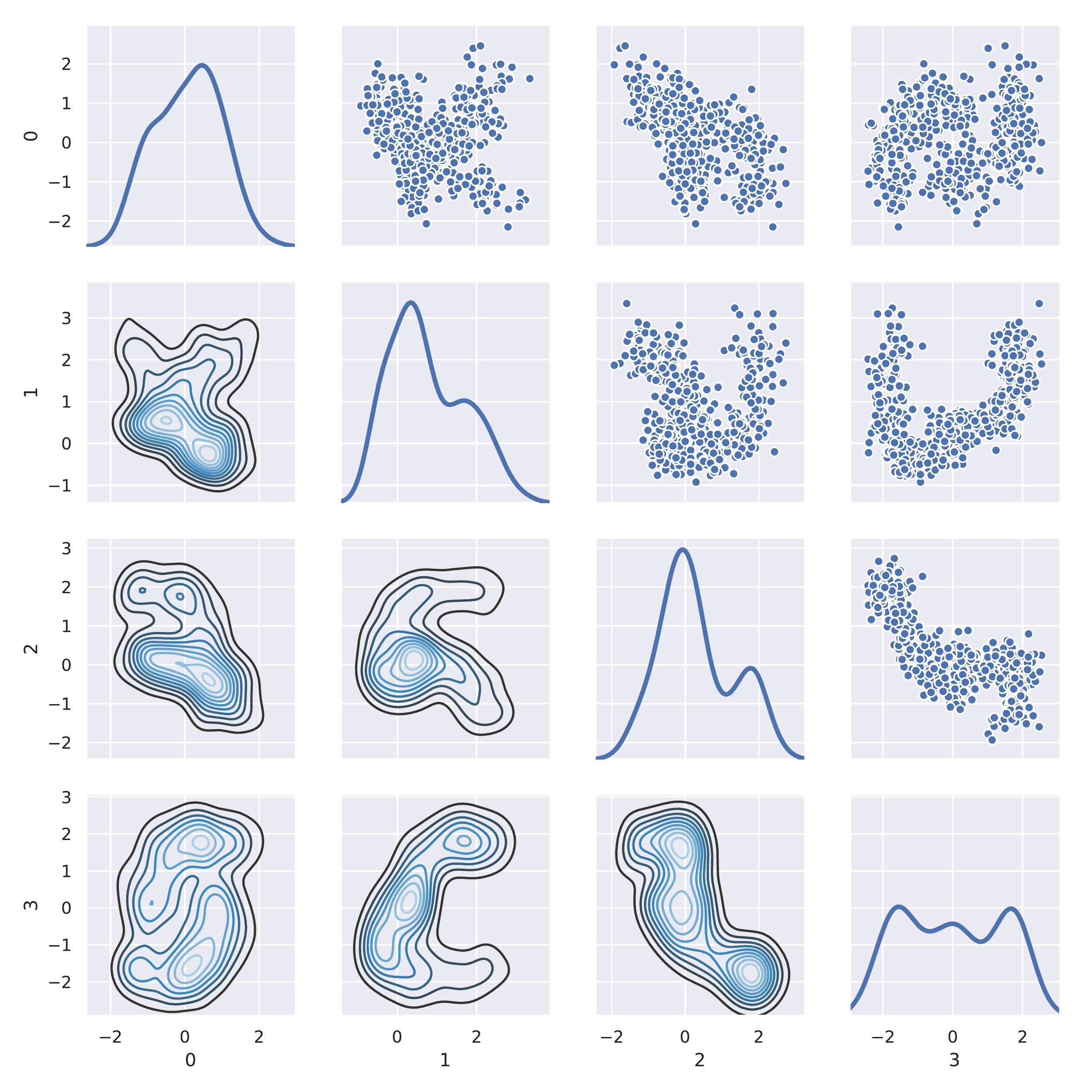}}

\subfloat[VAE (R)]{\includegraphics[width=0.49\textwidth]{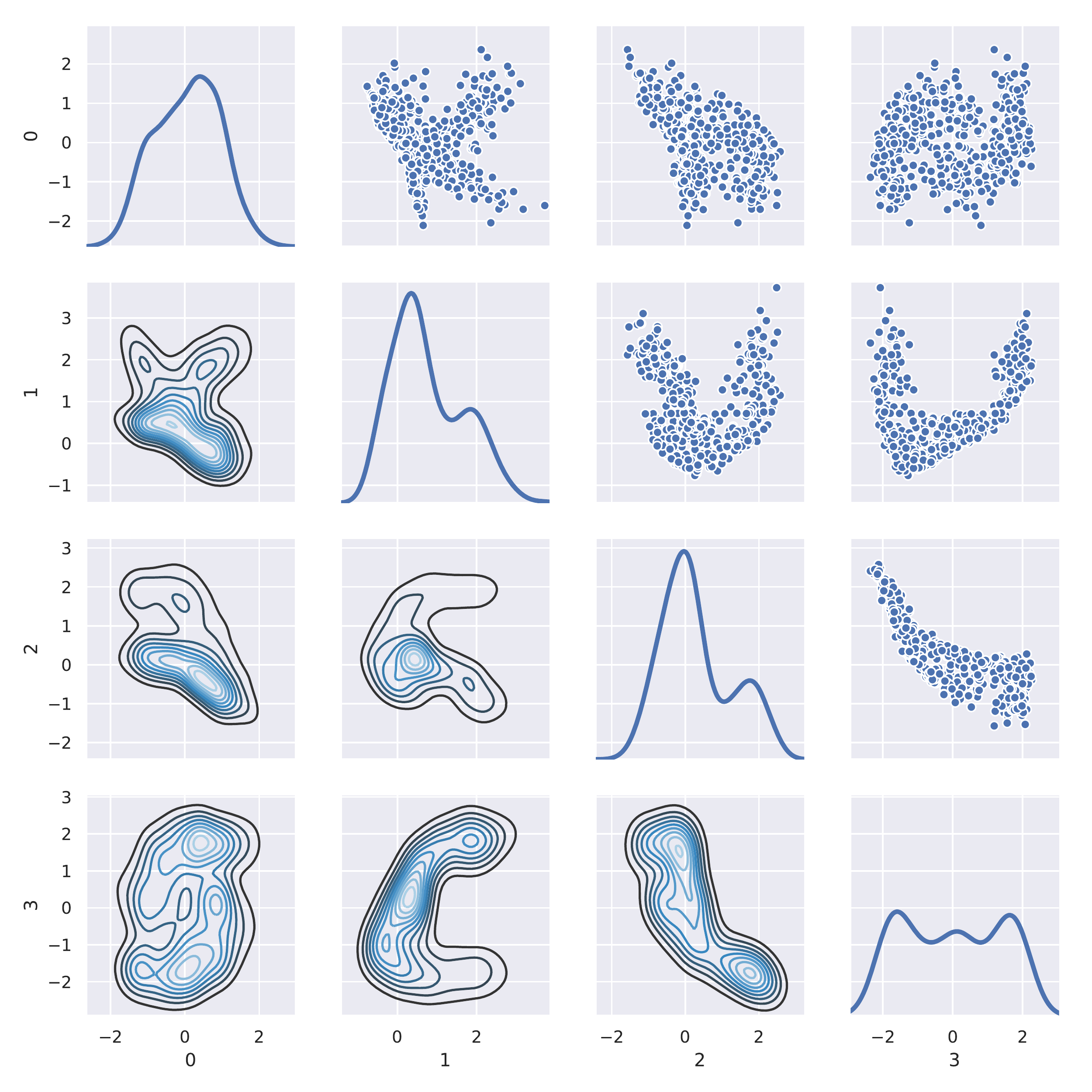}}
\subfloat[VAE (S)]{\includegraphics[width=0.49\textwidth]{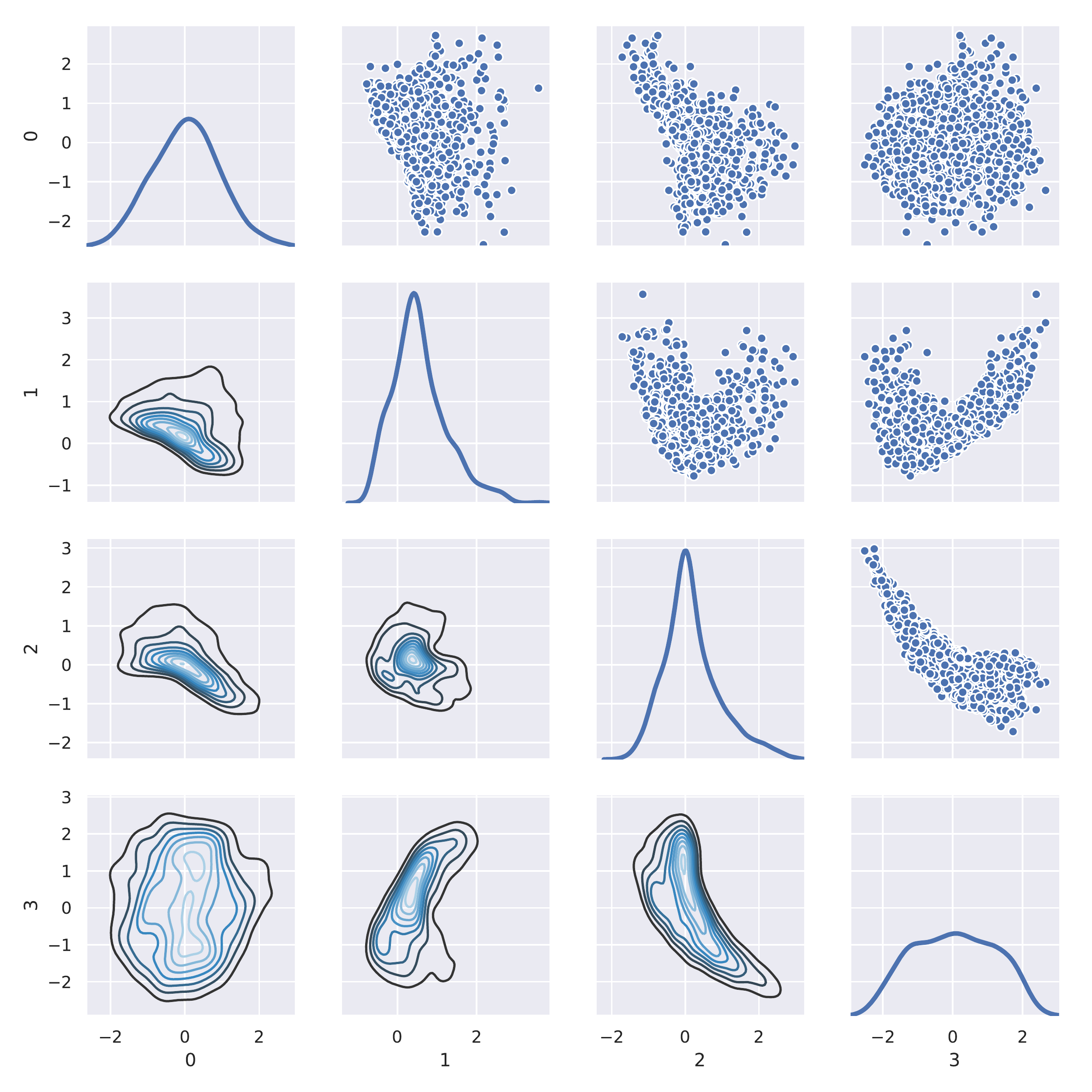}}

\subfloat[Comb-VAE (R)]{\includegraphics[width=0.49\textwidth]{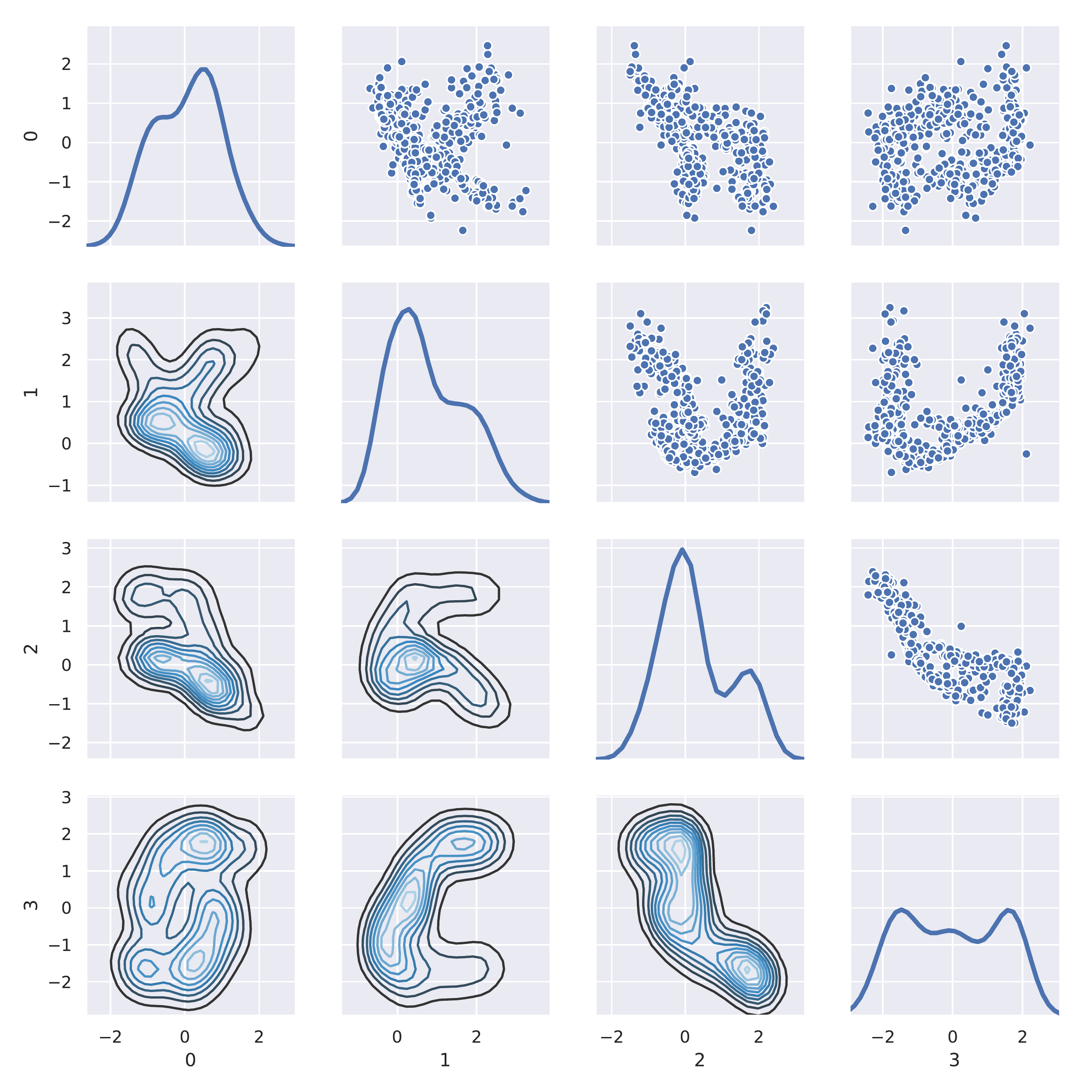}}
\subfloat[Comb-VAE (S)]{\includegraphics[width=0.49\textwidth]{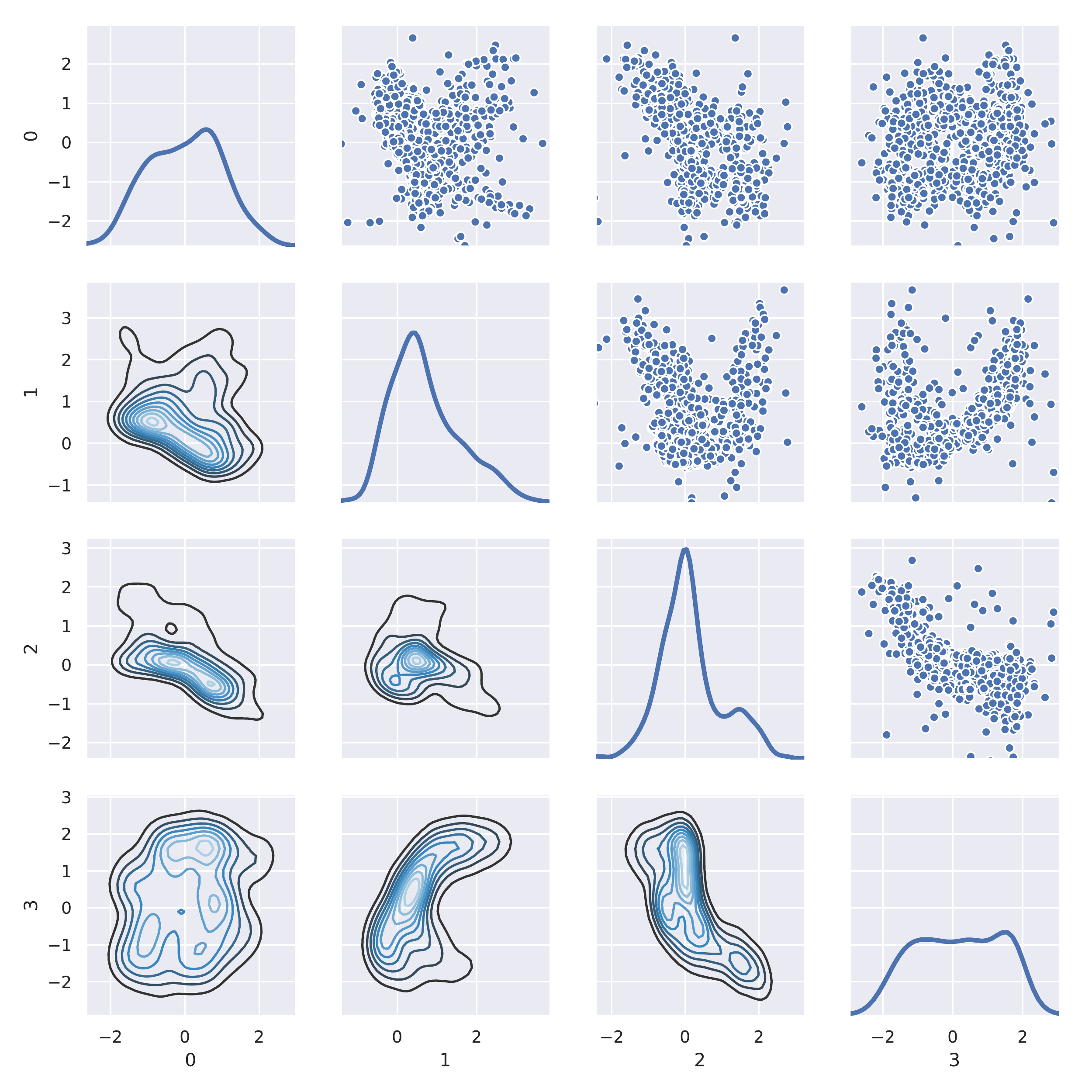}}
\caption{Pairwise plots between all sets of variables for our artificially dataset in the generative setting. }
\label{fig:generative_plot}
\end{figure}

In \FIG\ref{fig:generative_plot} we show marginal distribution, pairwise pointplots and pairwise joint distribution for our artificially dataset used in the generative setting. Top row show the ground true data, middle row shows reconstructions and samples from standard VAE model and bottom row show reconstructions and samples from our proposed Comb-VAE model. We observe that reconstruction are similar for the two models, but the quality of the generative samples are much better for Comb-VAE.

\paragraph{Generative modeling of image data}

\begin{figure}[h!]
\centering
\subfloat[VAE]{\includegraphics[width=0.49\textwidth]{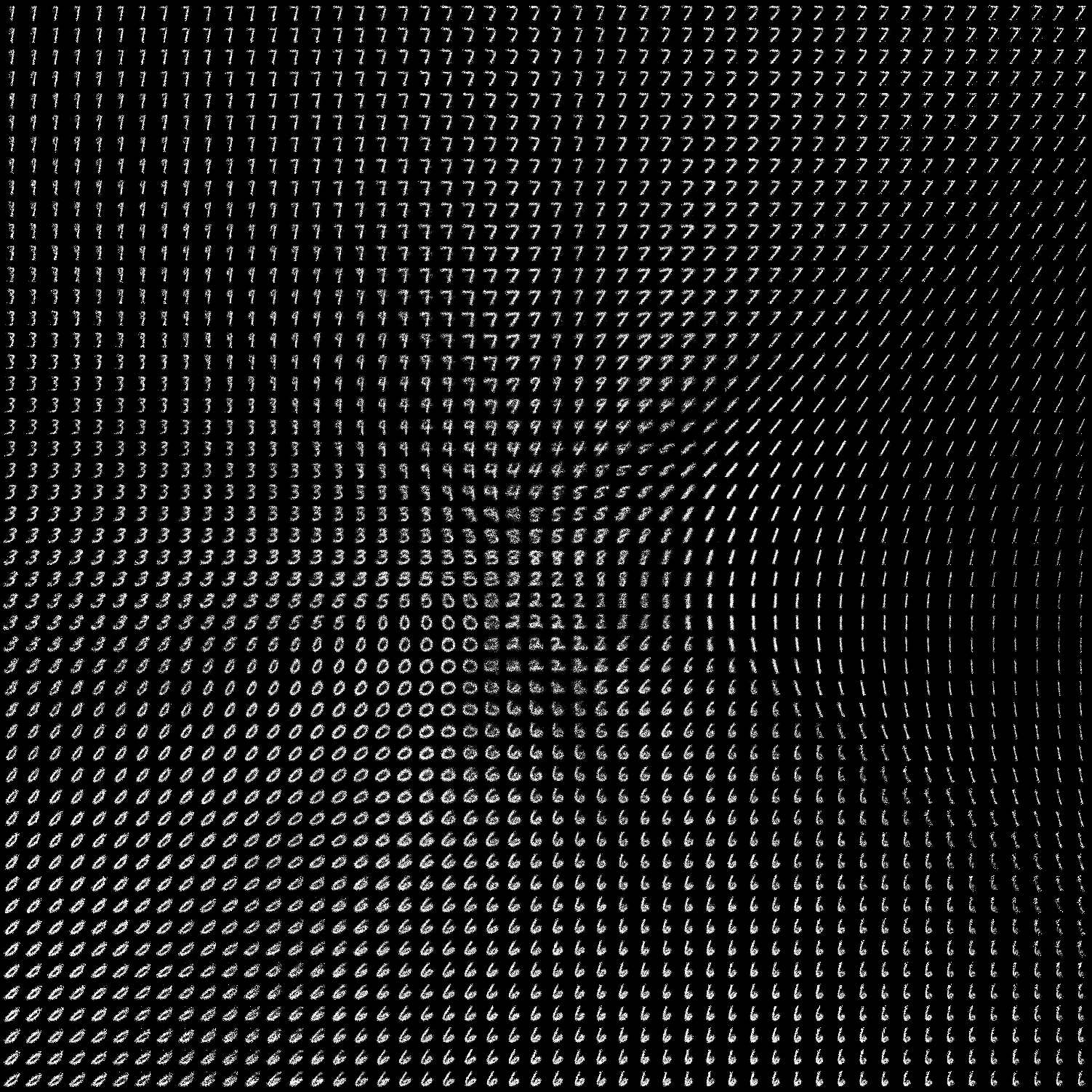}}
\subfloat[Comb-VAE]{\includegraphics[width=0.49\textwidth]{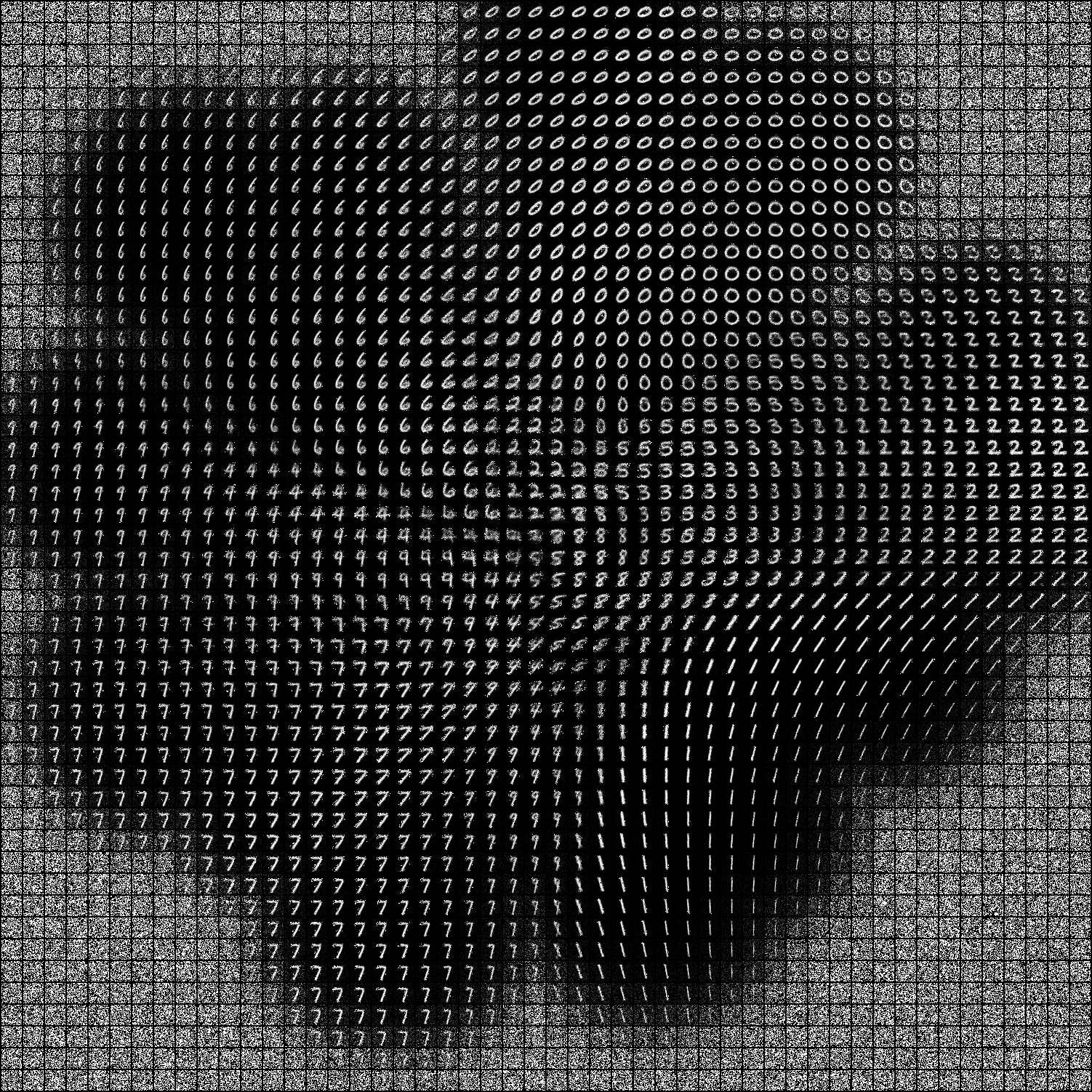}}
\caption{Meshgrid of latent space. For each subplot we sampled a mesh grid $[-4,4]\times[-4,4]$ of $[50,50]$ points, which we used to generate samples from. }
\label{fig:mnist_grid}
\end{figure}

In \FIG\ref{fig:mnist_grid} we show a meshgrid of samples from VAE and Comb-VAE on the MNIST dataset. We clearly see how proper extrapolation of variance in Comb-VAE can be used to "mask" when we are inside our data region and when we are outside. For standard VAE we observe a near constant variance added to the images. 

\section{On the parametrization of the $t$-distributed predictive distribution}
We parametrize the Student-$t$ distribution by letting the variance $\sigma^2$ have an inverse-Gamma distribution with shape and scale parameters $\alpha$ and $\beta$. We use that if $\sigma^2\sim\textsc{Inv-Gamma}(\alpha,\beta)$ then $\frac{1}{\sigma^2}\sim\Gamma(\alpha,\beta)$. Then 
\begin{align*}
	p(y|\mu,\alpha,\beta)&=\int_{0}^{\infty}\mathcal{N}(y|\mu,\sigma^2)\frac{\beta^\alpha}{\Gamma(\alpha)}(\sigma^2)^{-(\alpha+1)}\exp(-\frac{\beta}{\sigma^2})\text{d}\sigma^2\\
	&=\frac{\beta^\alpha}{\Gamma(\alpha)\sqrt{2\pi}}\int_{0}^\infty(\sigma^2)^{-(\alpha+1)-\frac{1}{2}}\exp(-\frac{1}{\sigma^2}(\beta + \frac{1}{2}(y-\mu)^2))\text{d}\sigma^2\\
	&=\frac{\beta^\alpha}{\Gamma(\alpha)\sqrt{2\pi}}\frac{\Gamma(\alpha+\frac{1}{2})}{\Big(\beta + \frac{1}{2}(y-\mu)^2\Big)^{\alpha+\frac{1}{2}}},
\end{align*}
where we substituted the variable $\sigma^2$ with $\frac{1}{\sigma^2}$ and used that the remaining was a Gamma integral.

\section{Parameters of the locality sampler}
\begin{figure}[h!]
\subfloat{\includegraphics[width=0.2\textwidth]{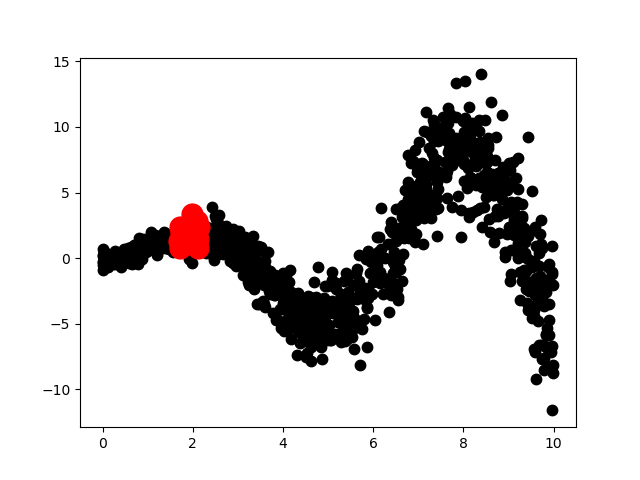}}
\subfloat{\includegraphics[width=0.2\textwidth]{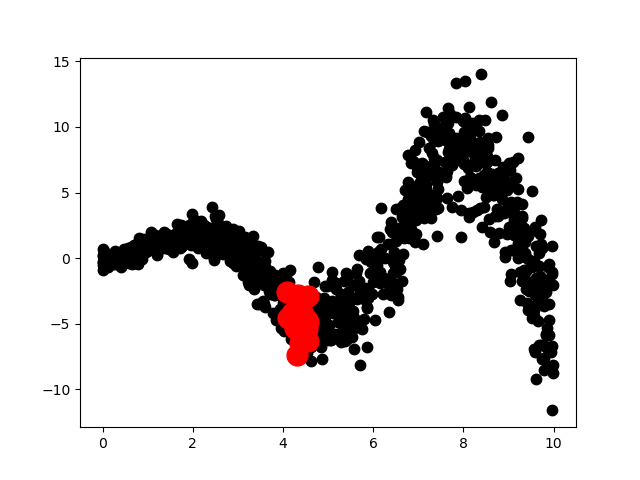}}
\subfloat{\includegraphics[width=0.2\textwidth]{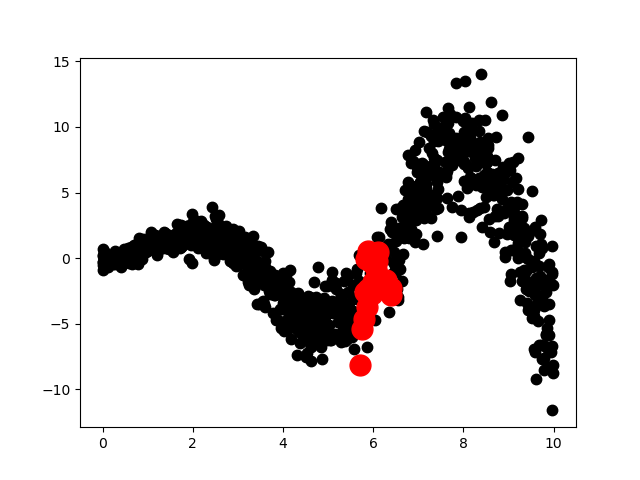}}
\subfloat{\includegraphics[width=0.2\textwidth]{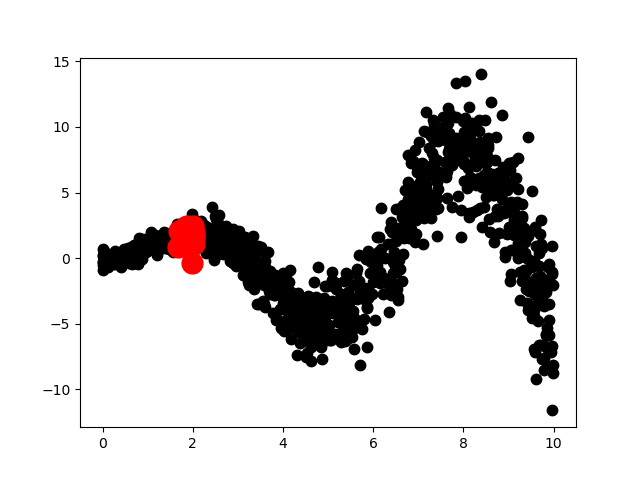}}
\subfloat{\includegraphics[width=0.2\textwidth]{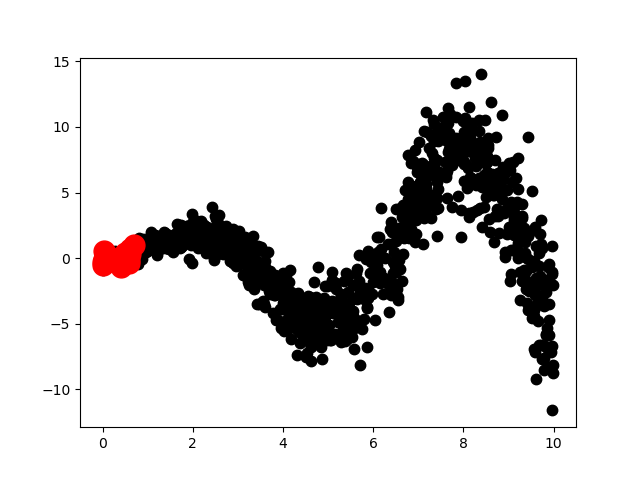}}
\vspace{-0.5cm}
\subfloat{\includegraphics[width=0.2\textwidth]{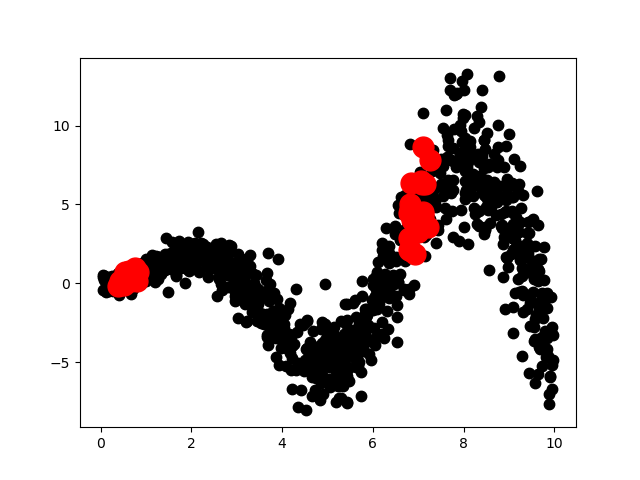}}
\subfloat{\includegraphics[width=0.2\textwidth]{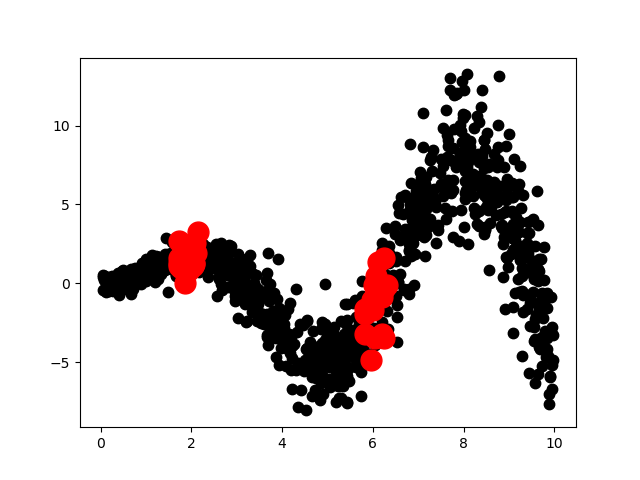}}
\subfloat{\includegraphics[width=0.2\textwidth]{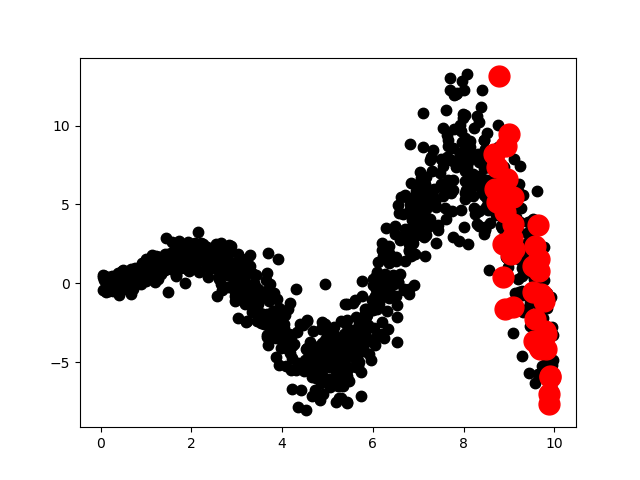}}
\subfloat{\includegraphics[width=0.2\textwidth]{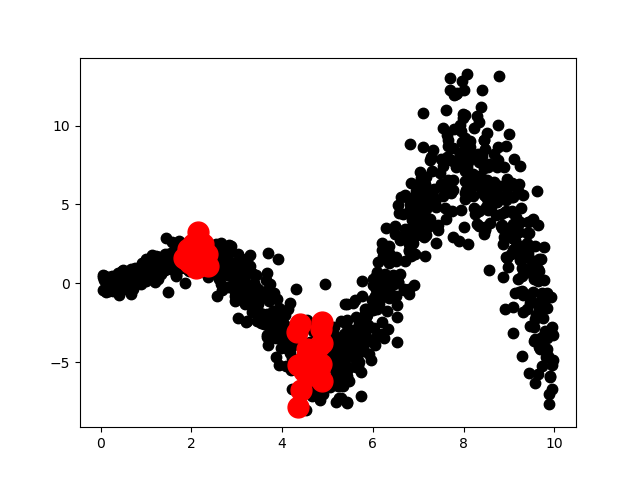}}
\subfloat{\includegraphics[width=0.2\textwidth]{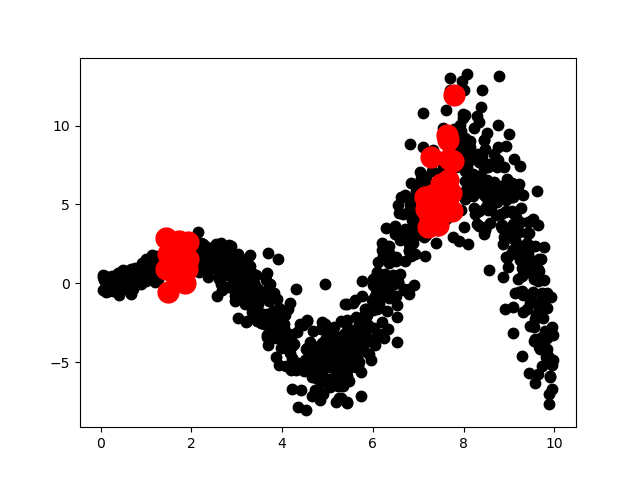}}
\vspace{-0.5cm}
\subfloat{\includegraphics[width=0.2\textwidth]{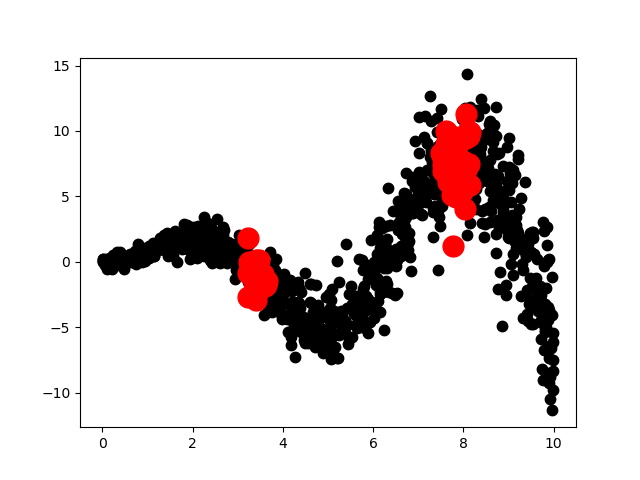}}
\subfloat{\includegraphics[width=0.2\textwidth]{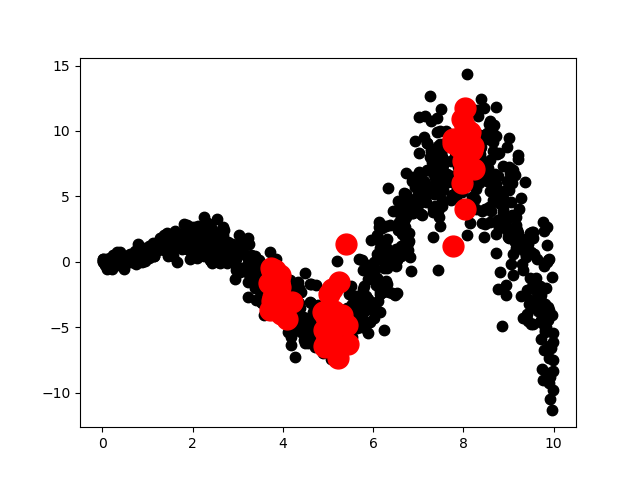}}
\subfloat{\includegraphics[width=0.2\textwidth]{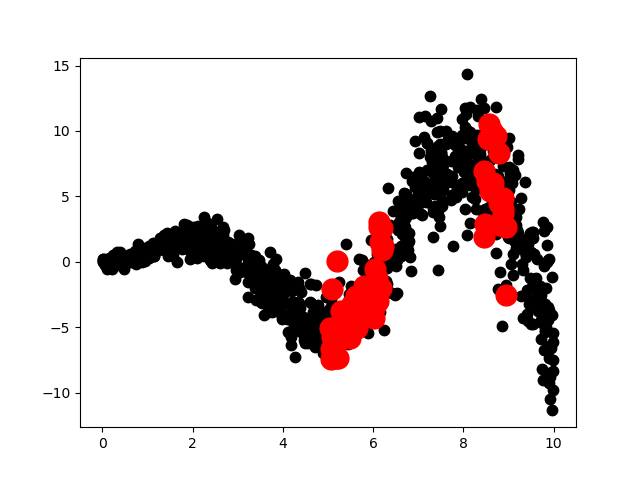}}
\subfloat{\includegraphics[width=0.2\textwidth]{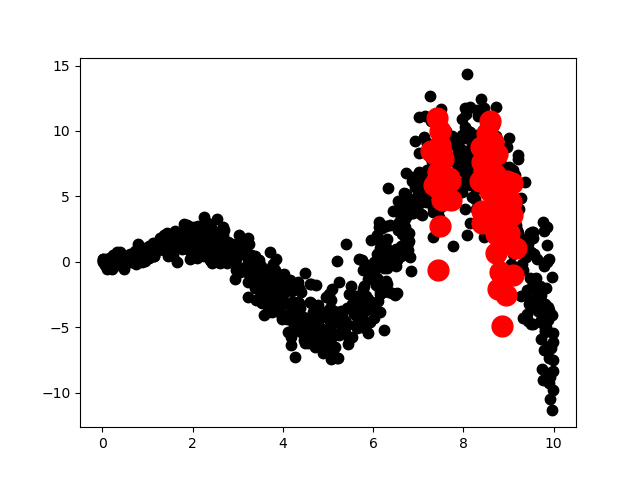}}
\subfloat{\includegraphics[width=0.2\textwidth]{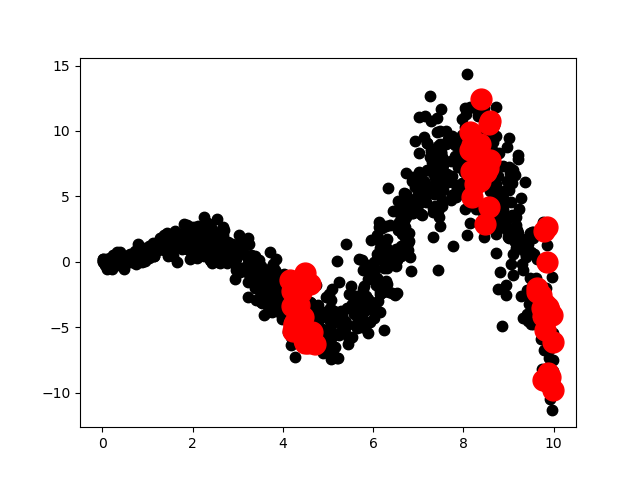}}
\caption{The effect of changing the size of the primary sampling unit. From top to bottom: $psu=[1,2,3]$. Each column corresponds to a sample from the locality sampler.}
\label{fig:psu}
\end{figure}

\begin{figure}[h!]
\subfloat{\includegraphics[width=0.2\textwidth]{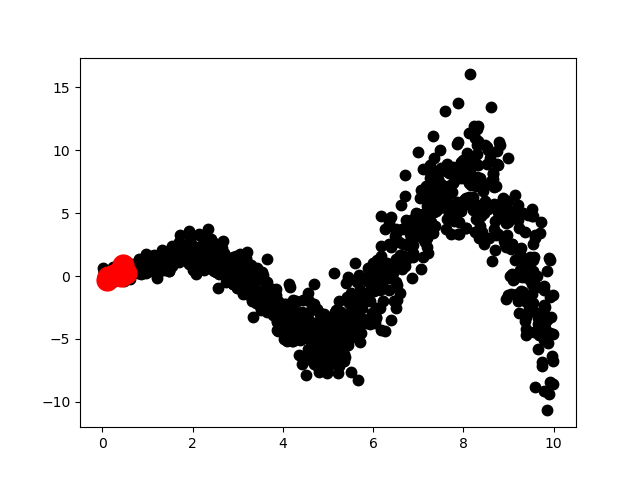}}
\subfloat{\includegraphics[width=0.2\textwidth]{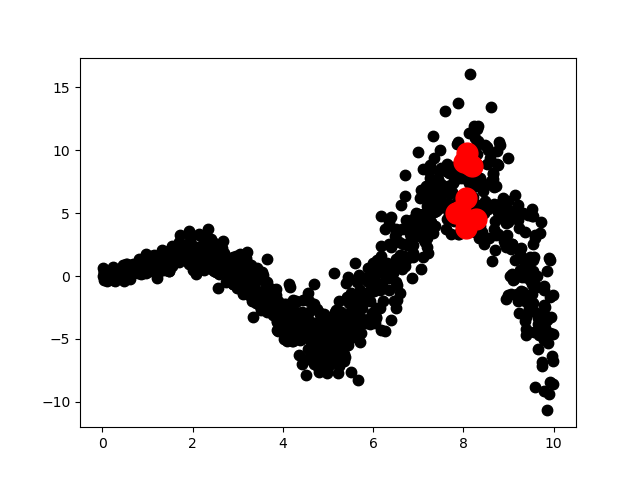}}
\subfloat{\includegraphics[width=0.2\textwidth]{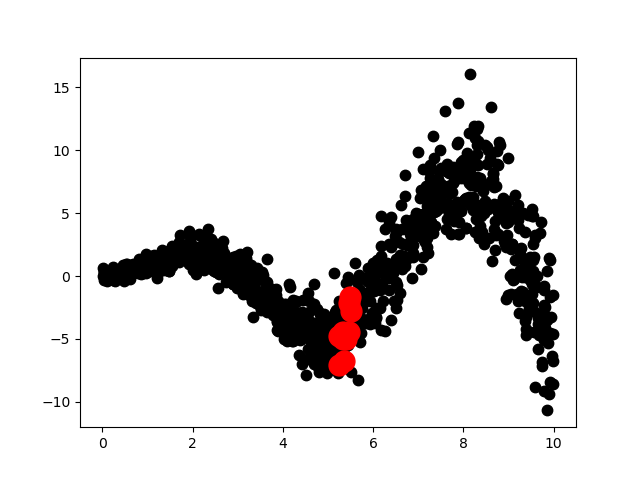}}
\subfloat{\includegraphics[width=0.2\textwidth]{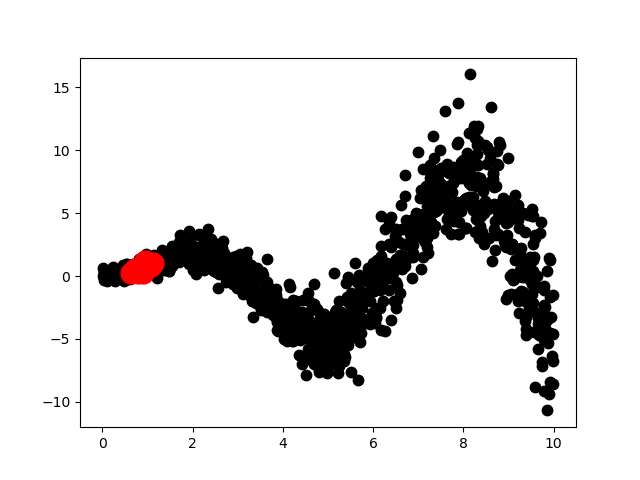}}
\subfloat{\includegraphics[width=0.2\textwidth]{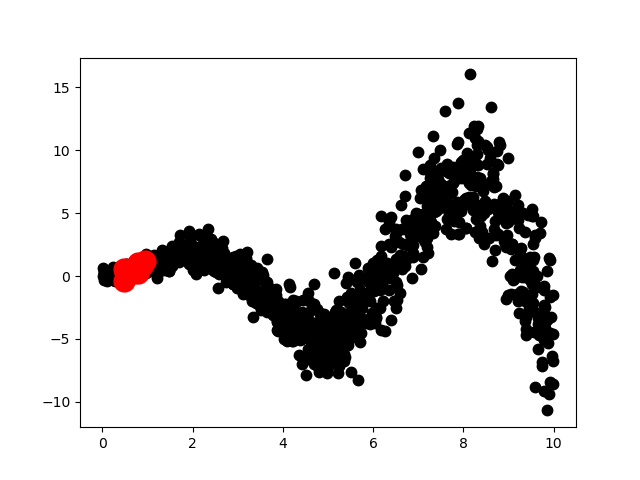}}
\vspace{-0.5cm}
\subfloat{\includegraphics[width=0.2\textwidth]{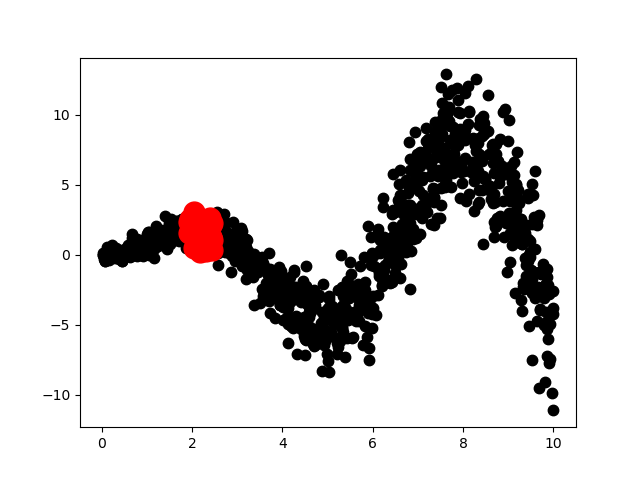}}
\subfloat{\includegraphics[width=0.2\textwidth]{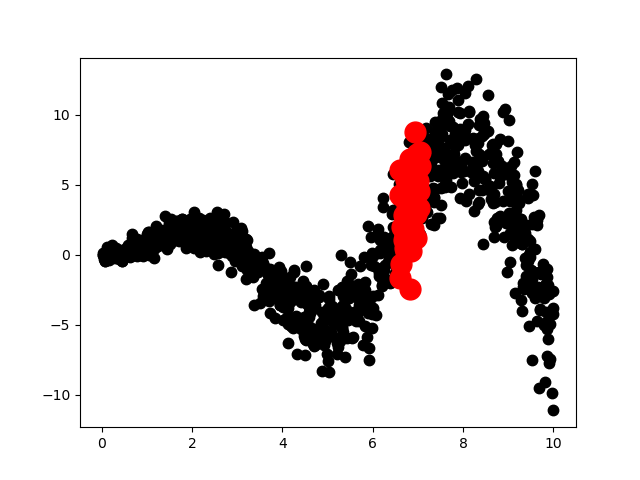}}
\subfloat{\includegraphics[width=0.2\textwidth]{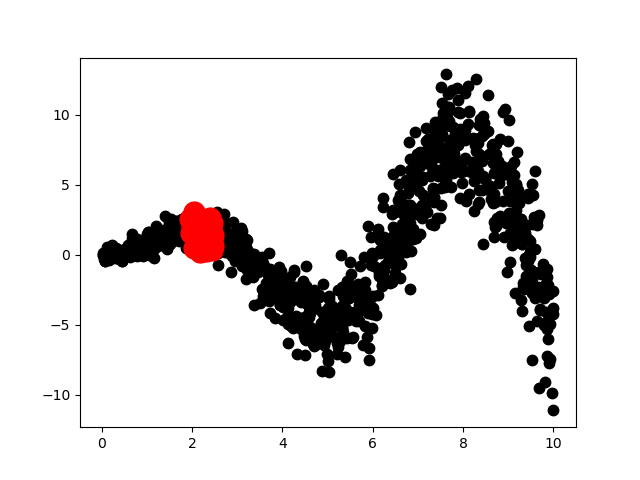}}
\subfloat{\includegraphics[width=0.2\textwidth]{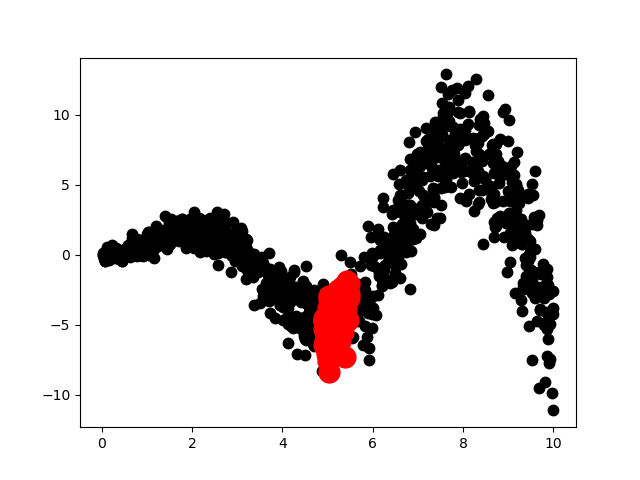}}
\subfloat{\includegraphics[width=0.2\textwidth]{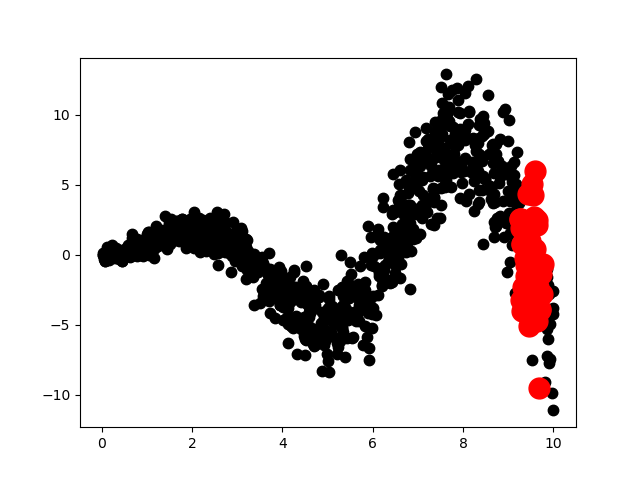}}
\vspace{-0.5cm}
\subfloat{\includegraphics[width=0.2\textwidth]{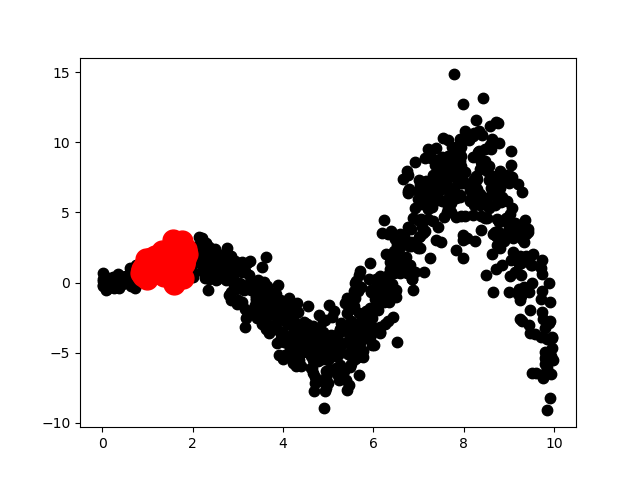}}
\subfloat{\includegraphics[width=0.2\textwidth]{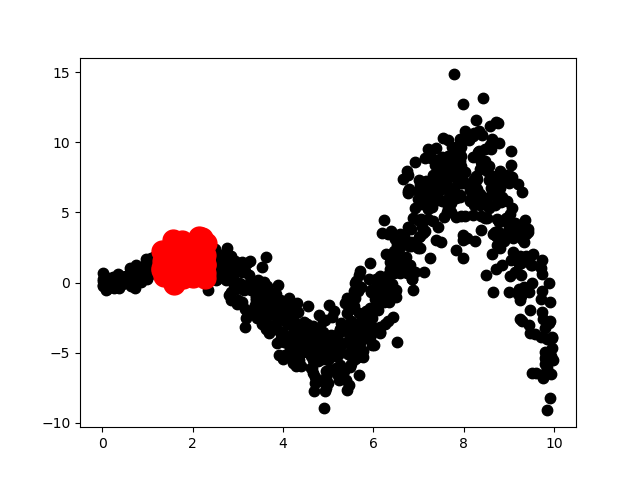}}
\subfloat{\includegraphics[width=0.2\textwidth]{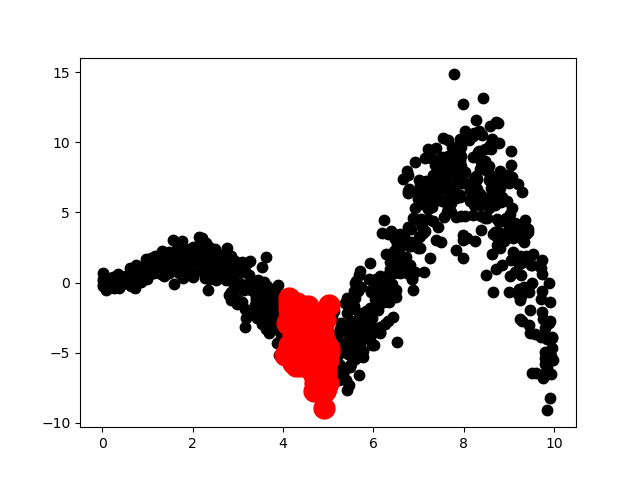}}
\subfloat{\includegraphics[width=0.2\textwidth]{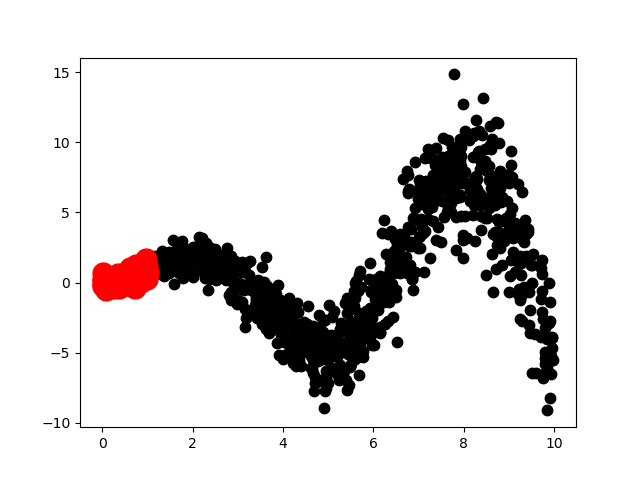}}
\subfloat{\includegraphics[width=0.2\textwidth]{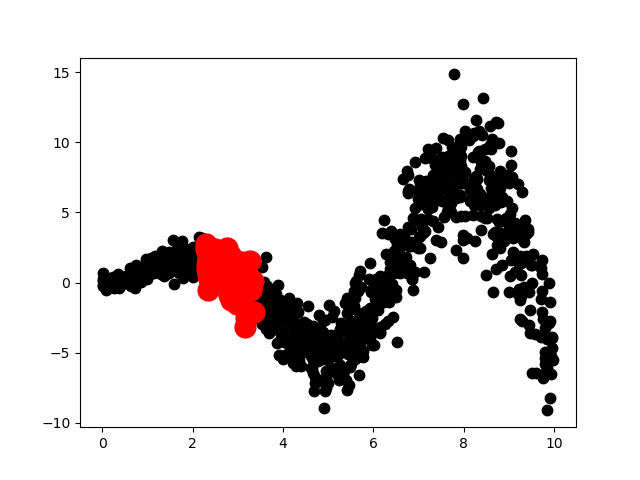}}
\caption{The effect of changing the size of the secondary sampling unit. From top to bottom: $ssu=[10,50,100]$. Each column corresponds to a sample from the locality sampler.}
\label{fig:ssu}
\end{figure}

In Algorithm \ref{alg:locality_sampler} a pseudoimplementation of our proposed locality sampler can be seen. The two important parameters in this algorithm are the primary sampling units (psu) and secondary sampling units (ssu). In \FIG\ref{fig:psu} and \ref{fig:ssu} we visually show the effect of these two parameters. 

For all our experiments we set psu=3 and ssu=40 when we are training the mean function and for variance function we set psu=1 and ssu=10.  

\begin{algorithm}
	\caption{Locality-sampler}
	\label{alg:locality_sampler}
	\textbf{Input} $N$ datapoints, a metric $d$ on feature space $\mathbb{R^D}$, integers $m,n,k$.
	\begin{algorithmic}[1]
		\State For each datapoint calculate the $k$ nearest neighbors under the metric $d$.
		\State Sample $m$ primary sampling units with uniform probability without replacement among all $N$ units.
		\State For each of the primary sampling units sample $n$ secondary sampling units among the primary sampling units $k$ nearest neighbors with uniform probability without replacement.
	\end{algorithmic}
	\textbf{Output} All secondary sampling units is a sample of at most $m\cdot n$ points. If a new sample is needed repeat from Step 2.
\end{algorithm}

\newpage

\section{Implementation details for regression experiments}
All neural network based models were implemented in Pytorch \citep{pytorch}, except for the Baysian neural network which was implemented in Tensorflow \citep{tensorflow}. GP models where implemented in GPy \citep{gpy}. Below we have stated details for all models:

\begin{description}
\item[\textbf{GP}] Fitted using a ARD kernel and with default settings of GPy.
\item[\textbf{SGP}] Fitted using a ARD kernel and with default settings of GPy. Number of inducing points was set to $\min(500, |\mathcal{D}_{train}|)$.
\item[\textbf{NN}] Model use two networks, one for the predictive mean and one for the predictive variance. Model was trained to optimize the log-likelihood of data. 
\item[\textbf{BNN}] We use a standard factored Gaussian prior for the weights and use the Flipout approximation \citep{wen2018flipout} for the variational approximation.
\item[\textbf{MC-Dropout}] Model use a single network, where we place dropout on all weights. The dropout weight was set to 0.05. The model was trained to optimize the RMSE of data.
\item[\textbf{Ens-NN}] Model consist of an ensemble of 5 individual NN models, each modeled as two individual networks. Each are trained to optimize the log-likelihood of data. Only difference between ensemble models is initialization. 
\item[\textbf{Combined}] Model use three networks, one for the mean function, one for the $\alpha$ parameter and one for the $\beta$ parameter. We set the number of inducing points to $\min(500, |\mathcal{D}_{train}|)$. For the $\gamma$ in the scaled-and-translated sigmoid function $\nu(x)$ we initialize it to 1.5, and try to minimize it during training. Model was trained to optimize the log-likelihood of data. 
\end{description}

For each neural network based approach, we follow the experimental setup \citep{Hernandez2015backprob}. All individual networks was modeled a single hidden layer MLPs with 50 neurons for all other datasets than "Protein" and "Year" where we use 100 neurons. Except for the output of each network, the activation function used is ReLU. For the output of the mean networks, no activation function is applied. For the output of the variance network, the Softplus activation function is used to secure positive variance. All neural network based models where trained using the Adam optimizer \citep{adam} with a learning rate of $10^{-1}$ using a batch size of 256. All models were trained for 10.000 iterations. 

The code can found here: \url{https://github.com/SkafteNicki/john}.

\section{Generative network architecture}
Pixel values of the images were scaled to the interval [0,1]. Each pixel is assumed to be i.i.d. Gaussian distributed. For the encoders and decoders we use multilayer perceptron networks, see table below.

\begin{table}[h!]
\centering
\begin{tabular}{l|cccc}
										& Layer 1 			& Layer 2				& Layer 3 				&	Layer 4 \\ \hline
$\bmu_{encoder}$			&	512	(BN + ReLU)		& 256	(BN + ReLU)		& 128 (BN + ReLU)			&	d  (BN + Linear) \\
$\bsigma^2_{encoder}$	&	512	(BN + ReLU)		& 256	(BN + ReLU)		& 128	(BN + ReLU)			& d  (Softplus) \\ \hline
$\bmu_{decoder}$			& 128	(BN + ReLU)		& 256	(BN + ReLU)		& 512	(BN + ReLU)			& D (ReLU) \\
$\bsigma^2_{decoder}$ & 128	(BN + ReLU)		& 256	(BN + ReLU)		& 512	(BN + ReLU)			& D (Softplus) \\
\end{tabular}
\end{table}

The numbers corresponds to the size of the layer and the parenthesis states the used activation function and whether or not batch normalization was used. $D$ indicates the size of the images \ie $D=width \times height \times channels$. For MNIST and FashionMNIST these are 28,28,1 and for CIFAR10 and SVHN these are 32,32,1. $d$ indicates the size of the latent space. For MNIST and FashionMNIST we set $d=2$ for visualization purpose and for CIFAR10 and SVHN we set $d=10$ to be able to capture the higher complexity of these datasets.

To train the networks we used the Adam optimizer \citep{adam} with learning rate $10^{-3}$ and a batch size of 512. We train for 20000 iterations without early stopping. Additionally, we use warm-up for the KL term \citep{ladder}, by scaling it with $w=\text{min}\left(1,\frac{it}{warmup}\right)$ where $it$ is the current iteration number and $warmup$ was set to half the number of iterations. This secures that we converge to stable reconstructions before introducing too much structure in the latent space.

\section{Artificial Data}
\begin{figure}[h!]
	\centering
	\subfloat{\includegraphics[width=0.48\textwidth]{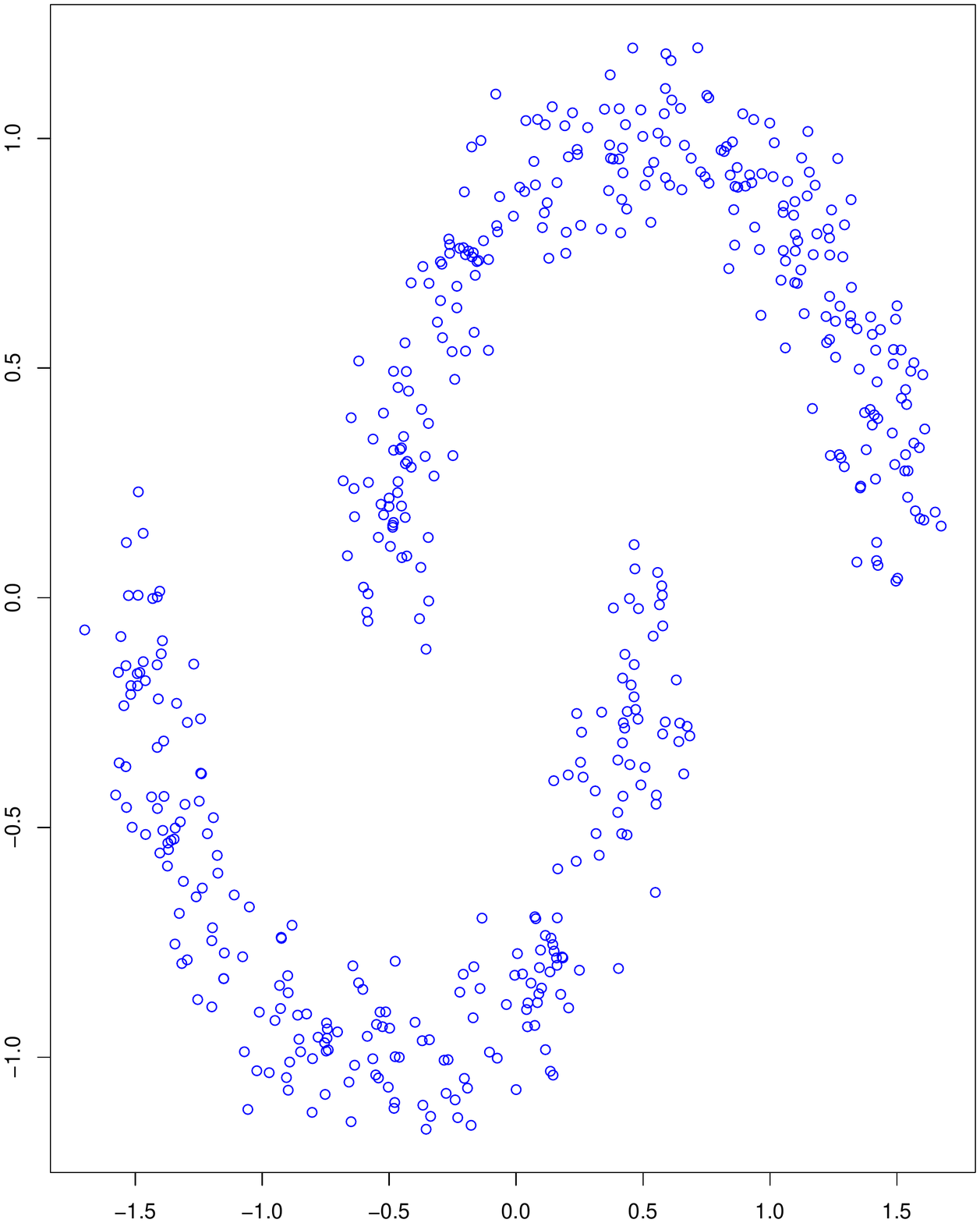}}
	\subfloat{\includegraphics[width=0.48\textwidth]{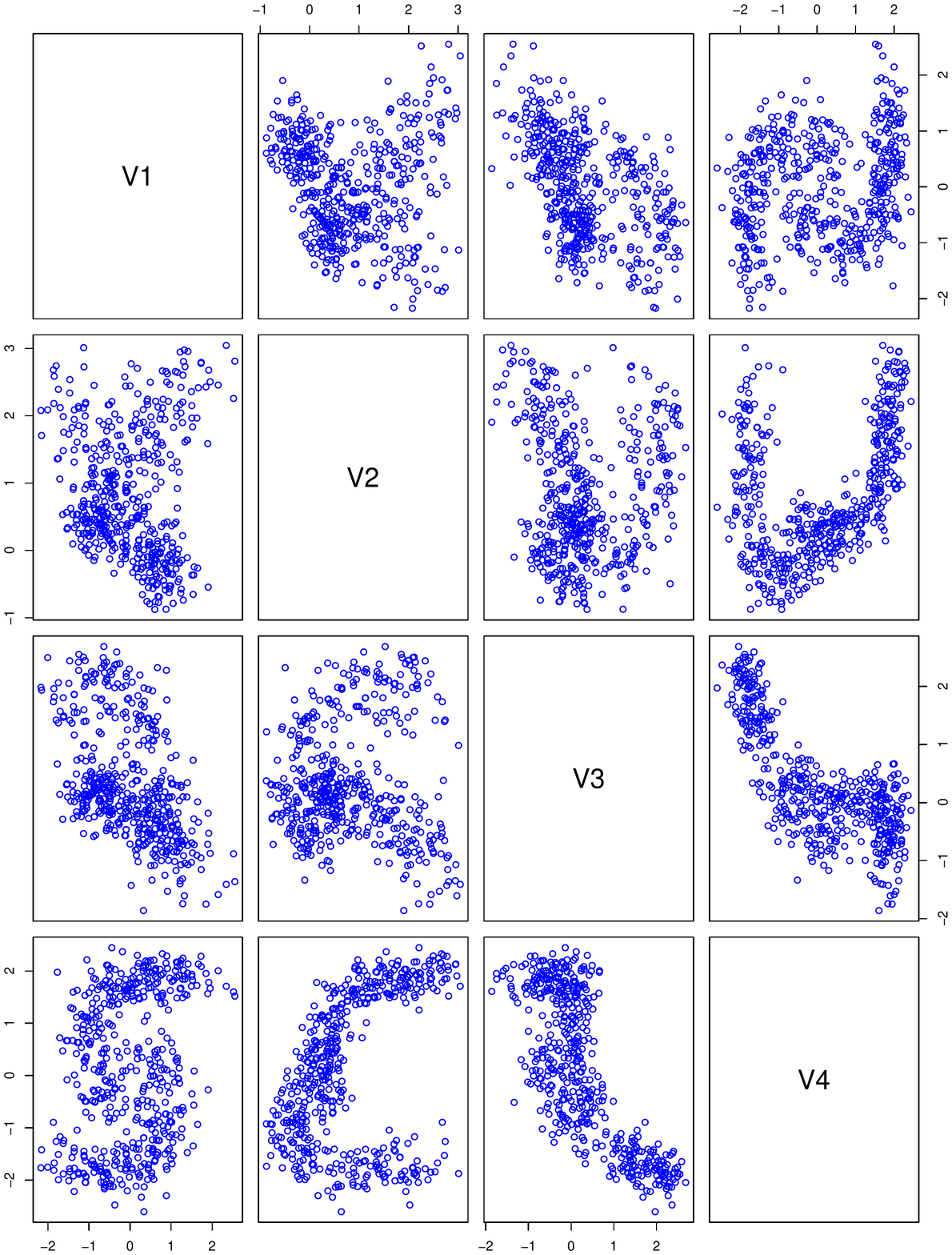}}
	\caption{An example of the two moon data, and its transformation into $\mathbb{R}^4$. Shown as pairwise scatterplots.}
	\label{fig:art-data}
\end{figure}

The data considered in Section 4 are generated in the following way: first we sample points in $\mathbb{R}^2$ in a two-moon type way. See details in Algorithm \ref{twomoonsampler}. We generate 500 points in this way to establish a 'known' latent space. We then map these to four dimensions $(v_1,v_2,v_3,v_4)$ by 
\begin{align}
	v_1(z_1,z_2) &= z_1 - z_2 + \epsilon\cdot\sqrt{0.03 + 0.05\cdot(3 + z_1)},\\
	v_2(z_1,z_2) &= z_1^2 + \frac{1}{2}z_2 + \epsilon\cdot\sqrt{0.03 + 0.03\cdot\|\mathbf{z}\|_2},\\
	v_3(z_1,z_2) &= z_1z_2 - z_1 + \epsilon\cdot\sqrt{0.03 + 0.05\cdot\|\mathbf{z}\|_2},\\
	v_4(z_1,z_2) &= z_1 + z_2 + \epsilon\cdot\sqrt{0.03 + \frac{0.03}{0.2 + \|\mathbf{z}\|_2}},
\end{align}
where all $\epsilon\sim\mathcal{N}(0,1)$ and independent.
\begin{algorithm}
	\caption{Two moon sampler}\label{twomoonsampler}
	\begin{algorithmic}[1]
		\State Sample $U\sim\text{Bernoulli}(0.5)$.
		\If {$U=1$}
			\State Set $c=(0.5,0)$ and sample $\alpha_1\sim\text{unif}[0,\pi]$.
			\State Let $z = c + (\cos(\alpha_1),\sin(\alpha_1))$, and sample $\alpha_2\sim\text{unif}[0,2\pi]$ and $u\sim\text{unif}[0,1]$.
			\State Let $z = z + \frac{u}{4}\cdot (\cos(\alpha_2),\sin(\alpha_2))$.
		\Else
			\State Set $c=(-0.5,0)$ and sample $\alpha_1\sim\text{unif}[\pi,2\pi]$.
			\State Let $z = c + (\cos(\alpha_1),\sin(\alpha_1))$, and sample $\alpha_2\sim\text{unif}[0,2\pi]$ and $u\sim\text{unif}[0,1]$.
			\State Let $z = z + \frac{u}{4}\cdot (\cos(\alpha_2),\sin(\alpha_2))$.
		\EndIf
		\Return $z$
	\end{algorithmic}
\end{algorithm}
A typical dataset from this procedure is shown in Figure \ref{fig:art-data}.

